\documentclass[fleqn,10pt]{wlscirep}

\usepackage[utf8]{inputenc} 
\usepackage[T1]{fontenc}    
\usepackage{hyperref}       
\usepackage{url}            
\usepackage{booktabs}       
\usepackage{amsfonts}       
\usepackage{nicefrac}       
\usepackage{microtype}      
\usepackage{graphicx}
\usepackage{placeins}
\usepackage{subcaption}

\newcommand{\rulesep}{\unskip\ \hrule\ }

\usepackage{xspace}
\usepackage{amssymb}
\newcommand{\molformer}{\textsc{MoLFormer}\xspace}
\newcommand*{\greencheck}{\textcolor{green}{$\checkmark$}\xspace}
\newcommand*{\redcross}{\textcolor{black}{$\times$}\xspace}

\definecolor{rc1}{RGB}{4, 142, 255}
\definecolor{rc2}{RGB}{255, 104, 3}
\newcommand\rankone[1]{\textcolor{rc1}{\textbf{#1}}}
\newcommand\ranktwo[1]{\textcolor{rc2}{\textbf{#1}}}
\newcommand{\RNum}[1]{\uppercase\expandafter{\romannumeral #1\relax}}

\usepackage{siunitx}

\title{How Much Structural Information \\ Large Scale Molecular Language Representations Can Capture? }
\title{Can Large Scale Molecular Language Models Universally Predict Chemical Properties?}
\title{Do Large Scale Molecular Language Representations Capture Important Structural Information? }

\title{\molformer: Large-Scale Chemical Language Representations that capture structural properties of Molecules}

\title{Are Large Scale Chemical Language Representations \\ Universal Predictors of  Molecular Properties?}
\title{\molformer: Large-Scale Chemical Language Representations  Capture Molecular Structure and  Properties}
\title{Large-Scale Chemical Language Representations  Capture Molecular Structure and  Properties}
\author[1,*]{Jerret Ross}
\author[1]{Brian Belgodere}
\author[1]{Vijil Chenthamarakshan}
\author[1]{Inkit Padhi}
\author[1]{Youssef Mroueh}
\author[1,*]{Payel Das}
\affil[1]{IBM Research, Yorktown Heights, NY 10598, USA}

\affil[*]{rossja@us.ibm.com and daspa@us.ibm.com}

\begin{abstract}
Predicting the  properties of a chemical molecule is of great importance in many applications, including drug discovery and material design. 
Machine learning-based models promise to enable more accurate and faster molecular property predictions than the current state-of-the-art techniques, such as Density Functional Theory calculations or wet-lab experiments. 
Various supervised machine learning models, including graph neural nets, have demonstrated promising  performance in molecular property prediction tasks. However, the vast chemical space and the limited availability of property labels make supervised learning challenging, calling for learning a general-purpose molecular representation. Recently, unsupervised transformer-based language models pre-trained on large unlabeled corpus have produced state-of-the-art results in many downstream natural language processing tasks. Inspired by this development, we present molecular embeddings obtained by training an efficient transformer encoder model, MoLFormer, which uses rotary positional embeddings. This model employs a linear attention mechanism, coupled with highly distributed training, on SMILES sequences of 1.1 billion unlabeled molecules from the PubChem and ZINC datasets. Experiments show that utilizing the learned molecular representation outperforms existing baselines on downstream tasks, including supervised and self-supervised graph neural net baselines and language models, on several  classification and regression tasks from ten benchmark datasets while performing competitively on two others. Further analyses, specifically through the lens of attention, demonstrate that MoLFormer trained on chemical SMILES indeed learns the spatial relationships between atoms within a molecule. 
These results provide encouraging evidence that the large-scale molecular language models can capture sufficient  chemical and structural information to predict various distinct molecular properties, including quantum-chemical properties.
\end{abstract}

\begin{document}
\pagenumbering{gobble}
\maketitle

\section*{Main}\label{sec:introduction}

Machine Learning (ML) has emerged as an appealing, computationally efficient approach for predicting molecular properties, with implications in drug discovery  and material engineering. 
ML models for molecules can be trained directly on pre-defined chemical descriptors,
such as unsupervised molecular fingerprints \cite{rogers2010extended}, or hand-derived derivatives of geometric features such as a Coulomb Matrix (CM) \cite{rupp2012fast}. 
However, more recent ML models have focused on automatically learning the features either from the natural  graphs that encode the connectivity information or from the line annotations of molecular structures, such as the popular SMILES\cite{weininger1988smiles} (Simplified Molecular-Input Line Entry System) representation. SMILES defines a character string representation of a molecule by performing a depth-first pre-order spanning tree traversal of the molecular graph, generating symbols for each atom, bond, tree-traversal decision, and broken cycles. Therefore, the resulting character string corresponds to a flattening of a spanning tree of the molecular graph. Learning on SMILES has been widely adopted for molecular property prediction \cite{goh2017smiles2vec, ozturk2018deepdta, paul2018chemixnet, shin2019self} as SMILES is generally more compact than other methods of representing structure, including  graphs. 
Additionally, meaningful substructures such as branches, cyclic structures, and chirality information are explicitly represented in SMILES strings, which is not the case for the graph representation.  

However, the SMILES grammar is complex and restrictive;  most sequences  over the appropriate character set do not belong to well-defined molecules. Alternative string-based representations exist, such as SMARTS \cite{daylight2007smarts} and SELFIES \cite{Krenn_2020_selfies}.
\textcolor{black}{Comparing  benefits of these alternative representations with respect to SMILES is an active area of research. For example  \cite{gao2022sample}, focusing on molecular optimization tasks on the learned representation space, suggested no obvious shortcoming of SMILES with respect to SELFIES in terms of optimization ability and sample efficiency, particularly when the language model is more advanced.} 
Nevertheless, string-based representations are thought to not be topologically-aware, while graphs are.   Due to these limitations, deep chemical language models  may focus on learning the 
grammar of molecular strings and not the implicit topological structure of the molecular graphs. Accordingly, while string-based  
deep neural nets have been  employed in  predicting molecular  properties   \cite{ozturk2018deepdta,paul2018chemixnet, shin2019self, JO202065}, they are typically outperformed 
by  graph neural networks (GNNs) \cite{Duvenaudneurips15} and their variants \cite{ defferrard2017convolutional,kipf2017semisupervised,li2015gated,velickovic2018graph, hamilton2018inductive, gilmer2017neural,schlichtkrull2017modeling, liao2019lanczosnet,chen2019utilizing}. 
GNN frameworks can be generally viewed as “message passing”, which  includes local neighborhood information aggregation and information updates across different levels of granularity, e.g., nodes, edges, or the full graph, according to the graph's connectivity structure. 

 One challenge with supervised training of GNNs and language models for molecular property prediction is the scarcity of labeled data. Label annotation of molecules is typically expensive and this problem is compounded by the fact that   
 the  size of the space consisting of plausible chemicals in need of annotation is astronomically large ($10^{60}$ to $10^{100}$) \cite{kirkpatrick2004chemical}. Such a scenario creates the need for  molecular representation learning which can be generalizable to various property prediction tasks in an un-/self-supervised setting. 
  The recent success of large transformer-based \cite{vaswani2017attention} foundation  models~\cite{foundationmodels}, using the paradigm of learning a task-agnostic language representation, obtained by pre-training on large unlabeled corpora and subsequently using it for fine-tuning on downstream tasks of interest, has been extended to other domains.

 
 Pre-trained Language Models (LMs)\cite{chithrananda2020chemberta} and GNNs \cite{wang2021molclr}  have only recently started to emerge for predicting molecular properties. However, to what extent pre-trained LMs, trained on a large  corpus of billions of molecules, are able to capture the molecule-property relationships across 
 various downstream tasks remains unexplored. 
 
Towards this direction, here we present molecular SMILES transformer models referred to as \molformer (Molecular Language transFormer). We name our best performing \molformer variant \molformer-XL. \molformer-XL was obtained using an efficient linear attention mechanism trained on a large corpus of 1.1 billion molecules (see Figure \ref{fig:pipeline_overview}). 
Results show, for the first time, that pre-trained transformer encoders of molecular SMILES perform competitively with existing supervised or unsupervised LM and GNN baselines on predicting a wide variety of molecular properties, including quantum-mechanical properties.

Our main contributions are: 
\begin{itemize}\setlength{\itemsep}{-0.6pt}
 \item We train a large-scale and efficient  Molecular Language model transFormer (\molformer) on over a billion molecules, with relatively limited hardware resources (up to 16 V100 GPUs). 
 We owe our scalability and speedups to efficient linear time attention, adaptive bucketing of batches, and open-source parallelization provided in PyTorch Lightning 
 and NCCL. With the combination of bucketing and linear attention we are able to achieve a batch size of 1600 molecules per GPU. Using 16 GPUs we need $208$ hours to complete 4 epochs of pre-training for \molformer-XL. To complete training in the same amount of time without bucketing and linear attention we would be limited to less than $50$ molecules per GPU and require over $1000$ GPUs for the task. 
    \item We explore the difference between absolute and relative position embeddings in representing molecular SMILES. We also provide a new, efficient, and accurate linear attention approximation of the recently proposed relative position RoFormer \cite{su2021roformer}. 
    \item We perform extensive experimentation and ablation studies  on  several classification and regression tasks from 10 benchmark datasets, covering quantum mechanical, physical, biophysical, and physiological property prediction of small molecule chemicals from MoleculeNet~\cite{wu2018moleculenet}.
    \item  Our results provide encouraging evidence that  \molformer representations can accurately capture sufficient chemical and structural information to predict a diverse range of chemical properties. Furthermore, the performance of \molformer is either better or on par with state-of-the-art GNNs 
    that learn from  
    precise graph topology information and beyond (e.g., bond distances).
    \item We provide further analyses to demonstrate that \molformer can capture substructures, as well as spatial interatomic distances within a molecule from 
    SMILES annotations only.  
\end{itemize}


To our knowledge, the present study is the first one that explores the representational power of pre-trained chemical language models on predicting a broad range of downstream molecular properties from quantum chemical to physiological. In particular, predicting quantum-chemical properties from SMILES strings alone is non-trivial, as those properties are largely dependent on the accurate 3D molecular geometric information, which is considered privileged information and not available in general.  

\section*{Results and Discussion}\label{sec:experiments}
\subsection*{MoLFormer Framework}
\begin{figure}[!ht]
    \centering
    \includegraphics[width=\textwidth]{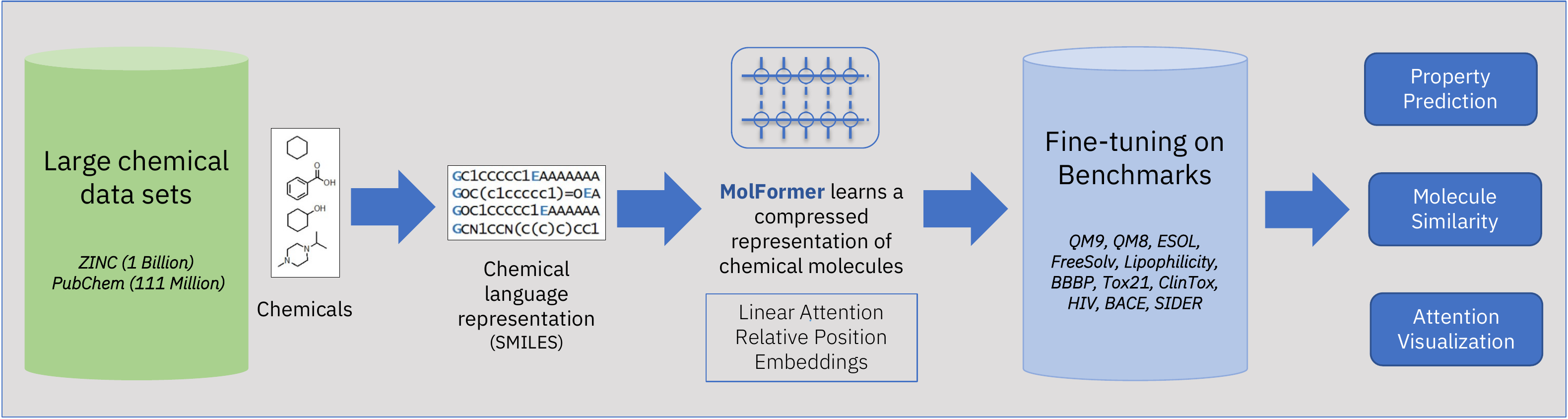}
    \caption{} 
    \label{fig:pipeline_overview}
\end{figure}

\begin{figure*}[ht]
\centering
  \includegraphics[scale=0.55]{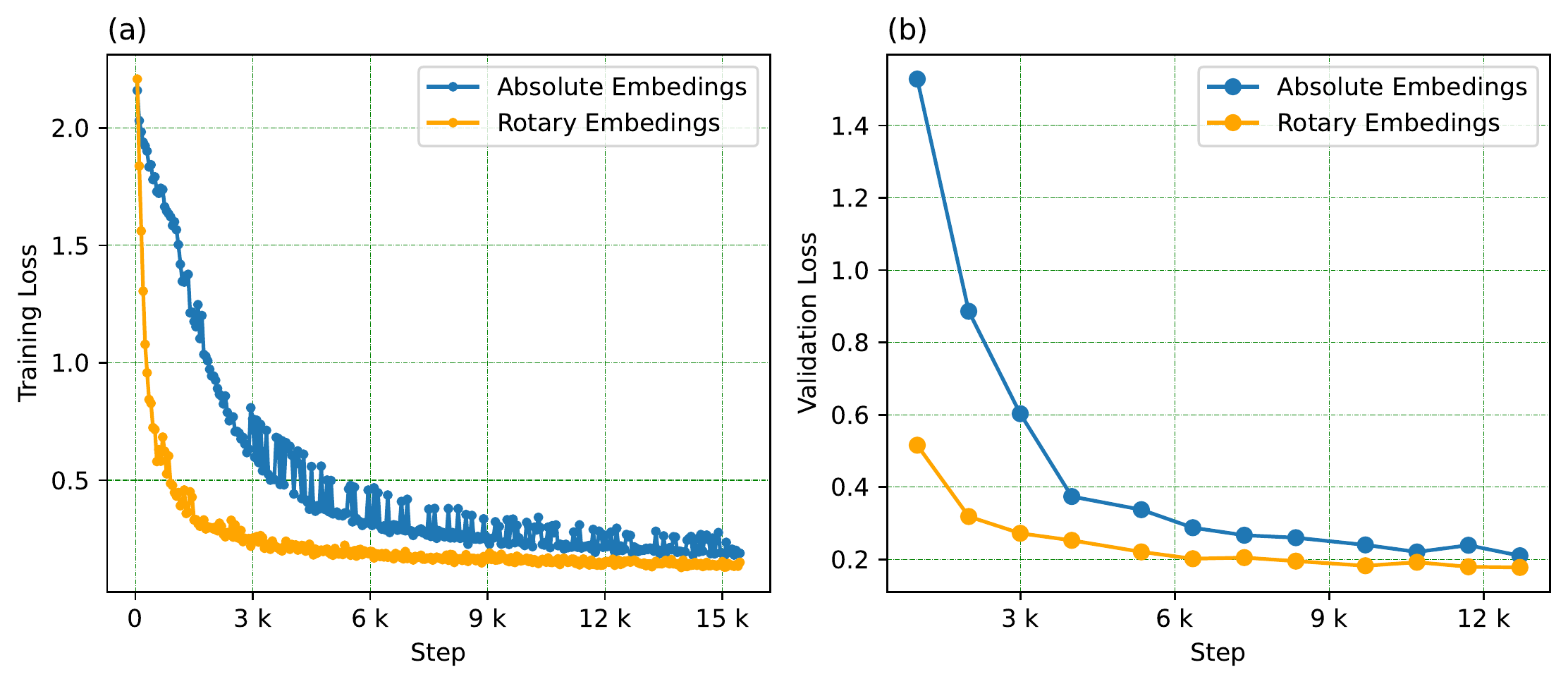} 
\caption{ 
} 
\label{fig:training_fig}
\end{figure*}

The  goal of \molformer is to learn a universal molecular representation from large scale  chemical SMILES data and then evaluate the representation on various downstream molecular property prediction tasks, as shown in Figure \ref{fig:pipeline_overview}. 
To do so,  \molformer model is developed using the masked language model framework~\cite{liu2019roberta, devlin-etal-2019-bert}, which randomly masks a certain percentage of tokens within a SMILES sequence during training and then predicts those tokens. The masked language modeling thus exploits self-supervision and enables contextual learning.  To allow better contextual learning and faster training, rotary positional embedding \cite{su2021roformer} was used instead absolute positional embedding, along with linear attention \cite{linear_attention} (See Methods and Supplementary Information for further details on model architecture and training). We saw increased stability and faster convergence in training loss behavior when pre-training using rotary embeddings in contrast to absolute embeddings as observed in  Figure \ref{fig:training_fig}.
To demonstrate the effectiveness of the pre-trained \molformer as a universal and task-agnostic molecular representation,
we benchmarked its adaptation performance on  numerous challenging classification and regression tasks from MoleculeNet~\cite{wu2018moleculenet}.  Details of the benchmark datasets can be found in SI Section~\ref{app:dataset}.

\subsection*{Derivation of \molformer Embeddings}
We encode a chemical SMILES by extracting the mean of all embeddings of the last hidden state from the encoder model. The resulting embedding is used  for all downstream tasks. The downstream tasks themselves can be divided into two categories, The first category being called \textit{Frozen}  and the second being called \textit{Fine-tuned}. The  \textit{Frozen} setting  is defined by training a fully connected model for each task, while keeping the encoder embeddings fixed. The second setting, \textit{Fine-tuned}, involves  fine-tuning the weights of the encoder model jointly with the fully connected model for each downstream task. The ideal configuration and hyperparameters for the frozen strategy are discovered through a grid search as described in SI Table \ref{tab:hyperparams}.  For the fine-tuned strategy, we use a $2$-layer fully connected network with a hidden dimension of $768$ (matching the encoder embedding) with Dropout (set to $0.1$) and GELU layers in-between, on top of a final single output dimension for regression tasks.

\subsection*{Performance of \molformer Embeddings on Downstream Tasks}
We evaluate the performance of \molformer embeddings and compare them with existing baselines on six classification and five regression tasks from the MoleculeNet benchmark \cite{wu2018moleculenet}, as discussed below. We refer to \molformer which has been pre-trained on the entire training set comprised of $\approx$ 1.1 B  molecules (all molecules from both PubChem and Zinc) as  \molformer-XL. Unless  stated otherwise, the \molformer-XL is trained with linear attention using rotary positional embeddings and the  performance reported is of the model fine-tuned on the downstream task (see Methods for details).  To predict various properties on the downstream tasks we fined-tuned the model as described in the previous section. We use the training, validation and testing  data split as defined by the MoleculeNet benchmark for all tasks (see SI \ref{app:dataset}). 

\paragraph{Classification Tasks}

We choose six classification tasks from the MoleculeNet benchmark with nine total baselines, four supervised and five self-supervised, for comparison against \molformer-XL. 
The supervised baselines consist of shallow machine learning models trained on molecular fingerprints (RF and SVM in Table \ref{tab:molclr_comp}) and graph neural nets. Among the pre-trained/self-supervised baselines, 
Hu, et al. \cite{hu2020strategies} pre-trains a Graph Isomorphism Network (GIN, a GNN that uses an MLP and weighted sum of node features in the aggregation)  on molecular graphs that includes edge features involved in aggregation.  N-gram graph \cite{liu2019ngram} uses  a simple unsupervised representation for molecules by first embedding  the nodes in a graph and then constructing a compact representation of the graph by assembling the vertex embeddings in short walks in the graph. MolCLR \cite{wang2021molclr} is a  self-supervised learning framework based on GIN, which uses contrastive loss \cite{chen2020simple, oord2018representation}. GraphMVP-C is the Graph Multi-View Pre-training (GraphMVP) framework proposed by reference ~\cite{liu2021pre}, where self-supervised learning (SSL) is performed by leveraging the correspondence and consistency between 2D topological structures and 3D geometric views.  \textcolor{black}{We have considered three other  geometry-aware GNN baselines, one supervised (DimeNet \cite{klicpera2020directional}), and two self-supervised (GeomGCL \cite{liu2021pre} and GEM \cite{fang2022geometry}).} ChemBERTa~\cite{chithrananda2020chemberta} is a pre-trained molecular language model trained on a smaller chemical dataset. Table \ref{tab:molclr_comp} documents the performance comparison of \molformer with   these  baselines on six classification benchmarks using the MoleculeNet scaffold data splits. 
\molformer-XL outperforms all baselines in three (BBBP,  ClinTox, and SIDER) out of six benchmarks and comes a close second in the other three (Tox21, HIV, and BACE).

\paragraph{Regression Tasks}
Next, we evaluate \molformer-XL on more challenging regression tasks from MoleculeNet. We report our performance on five regression benchmarks, namely QM9, QM8, ESOL, FreeSolv, and Lipophilicity, in Table \ref{tab:esol_freesolv_result}. In particular, QM9 and QM8 involve predicting several quantum chemical measures, which is considered challenging without having access to privileged 3D geometric information. Again we use the train, validation and test split as suggested in \cite{wu2018moleculenet} for these tasks. The baselines considered are  a molecular graph convolutional network (GC, a GNN that utilizes a mean-pooling over the node and its neighbors before the linear transformation)  \cite{altae2017low}, the  attentive-FP (A-FP) model \cite{xiong2019pushing}, and an MPNN variant \cite{gilmer2017neural} that  learns edge features such as pairwise interatomic distances. Results show that \molformer-XL upon task-specific fine-tuning  outperforms the existing supervised GNN baselines, specifically  GC, A-FP, and MPNN (augmented with bond distances for QM8 and QM9), on all five tasks. \textcolor{black}{Table \ref{tab:esol_freesolv_result_3D} further shows  \molformer outperforming  geometry-aware GNNs (DimeNet, GeomGCL, and GEM) on three physical property regression benchmarks. } These results, combined with \molformer-XL performance on the classification benchmarks  confirm its generalizability.


\paragraph{A Closer Look at QM9}

\textcolor{black}{Table \ref{tab:molformer_3dgnn} further compares MoLFormer-XL performance on  the QM9 atomization energies and enthalpy (internal energy/enthalpy corrected for reference atomic energy,  in eV) prediction tasks  with two exemplary supervised 3D GNNs , such as SchNet \cite{schnet} and Dimenet \cite{klicpera2020directional}. \molformer-XL trained on SMILES alone is outperformed by both those models in all of the four tasks. However, SchNet, and DimeNet, which directly encode 3D information  with specialized architecture for modeling quantum interactions, beats \molformer-XL only by roughly a factor of 8  and  by roughly a factor of 10, respectively.  This result, along with Tables \ref{tab:molclr_comp} and \ref{tab:esol_freesolv_result}, reinstates the power of learning an universal molecular representation from readily available information, such as SMILES, at a broader scale, while confirming the crucial role of   privileged geometric information for quantum-chemical energy prediction. Further, results from this comparison opens up the door for future investigations, with the goal of estimating emergence of geometric awareness in MoLFormer (see later sections) or how the expressiveness of SMILES-only MoLFormer can be further enhanced by adding partial or complete 3D geometric information.}

\subsection*{Ablation Studies}
    

 In this  section we discuss several different ablations of \molformer-XL in an attempt to provide insights into its impressive performance. The ablations we performed can be broadly divided in the following three categories  (1) the effect of size and the nature  of the pre-training data and model depth,  and (2) the results with (\textit{frozen}) and with (\textit{fine-tuned}) model fine-tuning on the downstream data, (3) the effect of absolute and rotary positional embeddings.   

\paragraph{Data/Model Size}First we investigate how pre-training dataset size affects the performance of \molformer-XL on several downstream tasks from the MoleculeNet benchmark.  To accomplish this we chose 3 different weighted combinations of the PubChem and Zinc datasets, specifically  a set consisting of  10\% of Zinc and 10\% PubChem, another with  100\% of PubChem mixed with 10\% of Zinc,  and then one with 100\% Zinc  molecules and 0\% PubChem. \textcolor{black}{We also investigate the influence of model depth by pre-training a $6$ layer model, named \molformer-Base, on the complete Zinc and Pubchem dataset}.All models are pre-trained with rotary embeddings and linear attention and then compared to \molformer-XL. Identical learning rates, data splits, optimization, etc. are used for pre-training and fine-tuning.  Tables  \ref{tab:molclr_comp_small} and \ref{tab:qm8_small} summarize these results. While \molformer-XL performs better on average, we  report two interesting observations. The first is that the model that is pre-trained on the second biggest data set, 100\% Zinc,  consistently performs worse than all other pre-trained models.  \textcolor{black}{A possible explanation for the poor performance of the model trained on only Zinc is due to the Zinc dataset consisting of a much smaller vocabulary than all other dataset combinations as well as much shorter molecules with little variance with respect to molecule length}. The other point of interest is that when \molformer-XL falls behind, it is only by a very small margin (See performance on ESOL, QM8, FreeSolv benchmarks in Table \ref{tab:qm8_small}).  \textcolor{black}{Tables \ref{tab:molclr_comp_small} and \ref{tab:qm8_small}  further show that \molformer-Base has a weaker performance than \molformer-XL in majority of tasks, implying a deeper model helps in learning.}

\paragraph{\textit{Fine-tuned} versus \textit{Frozen}} Table \ref{tab:qm9_variants} further summarizes  the two remaining ablation experiments using the QM9 benchmark. For simplicity we observe that the  \textit{fine-tuned} ablation experiments achieves such a convincing win over the \textit{frozen} experiments on all pre-training dataset sizes that we opted to only investigate fine-tuning for all other benchmarks. These results provide empirical insights onto the neural and data scaling behavior of \molformer.   

\paragraph{Position embeddings} The positional embeddings ablation results are collected in Table \ref{tab:qm9_variants}, 
 which show that \molformer with Rotary embeddings and fine-tuning is behind the Absolute positional embedding model for the smaller datasets, but then wins as the dataset size passes 1 Billion molecules.

\section*{Insights into \molformer}
\subsubsection*{Molecular Similarity Recovery} Next, we analysed the correlation between pairwise similarities estimated using the Tanimoto distance~\cite{}, a popular measure of pairwise  distance between chemicals, on the molecular fingerprints and those estimated using the Euclidean distance on the \molformer-XL embeddings. We further looked into the correlation between the number of atoms in the maximum common subgraph~\cite{} of a pair of molecules with their corresponding euclidean distance in the embedding space for a set of random molecules picked from PubChem. The results are summarized in Table \ref{tab:similarity} and show that \molformer-XL embeddings are better correlated with known molecule similarity measures when compared to ChemBERTa. These results are suggestive of \molformer embeddings being informative of chemical structure similarity. 

\subsubsection*{Attention Analyses} Finally, we inspect the average pooled attention matrices of \molformer-XL to explore the chemical information embedded in them. For this purpose, we utilize the cosine similarities between attention values and the spatial distances between atoms within a molecule from the QM9 test set. Spatial distances  are obtained from the corresponding energy-minimized geometries provided within QM9 benchmark~\cite{wu2018moleculenet}.  \molformer-XL is compared with a \molformer variant trained  with full attention and rotary embeddings on the entire PubChem+Zinc dataset.  Note that the \molformer models  here are not fine-tuned for the QM9 dataset. The  frozen \molformer  with full attention shows a much higher average MAE ($\ge$ 12) on QM9 downstream tasks, performance is particularly worse on internal energies (U and U$_0$), enthalpy (H), and free energy (G).   We present attention results separately for three different categories of interatomic spatial distances: short ($\le$ 2 \AA; that are mostly reflective of typical covalent bonds in the molecule, C-C single bond distance being 1.5 \AA), medium (2-4 \AA) and long ($\geq$ 4\AA), and summarize them in Table \ref{tab:quantitative_attention}. Interestingly,  attentions in 
\molformer with linear or full attention (and rotary positional embeddings) show strong similarity with interatomic distances in both the short and medium categories, while revealing  a weak (around 0.2) similarity with longer interatomic distances. This is an interesting observation, indicating that \molformer is able to capture spatial relations between atomic tokens that are not necessarily neighbors in the SMILES sequence. 
The observed attentions in \molformer-XL are slightly more in line with medium and long range distances, when compared to \molformer with full attention. This observation suggests \molformer-XL, with linear-attention, does in fact capture spatial relations between atoms more effectively.   

Figure \ref{fig:attention} further elaborates this point showing the average learned attention coefficients in an intermediate attention layer of \molformer-XL with rotary positional embeddings.  Attentions between different pairs of atomic tokens are compared to the corresponding covalent bond connectivity and 3D distances between atom pairs (complete attention matrices for the same molecules across all layers are shown in Figures \ref{fig:attention_map_seq1} and \ref{fig:attention_map_seq2} in SI). We chose   two  molecules from the QM9 test set whose attention values show a high cosine similarity with the medium range spatial distances for this visualization.   
Visual inspection  indicates that an aggregation of heads on the intermediate rotary attention  layer  corresponds well to  the  covalent bonding pattern, while also capturing the signature of  the   spatial relations between non-bonded atoms within a molecule. These attention analysis results suggest that 
\molformer-XL is able to  recover molecular structural information from corresponding SMILES sequence to a  significant extent. This capability likely stems from pre-training on a large corpus of chemical SMILES which also allows  \molformer-XL to learn fundamental properties of chemicals, including structural information and various downstream properties, ranging from quantum chemical to physiological. A similar observation has been reported in recent work on protein sequence modeling~\cite{Rives622803,vig2020bertology}. To our knowledge, this is the first confirmation that structural and diverse property information emerges in the representation learned by a chemical language model pre-trained on large-scale data.   
  


\section*{Conclusion}\label{sec:Conclusion}

In this work, we have explored the power of unsupervised large-scale pre-trained molecular language models at various molecular property prediction tasks.
Unlike graphs, molecular languages such as SMILES do not explicitly encode molecular topology. However, with well-designed self-supervised training on a large-scale corpus and with an   
 expressive architecture, such as a contextualized
 transformer-based  language model  with a linear attention mechanism, and a parallelized training protocol, our \molformer  can efficiently learn implicit rich structure-property relationship information.
 
 Specifically, \molformer 
 outperforms existing graph-based baselines on a wide variety of molecular regression and classification benchmarks.  To our knowledge, this is the first work that validates the power of large-scale self-supervised pre-trained molecular language models on  predicting molecular properties across the entire range from quantum chemical to physiological. Further, by analysing  the learned attentions, we show that  \molformer trained on SMILES sequences indeed is aware of interatomic relations within a molecule, even beyond the 2D topology. 
 Finally, on the large-scale learning end, we showcased with \molformer an efficient and environment-friendly use of computational resources, reducing  the number of GPUs needed to perform the training by a factor of 60 (1000 vs. 16).

\molformer has immediate potential for faster in silico screening of molecules across diverse targets, which is important for material design and drug discovery applications with 
positive societal impact. However, it should be noted that misuse of such technology  without  a proper experimental and scientific  validation in a wet lab can have harmful implications. 
Further, it has been shown that  accurate property prediction models (for example.,  for predicting toxicity) along with generative models  can be exploited for designing highly toxic molecules\cite{Urbina_dualuse}. This highlights the need for a responsible framework around the use of these emerging powerful technologies.
In addition, the present work calls for further exploration of the representational power of  \molformer in the context of its ability to  learn structural molecular information directly from chemical language and can be extended beyond the small organic molecules studied in this work. Future work will also aim to improve \molformer by employing larger models and larger training data, using improved and/or domain-specific self-supervised tasks, and using other string-based representations like SELFIES \cite{Krenn_2020_selfies}. 


 \section*{Methods}

\subsection*{Model Details}As we aim to train a large scale masked language model of chemical SMILES efficiently and effectively, while utilizing relatively limited hardware resources,  we leveraged  transformer-based neural nets~\cite{vaswani2017attention}. Transformers   process inputs through a series of blocks
alternating between  self-attention and feed-forward connections. Transformers encode the position in the sequence via a positional embedding, termed the absolute positional embedding. The input feature at a position $m$ is therefore concatenated with its corresponding absolute position embedding. 
Self-attention enables the network to construct complex representations
that incorporate context from across the sequence. 
Attention mechanisms transform the features in the sequence into queries ($q$), keys ($k$), and value ($v$) representations. These representations produce the output of  the attention at position $m$ as follows:
\[ \textnormal{Attention}_{m}(Q,K,V)=\frac{\sum_{n=1}^N\exp(\langle q_m,k_n\rangle )v_n}{\sum_{n=1}^N\exp(\langle q_m,k_n\rangle )}. \]

A well known computational bottlenecks of the vanilla transformer  \cite{vaswani2017attention} architecture is that the attention mechanism suffers from a quadratic computational cost with respect to the sequence length. Linear complexity attention models \cite{linear_attention,FAVOR} have tackled this issue utilizing kernel approximations and random feature approximations variants. This  led us to design \molformer that utilizes  an encoder based on a transformer with linear attention \cite{linear_attention}.
\molformer with linear attention  consists of $12$ layers, $12$ attention heads per layer, and has a hidden state size of $768$. A Generalized Feature map \cite{linear_attention} for the linear attention was chosen (see SI Section~\ref{app:optdetails} for details). 

As mentioned above, in a transformer architecture the dependency between tokens at different position of a (chemical) sequence is modeled under the supervision of position encoding.  
 The seminal work of  \cite{vaswani2017attention} investigated absolute  position embeddings to encode the position of a token in the sequence. More recent work  \cite{shaw-etal-2018-self,raffel,ke2021rethinking} showed that use of relative position embeddings between tokens results in improved performance. Rotary position embeddings were introduced  in RoFormer \cite{su2021roformer} as a means to enhance the relative encoding via position dependent rotations $R_{m}$ of the query and the keys at a position $m$. These rotations can be efficiently implemented as pointwise multiplications and do not result in a dramatic computational increase.

In order to leverage Rotary embeddings with linear transformers, the use of the following approximation was proposed in \cite{su2021roformer}:
\[ \textnormal{Attention}_{m}(Q,K,V)=\frac{\sum_{n=1}^N \langle R_{m} \varphi(q_m) , R_{n} \varphi(k_n) \rangle v_n }{\sum_{n=1}^N  \langle \varphi(q_m) , \varphi(k_n) \rangle  },\]
where $Q,K,V$ are the query, key, and value respectively, and $\varphi$ a random feature map.

After preliminary experimentation with this linear Roformer, we found it performed worse than its absolute position counterpart. We propose the following modification to  Roformer that we found to train more gracefully (the training loss falls faster and lower) than the original Roformer, as well as observing better performance than the model using absolute embeddings:
\[ \textnormal{Attention}_{m}(Q,K,V)=\frac{\sum_{n=1}^N \langle  \varphi(R_{m}q_m) ,  \varphi(R_{n}k_n) \rangle v_n }{\sum_{n=1}^N  \langle \varphi(R_{m}q_m) , \varphi(R_{n}k_n) \rangle  }.\]
When compared with \cite{su2021roformer} we rotate the original keys and queries instead of the transformed ones with the feature map $\varphi$. 

\subsection*{Datasets and Tokenization}
 We constructed several datasets for  pre-training by combining the  PubChem \cite{pubchem_kim} and ZINC \cite{irwin2005zinc} datasets with varying proportion from each. The PubChem dataset consists of $111$ million molecules, while the much larger ZINC dataset contains over $1$ billion molecules. To construct a vocabulary, we utilize the tokenizer from \cite{SchwallerFWD}. All molecules from both PubChem and ZINC are converted to a canonical format utilizing RDKit \cite{rdkit} then tokenized.  All unique tokens extracted from the resulting output gives us a  
 vocabulary of $2357$ tokens plus $5$ special tokens, resulting in a total of  $2362$ vocabulary tokens which are used for all pre-trained models considered in this paper, irrespective of pre-training dataset size. \textcolor{black}{In other words, all models have the same embedding capacity with a fixed vocabulary size. However, the total unique tokens that they are pre-trained on might only contain a subset of the model vocabulary capacity.} 
The post tokenization sequence length of the molecules range from $1$ to just over $2000$ tokens. We decide to restrict the sequence length range from $1$ token to $202$ tokens, special tokens inclusive, to reduce computation time. Since over $99.4$ percent of all molecules from our dataset contain less than $202$ tokens we hypothesize that the removal of molecules with more than $202$ tokens would be of minimal negative impact on pre-training.

\subsection*{ Large Scale Training and Parallelization}\label{sec:training}
For pre-training we use the masked language model method defined in \cite{devlin-etal-2019-bert}. \textcolor{black}{Initially $15\%$ of the tokens are selected for possible denoising. From that selection, $80\%$ of the tokens will be randomly selected and replaced with the \texttt{[MASK]} token, $10\%$ of the tokens will be randomly selected to be replaced with a random token, while the remaining $10\%$ of the tokens will be unchanged. }
 Training was performed for 4 epochs through the entire PubChem+ZINC dataset with a fixed learning rate of $1.6\text{e}^{-4}$ and a batch size of $1600$ molecules per GPU on a total of $16$ GPUs over $2$ servers connected via Infiniband fabric. It should be noted that as the number of GPUs utilized increased we found an increase in learning rate was necessary up to a factor of $8$.

In order to scale our training to large datasets (1 Billion+  data points), we relied on adaptive bucketing of mini-batches by sequence length, as well as parallelization via distributed training (see Supplementary Information (SI) ~\ref{app:trainingdetails} for details). Using Linear attention and bucketing allowed us to reduce the number of GPUs needed from roughly 1000 for quadratic attention with no bucketing to 16.

\section*{Data availability}
Datasets used for model pre-training and  finetuning on benchmark tasks are available at \url{https://github.com/IBM/molformer}. 

\section*{Code availability}
Python codes for MoLFormer training and fine-tuning, and python notebooks for MoLFormer attention visualization, as well as instances of pre-trained models are available  at \url{https://github.com/IBM/molformer}. For other enquiries contact the corresponding authors.
\newpage

\section*{Figures and Tables}

\begin{figure*}
     \centering
     \begin{subfigure}[b]{0.33\textwidth}
     
         \includegraphics[width=0.83\textwidth]{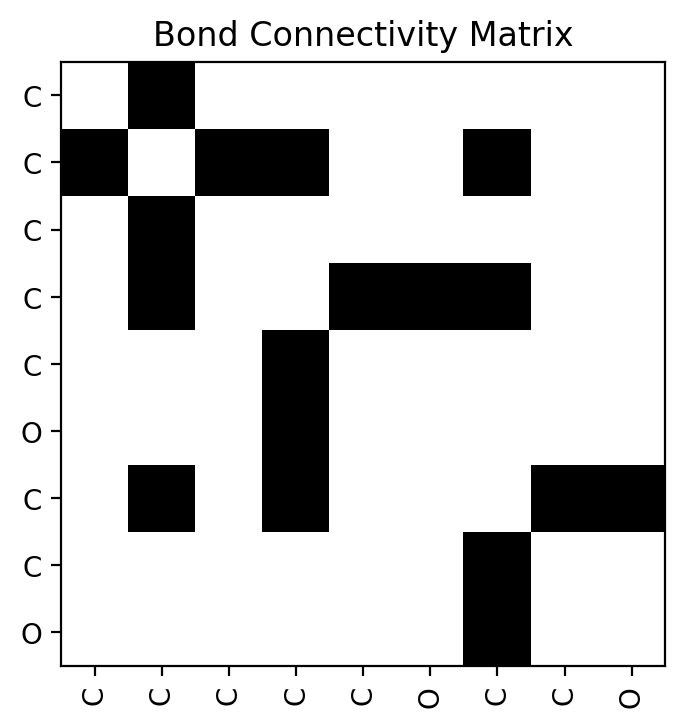}
     \end{subfigure}
     \hfill
     \begin{subfigure}[b]{0.33\textwidth}
     \caption{}
        \includegraphics[width=\textwidth]{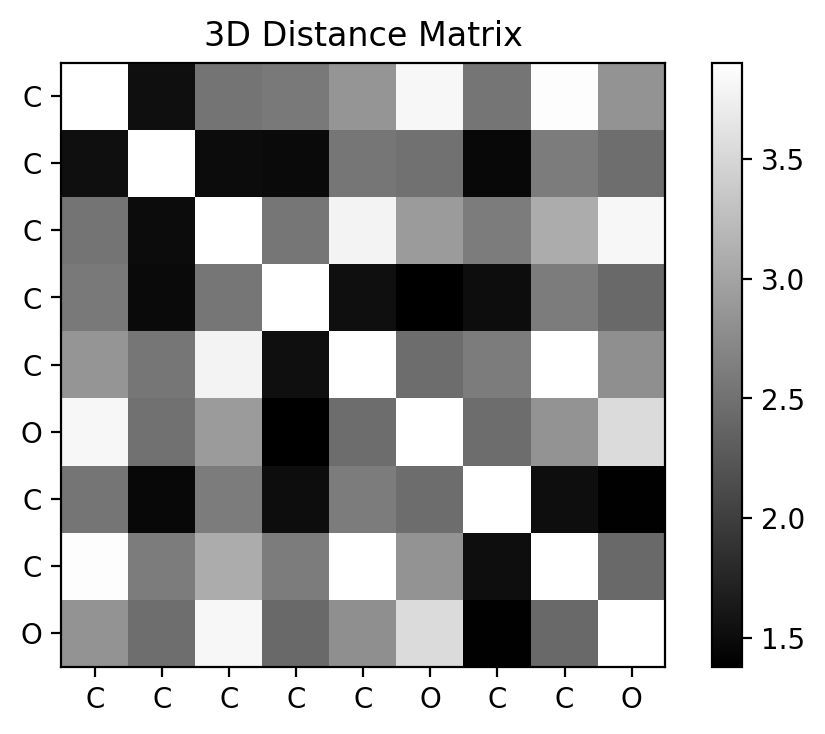}
     \end{subfigure}
     \hfill
     \begin{subfigure}[b]{0.32\textwidth}
        \includegraphics[width=\textwidth]{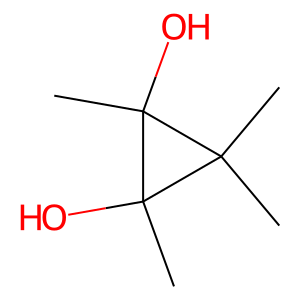}
     \end{subfigure}
     \hfill
     \begin{subfigure}[b]{0.33\textwidth}
        \includegraphics[width=\textwidth]{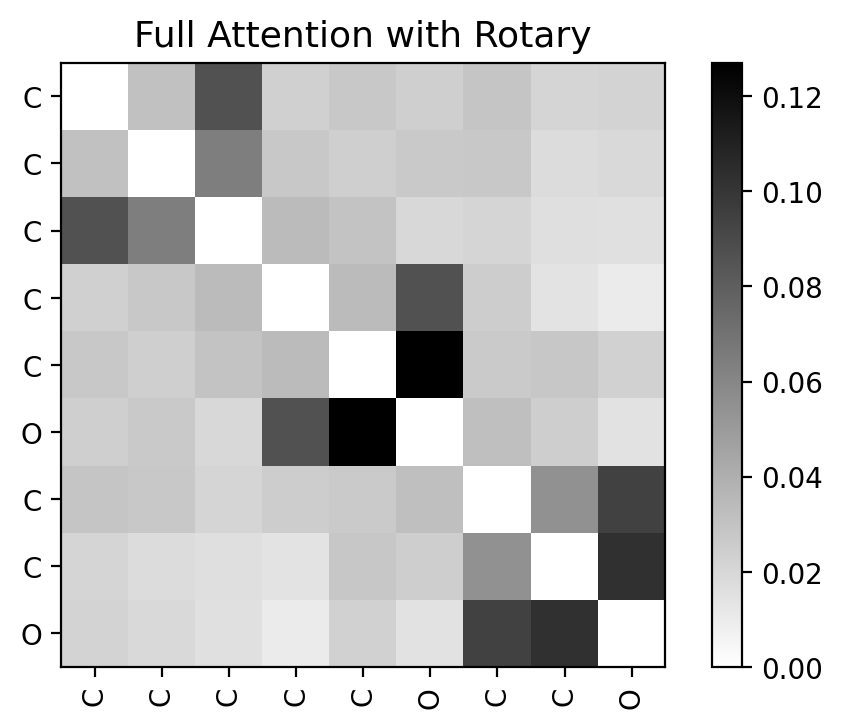}
        
     \end{subfigure}
     \hfill
      \begin{subfigure}[b]{0.33\textwidth}
         \includegraphics[width=\textwidth]{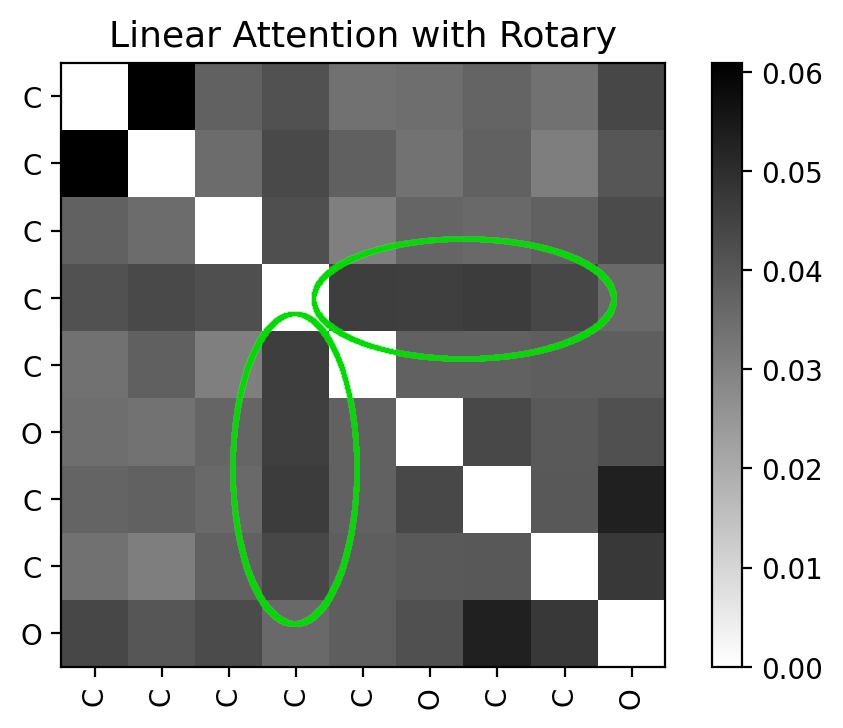}
         
     \end{subfigure}
     \hfill
     \begin{subfigure}[t]{0.32\textwidth}
        \centering
         \includegraphics[width=0.7\textwidth]{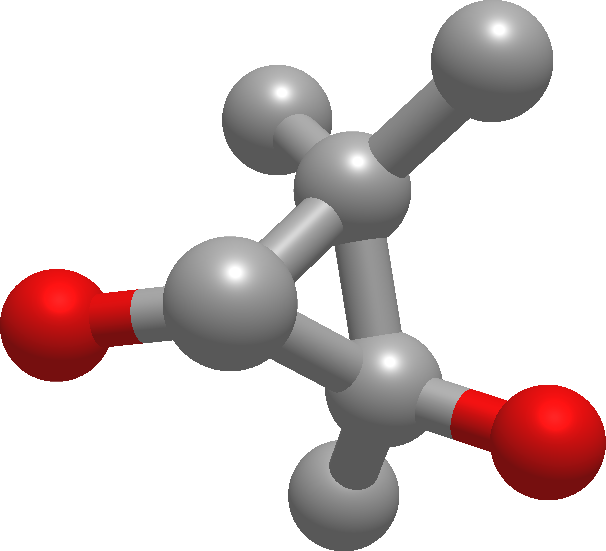}
     \end{subfigure}
     \newline
     \rulesep
     \begin{subfigure}[b]{0.33\textwidth}
         \includegraphics[width=0.83\textwidth]{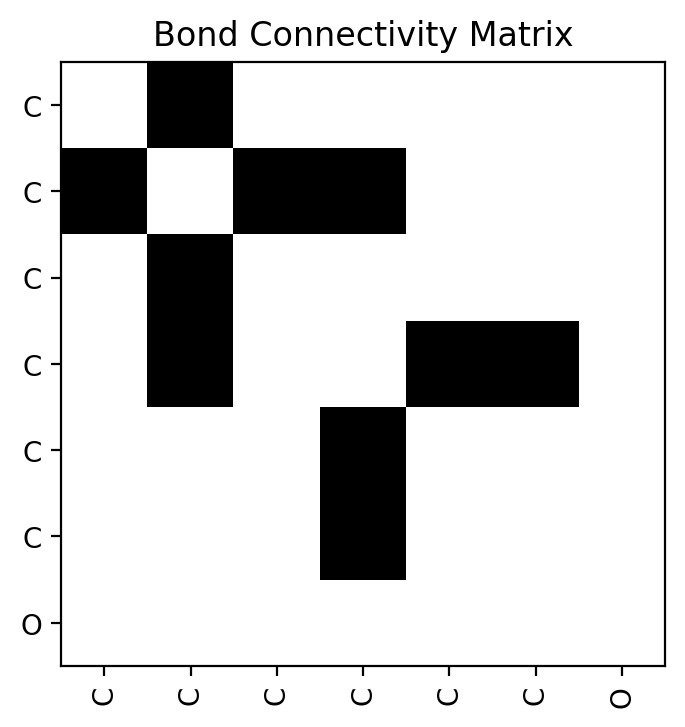}
     \end{subfigure}
     \hfill
     \begin{subfigure}[b]{0.33\textwidth}
     \caption{}
        \includegraphics[width=\textwidth]{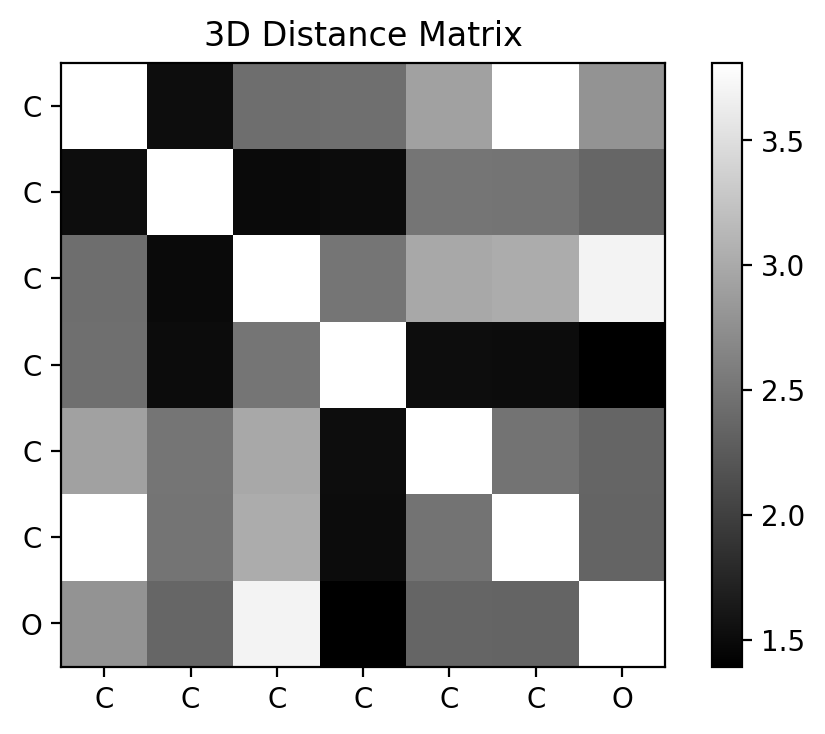}
     \end{subfigure}
     \hfill
     \begin{subfigure}[b]{0.32\textwidth}
        \includegraphics[width=\textwidth]{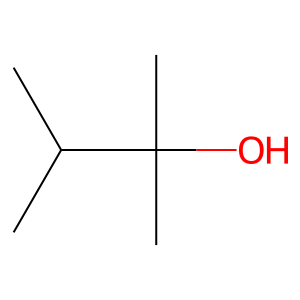}
     \end{subfigure}
     \hfill
     \begin{subfigure}[b]{0.33\textwidth}
        \includegraphics[width=\textwidth]{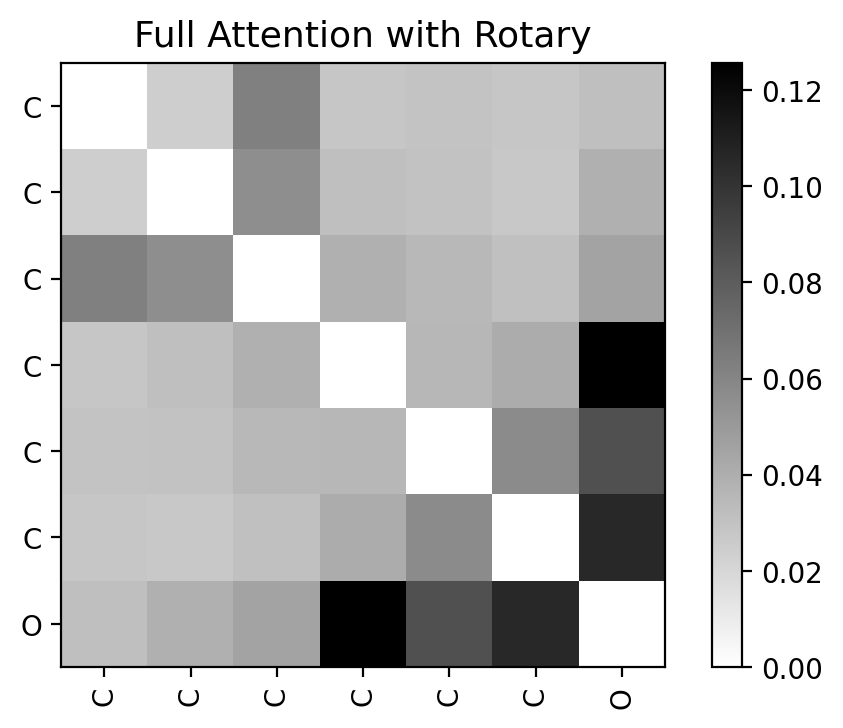}
     \end{subfigure}
     \hfill
      \begin{subfigure}[b]{0.33\textwidth}
      
         \includegraphics[width=\textwidth]{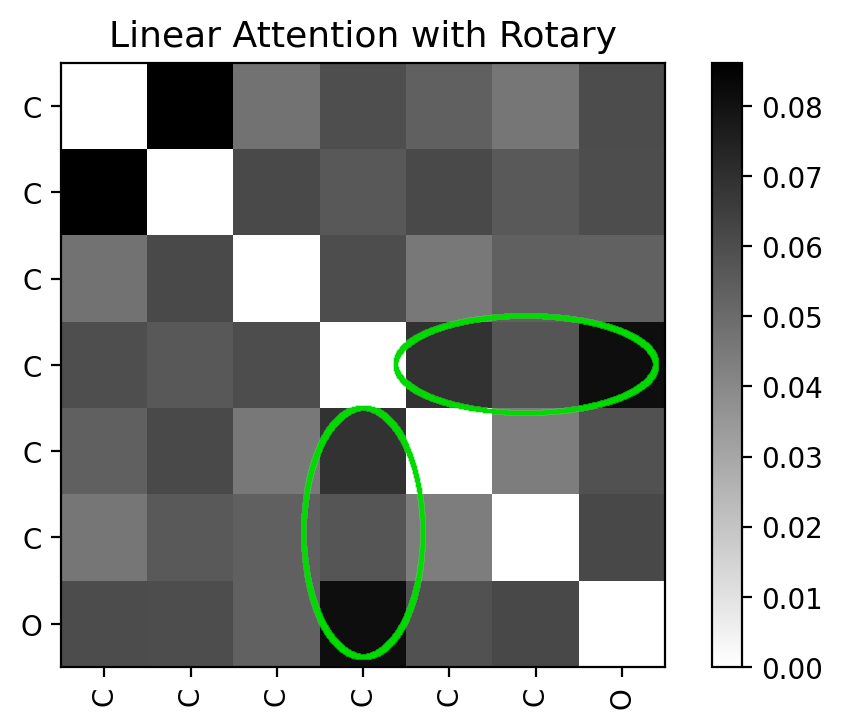}
     \end{subfigure}
     \hfill
     \begin{subfigure}[t]{0.32\textwidth}
        \centering
         \includegraphics[width=0.7\textwidth]{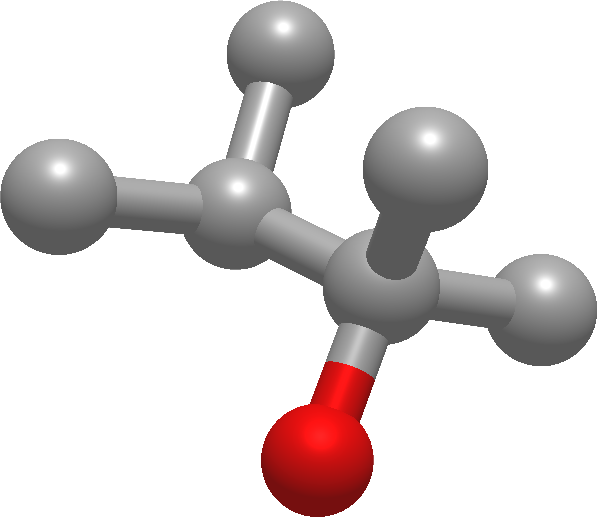}
     \end{subfigure}
     \caption{} 
     \label{fig:attention}
\end{figure*}

\newpage
\begin{table*}[ht]
\centering
\begin{tabular}{llllllll}
\hline Dataset & BBBP & Tox21 & ClinTox & HIV & BACE & SIDER \\
 Tasks & 1 & 12 & 2 & 1 & 1 & 27 \\
\hline RF & $71.4$ & $76.9 $ & $71.3 $ & $78.1 $ & $\mathbf{8 6 . 7} $ & $\mathbf{6 8 . 4} $ \\
SVM & $72.9 $ & $\mathbf{8 1 . 8} $ & $66.9 $ & $\mathbf{7 9 . 2} $ & $86.2 $ & $6 8 . 2 $ \\
MGCN \protect{\cite{DBLP:conf/aaai/Lu0WHLH19}}  & $\mathbf{8 5 . 0} $ & $70.7 $ & $63.4 $ & $73.8 $ & $73.4 $ & $55.2 $ & \\
D-MPNN \protect{\cite{doi:10.1021/acs.jcim.9b00237}} & $71.2 $ & $68.9 $ & $\mathbf{9 0 . 5} $ & $75.0 $ & $85.3 $ & $63.2 $ \\
\textcolor{black}{DimeNet \cite{klicpera2020directional}} & \textcolor{black}{--}
& \textcolor{black}{$78.0$}
& \textcolor{black}{$76.0$}
& \textcolor{black}{--}
& \textcolor{black}{--}
& \textcolor{black}{$61.5$}
 \\
\hline Hu, et al. \cite{hu2020strategies}  & $70.8 $ & $78.7 $ & $78.9 $ & $80.2 $ & $85.9 $ & $65.2 $ \\
N-Gram \cite{liu2019ngram} & $9 1 . 2 $ & $76.9 $ & $85.5 $ & $\mathbf{8 3 . 0} $ & $87.6 $ & $63.2 $\\
MolCLR \cite{wang2021molclr} & $73.6 $ & $79 . 8 $ & $9 3 . 2 $ & $80.6 $ & $\mathbf{8 9 . 0} $ & $6 8 . 0 $ \\
GraphMVP-C \cite{liu2021pre} & $72.4 $ & $74.4 $  & $77.5 $ & $77.0 $ & $81.2 $ & $63.9 $ \\
\textcolor{black}{GeomGCL \cite{liu2021pre}} & \textcolor{black}{--}
&\textcolor{black}{$\mathbf{85.0}$}
&\textcolor{black}{$91.9$}
 & \textcolor{black}{--}
& \textcolor{black}{--}
&\textcolor{black}{$64.8$}
\\
\textcolor{black}{GEM \cite{fang2022geometry}} & \textcolor{black}{$72.4$}
& \textcolor{black}{$78.1$}
& \textcolor{black}{$90.1$}
& \textcolor{black}{$80.6$}
& \textcolor{black}{$85.6$}
& \textcolor{black}{$67.2$}
 \\
ChemBerta \cite{chithrananda2020chemberta} & 64.3 & -- & 90.6 & 62.2 & -- & -- \\
\molformer-XL & $\mathbf{93.7}$ & $84.7$ & $\mathbf{94.8}$ & $82.2 $ & $88 . 21$ & $\mathbf{6 9 . 0} $\\
\hline
\end{tabular}
\caption{} 
\label{tab:molclr_comp}
\end{table*}

\begin{table*}[ht]
\centering
\begin{tabular}{lcccccc}
\toprule
Dataset & QM9 & QM8 & ESOL & FreeSolv & Lipophilicity \\
\midrule
GC & 4.3536 & 0.0148 & 0.970 & 1.40 & 0.655 \\
A-FP & 2.6355 & 0.0282  & 0.5030 & 0.736 & 0.578\\
MPNN & 3.1898  & 0.0143 & 0.58 & 1.150 & 0.7190  \\
\begin{tabular}{@{}c@{}}\molformer-XL \end{tabular} & \textbf{1.5894} & \textbf{0.0102} & \textbf{0.2787} &\textbf{0.2308} &\textbf{0.5289} \\

\bottomrule
\end{tabular}
\caption{
}
\label{tab:esol_freesolv_result}
\end{table*}

\begin{table}[ht!]
\centering
\begin{tabular}{lrcccccc}
\toprule
Distance-Category & Attention& 1 & 3 & 5 & 7 & 9 & 11 \\
\midrule
Short & Full (\greencheck Rotary) & \textbf{0.615} &	0.604 &	0.603 &	\textbf{0.615} &	0.601 &	\textbf{0.598} \\
 & Linear (\greencheck Rotary) & 0.596 & 0.597 & 0.602 & 0.597 & 0.600 & 0.594 \\
\hline
Medium & Full (\greencheck Rotary) & 0.716 &	0.724 &	0.724 &	0.716 &	\textbf{0.727} &	0.727 \\
 & Linear (\greencheck Rotary) & \textbf{0.729} & \textbf{0.728} & \textbf{0.724} & \textbf{0.727} & 0.726 & \textbf{0.730} \\
\hline
Long & Full (\greencheck Rotary) & 0.204 &	0.207 &	0.208 &	0.205 & 	0.208 &	0.210 \\
 & Linear (\greencheck Rotary)  & \textbf{0.211} & \textbf{0.210} & \textbf{0.210} & \textbf{0.211} & \textbf{0.209} & \textbf{0.210} \\
 \bottomrule
\end{tabular}
\caption{}
\label{tab:quantitative_attention}
\end{table}
\newpage
\section*{Figure and Table Captions}
~~\\
\newline \textbf{Figure 1.} Overview of MoLFormer pipeline. The transformer neural network based model is trained on the SMILES sequences corresponding to a large collection of chemical molecules from PubChem and Zinc, two public chemical databases in a self-supervised fashion. \molformer was designed with an efficient linear attention mechanism and relative positional embeddings, with the goal of learning a meaningful and compressed representation of  chemical molecules. This foundation model was then adopted to different downstream molecular property prediction  tasks via fine-tuning on task-specific data. The representative power was further tested by recovering molecular similarity using the \molformer encodings, as well as by analyzing the correspondence between the interatomic spatial distance and attention value for a given molecule. 
\newline \textbf{Figure 2.} (a) Training and (b) validation  losses of our Linear attention \molformer with \emph{rotary} (relative) and absolute position embeddings on PubChem. 
We see that both rotary and absolute \molformer have graceful training curves. Our Rotary Linear attention  \molformer leads to lower training and validation losses than  \molformer with  absolute position embeddings. 
\newline \textbf{Figure 3.} Visualization of the learned attention map (using either full or linear attention) under rotary embedding and corresponding molecular structure (bond connectivity and 3D distance in Angstroms)  for two random molecules: \texttt{`CC1(C)C(C)(O)C1(C)O'}  (a) and \texttt{`CC(C)C(C)(C)O'} (b). The attention map (only tokens that map to constituent atoms are shown for clarity), comprised of the average-pooled heads of an intermediate attention layer, exhibits awareness of both covalent bond connectivity and interatomic long-range spatial  relationship. The linear attention variant captures (encircled in green) the medium 3D range distance better in comparison to its counterpart.
\newline \textbf{Table 1.} Comparison of fine-tuned \molformer with existing supervised and pre-trained/self-supervised baselines on multiple classification benchmarks. Bold indicates the top-performing model. All models were evaluated by AUC-ROC on scaffold splits.  Baseline performances are adopted from references \protect{\cite{wang2021molclr, liu2021pre, chithrananda2020chemberta}}, '-' signifies that the values were not reported for the corresponding task.  .
\newline \textbf{Table 2.} Performance of fine-tuned \molformer and other supervised GNN baselines  on  QM9, QM8, ESOL, FreeSolv, and Lipophilicity regression benchmarks. For QM9 and QM8, we report average MAE,  while RSME is reported for the remaining tasks. Baseline performances are taken from references \protect{\cite{wu2018moleculenet, xiong2019pushing}}. Bold indicates the top-performing model.. 
\newline \textbf{Table 3.} Comparison of \molformer models with respect to  cosine similarity between the interatomic spatial distance map and the attention map, across three different distance categories for 7806 molecules from QM9 test set. Short, Medium, and Long distance categories are defined with interatomic distances in the range of $\le$2, 2-4, and 4-10 \AA, respectively. Bold indicates the top-performing model.

\newpage
\section*{Acknowledgements}
We thank IBM Research for supporting this work.
\section*{Author information}



Contributions
\newline All authors conceived the project, developed the MoLFormer framework, and designed experiments.  J.R., B.B., V.C., and I.P. performed model training, fine-tuning, and inference experiments. I.P. and P.D. performed attention map analyses. All authors analysed the results and wrote the paper.



\section*{Ethics declarations}
Competing interests
\newline The authors declare no competing interests.
\newpage
\bibliography{confs}


\newpage
\section*{Extended data}
\begin{table*}[ht!]
\centering
\begin{tabular}{llllllll}
\hline Dataset & BBBP & HIV & BACE &\textcolor{black}{SIDER}& \textcolor{black}{Clintox}& \textcolor{black}{Tox21}\\
\hline 
10\% ZINC + 10\% PubChem  & 91.5 & 81.3  & 86.6& \textcolor{black}{68.9} & \textcolor{black}{94.6} &\textcolor{black}{84.5}\\
10\% ZINC + 100\% PubChem  & 92.2& 79.2& 86.3& \textcolor{black}{69.0} & \textcolor{black}{94.7} & \textcolor{black}{84.5}\\
100\% ZINC & 89.9 & 78.4 & 87.7& \textcolor{black}{66.8} & \textcolor{black}{82.2}& \textcolor{black}{83.2}\\
\textcolor{black}{\molformer-Base} & \textcolor{black}{90.9} & \textcolor{black}{77.7} & \textcolor{black}{82.8}  & \textcolor{black}{64.8} & \textcolor{black}{61.3} & \textcolor{black}{43.2}\\
\molformer-XL & \textbf{93.7} & \textbf{82.2} & \textbf{88.2} & \textcolor{black}{\textbf{69.0}} & \textcolor{black}{\textbf{94.8}} & \textcolor{black}{\textbf{84.7}}\\
\hline
\end{tabular}
\setcounter{table}{0}
\renewcommand{\tablename}{Extended data Table}
\caption{Comparison of \molformer-XL  with fine-tuned \molformer models \textcolor{black}{that are either of smaller size or} pre-trained on smaller datasets  on BBBP, HIV, Sider, Clintox, Tox21 and BACE classification benchmarks.}
\label{tab:molclr_comp_small}
\end{table*}

\begin{table*}[ht!]
\centering
\begin{tabular}{lccccc}
\toprule
Dataset & QM9 & QM8& ESOL & FreeSolv & Lipophilicity  \\
\midrule

10\% Zinc + 10\% Pub & 1.7754 & 0.0108 & 0.3295 & 0.2221 & 0.5472 \\
10\% Zinc + 100\% Pub & 1.9093 & \textbf{0.0102} & \textbf{0.2775} & \textbf{0.2050} & 0.5331 \\
100\% Zinc & 1.9403 & 0.0124 & 0.3023 & 0.2981 & 0.5440 \\
\textcolor{black}{\molformer-Base} & \textcolor{black}{2.2500} & \textcolor{black}{0.0111} & \textcolor{black}{0.2798} &\textcolor{black}{0.2596} & \textcolor{black}{0.6492} \\
\molformer-XL & \textbf{1.5984} & \textbf{0.0102} & 0.2787 & 0.2308 & \textbf{0.5298} \\
\bottomrule
\end{tabular}
\renewcommand{\tablename}{Extended data Table}
\caption{Performance comparison of fine-tuned \molformer-XL with fine-tuned \molformer models \textcolor{black}{are either of smaller size or} pre-trained on smaller datasets on  QM9 (avg MAE), QM8 (avg MAE), ESOL (RMSE), FreeSolv (RMSE), and Lipophilicity (RMSE) regression benchmarks. 
}
\label{tab:qm8_small}
\end{table*}

\begin{table*}[ht!]
\centering
\resizebox{\textwidth}{!}{\begin{tabular}{lrrr|rrr|rrr}
\toprule
Pre-training Data $\rightarrow$ &\multicolumn{3}{c|}{QM9 Only}& \multicolumn{3}{c|}{PubChem Only} & \multicolumn{3}{c}{PubChem+ZINC}\\
Dataset Size $\rightarrow$ &\multicolumn{3}{c|}{$111$ × $10^3$}& \multicolumn{3}{c|}{$111$ × $10^6$} & \multicolumn{3}{c}{$>1.1$ × $10^9$}\\
Measure $\downarrow$ & \begin{tabular}{@{}c@{}}Frozen \\ \redcross Rotary\end{tabular} & \begin{tabular}{@{}c@{}}Fine-tuned \\ \redcross Rotary\end{tabular} & \begin{tabular}{@{}c@{}}Fine-tuned \\ \greencheck Rotary\end{tabular} & \begin{tabular}{@{}c@{}}Frozen \\ \redcross Rotary\end{tabular} & \begin{tabular}{@{}c@{}}Fine-tuned \\ \redcross Rotary\end{tabular} & \begin{tabular}{@{}c@{}}Fine-tuned \\ \greencheck Rotary\end{tabular} & \begin{tabular}{@{}c@{}}Frozen \\ \redcross Rotary\end{tabular} & \begin{tabular}{@{}c@{}}Fine-tuned \\ \redcross Rotary\end{tabular} & \begin{tabular}{@{}c@{}}Fine-tuned \\ \greencheck Rotary\end{tabular}\\ 
\midrule
\midrule
Avg MAE & 8.3808 & \textbf{2.4621} & 2.6604 &  8.2600 & \textbf{2.9680} & 3.3990 & 2.5497 & 1.8620 & \textbf{1.5894}  \\
Avg std MAE  & 0.2390 & \textbf{0.0843} & 0.0937 & 0.2447 & \textbf{0.0801} & 0.1355 & 0.0978 & 0.0611 & \textbf{0.0567} \\
\bottomrule
\end{tabular}}
\renewcommand{\tablename}{Extended data Table}
\caption{Comparison of different \molformer variants on QM9  test set, in terms of average MAE and average standard MAE. Variants considered are \molformer  pre-trained using QM9 only, PubChem only, and PubChem+ZINC dataset. The variants with and without fine-tuning on downstream task are compared, as well as models with, (\greencheck)Rotary, and without ,(\redcross)Rotary, rotary embeddings. Our best candidate variant (for Table \ref{tab:qm9_results_with_baseline}) is chosen based on the average MAE (Mean Absolute Error) score, lower is better.}
\label{tab:qm9_variants}
\end{table*}

 \begin{table}[ht!]
\centering
\begin{tabular}{lrr}
\hline
Correlation & ChemBERTa & \molformer-XL \\
\hline
 Fingerprint  & 0.48 & \textbf{0.64}\\
 MCS  & -0.44 & \textbf{-0.60}\\
 \hline
\end{tabular}
\renewcommand{\tablename}{Extended data Table}
\caption{ Correlation  with structural similarity metrics on 10000 \textcolor{black}{randomly selected} pairs of molecules from the PubChem dataset.  Reported correlations are between (1) the pairwise similarities estimated using molecular Fingerprints and those using \molformer-XL (or ChemBERTa) embeddings and (2) the number of atoms in the maximum common subgraph (MCS) of two molecules and their corresponding Euclidean distance in the embedding space.}
\label{tab:similarity}
\end{table}
\newpage
\appendix
\onecolumn

\begin{center}
   \large{\textbf{ Supplementary Information}}
\end{center}
Supplementary Figs. 1–6,  Discussion, and Tables 1–11.

\newpage
\appendix
\onecolumn
\begin{center}
   \large{\textbf{ Supplementary Information}}
   \end{center}
\section*{Related Work}

\paragraph{Large Scale Training of Language Model}
The recent advancement of transformer-based masked language models (MLMs) \cite{liu2019roberta, devlin-etal-2019-bert} and prefix language models (PLMs) \cite{2020t5} have shown remarkable performance on various natural language understanding tasks. Self-supervised pre-trained representation learning of sequences through MLMs randomly masks input tokens during training and  predicts these masked tokens, whereas PLMs require adding task-specific text tags to the input sequences. These language models show substantial performance improvements on downstream tasks via increasing transformer models size and pre-training using large-scale data corpora. Recent efforts have addressed the resulting cost and memory challenges encountered due to scaling up models and data. One such effort is the linear-time attention transformers introduced in \cite{FAVOR, Beltagy2020Longformer, kitaev2020reformer, wang2020linformer} which address the quadratic memory challenges within the attention mechanism and allowing for more efficiency in training MLMs.

\paragraph{Molecular Representation Learning}
To represent molecules in vector space, traditional chemical fingerprints such as ECFP \cite{rogers2010extended}, have been used. 
Deep neural nets were further trained on chemical fingerprints for supervised learning. Recurrent Neural Network (RNN) based models have been used for molecular representation learning using SMILES and other linear molecular annotations as inputs \cite{bjerrum2017smiles}. At the same time, graph convolutional networks have  been used to learn the neural fingerprints of molecules \cite{Duvenaudneurips15, coley2017convolutional}.
Previous work\cite{gilmer2017neural} implemented a single common framework to learn from graphs, referred to  as a message passing framework, which computes node embeddings by aggregating neighborhood information during the message passing phase and computes a feature vector of the graph during the readout phase. Many attempts to extend GNNs have been made, which include %
 variations of the original message passing concept to learn non-local effects; for instance,  in \cite{xiong2019pushing} an attention mechanism was introduced. One challenge faced by GNNs is achieving  higher expressivity that can distinguish between two given graphs to that of the hierarchy of the Weisfeiler-Lehman (WL) graph isomorphism, while maintaining scalability. It has been shown that typical message passing models have limited expressiveness and are not better than the first WL test (1-WL) \cite{morris2019weisfeiler}. Powerful deep models that represent 
higher order interactions between graph nodes have been suggested 
\cite{morris2019weisfeiler, maron2020provably}, but with a large increase in computational cost. 

Molecular graphs can be further augmented with the 3D coordinates of atoms. Such augmentation is considered as privileged information due to the cost associated with deriving the 3D molecular geometry. To better model the spatial interactions among atoms the message passing framework was extended in \cite{gilmer2017neural} to include pairwise interatomic distances as edge features when geometric information was available. 
More recently, variations of the message passing networks (MPNN) were proposed to better model the spatial interactions within molecules and increase the models expressive power,   e.g., by 
using continuous filter convolutional layers (SchNet) \cite{schnet} or by  
using directional message passing (DimeNet) \cite{klicpera2020directional} but  at the cost of increased computational complexity. However, those models are not generalizable to settings where 3D structural information is not readily available and/or is expensive to compute (e.g. for larger molecules). Since the goal of this work is to learn a generalizable molecular representation from a large amount of unlabeled data without relying on expensive 3D information, we mainly focus  on comparing the proposed \molformer  with existing  supervised and un/self-supervised  baselines that utilize different input representation (SMILES, graphs, fingerprints) and can be generalizable to a wide variety of tasks,  from quantum mechanical to physiological. 

\paragraph{Pre-trained Molecular Language and Graph Models} 

The recent success of language representation models in downstream NLP tasks has inspired  extending this paradigm to other domains.  By combining the power of  pre-training on large unlabeled corpus and contextual language models (LMs) using advanced neural nets, such as transformers,  a domain-specific ``language'' embedding is obtained 
as the exclusive input for several downstream tasks. 

Examples include understanding the language of life through advanced LMs trained on protein sequences.  Here features extracted by LMs directly from single protein sequences reach state-of-the-art performance in downstream prediction tasks, even when those were used without evolutionary information \cite{vig2020bertology, Rives622803, Elnaggar2020.07.12.199554}. Similar large-scale unsupervised pre-training on SMILES sequences have been explored   for molecular property prediction \cite{chithrananda2020chemberta, Xue2020.12.23.424259,smilesbert, kim2021merged, irwin2022chemformer}; however, those models did not attempt to predict a diverse range of molecular properties while exploiting the available chemical sequences at scale. Unsupervised/semi-supervised representation learning  has been tested on molecular graphs as well \cite{liu2019ngram, rong2020selfsupervised, wang2021molclr}. A more recent line of work has leveraged the power of contrastive self-supervised pre-training using 2D graph topology and 3D conformal geometry ~\cite{liu2021pre} (referred as GeomGCL), which showed performance improvement on molecular regression and classification tasks compared to prior pre-training baselines. 

\textcolor{black}{Previous work \cite{zhang2021motif} has further considered use of motif prediction during self-supervised pre\-training on molecular graphs. Another study \cite{wang2022improving} has also investigated the effect of including substructure information, as well as cheminformatics measures, in a molecular contrastive learning framework. References \cite{zhu2021dual, stark20223d} have shown the benefit of pre-training by maximizing information between different molecular views, e.g. 1D SMILES and 2D graph, or 2D graph and 3D geometry. Fang, et al. \cite{fang2022geometry} has also explored the advantages of geometry-based pre-training as well as including higher-order information in learning through modeling the atom–bond–angle relations in the graph (referred as GEM).}



\section{Model and Methods}\label{app:trainingdetails}
\subsection{\molformer Model and Pre-Training Details}\label{app:pretraining}
In this section we include additional details and insights of \molformer pre-training.  
\subsubsection{Optimizer}\label{app:optdetails}
For optimization we  used the Fused Lamb optimizer from \cite{you2019large} as implemented via APEX 
 due to Lamb's superior performance in several aspects of training. For example, learning rate warm ups were found to be unnecessary and training was found to be robust when large batch sizes were used. All other optimizers were unable to maintain their stability without large amounts of modification any time a training configuration needed to be changed.
\subsubsection{Linear Attention}\label{app:lineardetails}
Preliminary experiments showed an acceptable balance between computation speed and minimal performance deficit when compared to the FAVOR \cite{FAVOR} feature map.   Generalized Features are a simplification of the feature map in FAVOR \cite{FAVOR}. 
The feature map size we settled on is $32$.
\subsubsection{Rotary versus Absolute position embeddings}\label{app:embeddingdetails}
 We show in Figure \ref{fig:training_figapp} that \molformer with linear attention and rotary embeddings has a better validation loss than its absolute position counterpart. This observation lead us to adopt \molformer with linear attention and rotary embedding throughout the paper.  
\begin{figure*}[ht!]
 \setcounter{figure}{0}
\begin{subfigure}{\textwidth}
  \centering
  \includegraphics[width=.8\linewidth]{figures/PubChem.pdf} 
  \caption{PubChem Pre-training}
  \label{fig:training_pubchem}
\end{subfigure}\\
\begin{subfigure}{\textwidth}
  \centering
  \includegraphics[width=.8\linewidth]{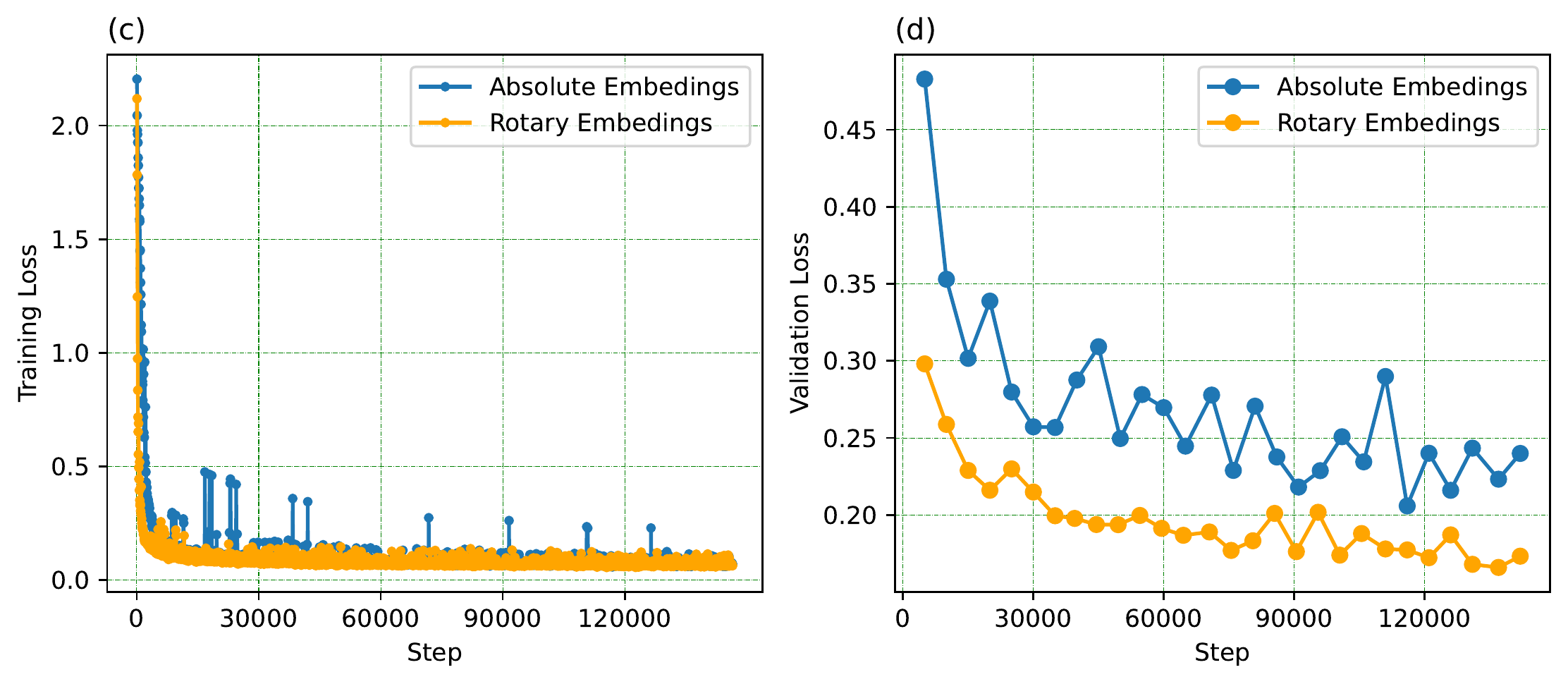}  
  \caption{PubChem+ZINC Pre-training}
  \label{fig:training_pubchem_zinc}
\end{subfigure}
\caption{Training and validation losses of our Linear attention \molformer with \emph{rotary} (relative) and absolute position embeddings on PubChem and PubChem+ZINC ($>$1 billion data points) datasets. We see that both rotary and absolute \molformer have graceful training curves. Our Rotary Linear attention  \molformer leads to lower training and validation losses than  \molformer with  absolute position embeddings. This observation lead us to focus on \molformer with rotary position embeddings. }
\label{fig:training_figapp}
\end{figure*}

\subsection{Parallelization and Computing Environment }
All experiments were performed on a GPU cluster where each node contains either $8$ NVIDIA Tesla V100 (32GB) or $8$ Ampere A100 (40GB) GPUs connected via NVLink. The V100 nodes are equipped with dual 28-core (Intel Xeon Gold 6258R) CPUs, the A100 nodes are equipped with dual 64-core (AMD EPYC 7742) CPUs, and all nodes are connected by 2 non-blocking EDR InfiniBand (100Gbps) network adapters as well as 2 100Gbps Ethernet adapters. All nodes are installed with RHEL 8.3, CUDA 10.2, and cuDNN 7.5. 

Due to the size of the datasets utilized in pre-training, our training environment relies on the Distributed Data Parallel functions provided by Pytorch 
and Pytorch Lightning utilizing the NCCL backend. By utilizing RDMA to enable GPU-direct technology we were able to efficiently scale to multi-node multi-GPU training. Additionally, we utilized HuggingFace Datasets 
to localize the data onto the machines where pre-training took place to improve performance during pre-training. Our pre-training task consists of training on the full dataset to 4 epochs. Training a single epoch of just PubChem on a single NVIDIA V100 GPU would take approximately $60$ hours. Utilizing Distributed Data Parallel, pre-training on the full PubChem dataset alone took approx. 22 hours on 16 NVIDIA V100 GPUs this averages to about 5.5 hours per epoch. The speed up achieved by parallelizing training to 16 GPUs gave us a factor of $10.9$. Pre-training for $4$ epochs on the combined PubChem+ZINC datasets took approx 208 hours on a 16 NVIDIA V100 GPUs which averages to about $52$ hours of compute for a single epoch. All fine-tunning tasks were able to be performed on single GPUs (either V100 or A100) and completed in approx. $12$ hours.

\subsection{ Memory Efficient Training with Adaptive Bucketing By Sequence Length } 
We observed that the distribution of molecule lengths in our dataset centered around molecules that were less than $45$ tokens long after tokenization. This fact coupled with large batch sizes increased the likelihood that padding tokens would overwhelm each batch and result in large amounts of computational waste. To address this problem we decided to break each minibatch into multiple buckets. This process is done on a batch by batch basis, i.e. on the fly, which means full dataset preprocessing does not take place. It should be noted that gathering of statistics for the full dataset did take place before training and buckets were defined by sequence interval length gathered from that process. The first bucket would contain SMILES strings of length $1$ to $42$, the next bucket would be of size $43$ to $66$, the third bucket would be of size $67$ to $122$ and finally the last bucket would be of size $123$ to $202$. Due to the length distribution of our dataset buckets $1$ and $2$ would always be present in all training steps while bucket $3$ would be present for the majority of minibatches. Molecules that fell into bucket $4$ appeared in most minibatches but would usually only represent a very small percentage of molecules found within the minibatch. With this information we decided to not utilize bucket $4$ in the training procedure until it reached a threshold of $50$ molecules thus preventing us from training on a bucket that consistently contained a very small amount of molecules which we believe aided training. Bucketing combined with gradient accumulation across buckets gave us stable training, maintained training randomization and reduced computational time needed compared to the traditional method of keeping GPU memory full at all times to maximize computation without consideration of the wasted computation that arises because of the large amount of padding tokens. To be more concrete, without bucketing on $1$ V$100$ GPU and on PubChem only a single epoch would take approx. $1200$ hours while with bucketing the same epoch only took around $60$ hours giving adaptive bucketing a speedup  of $20\times$. A similar concept, namely micro-batching \cite{falcon2019pytorch_}
exists but we  became aware of it after implementing  our adaptive  bucketing technique. We have not yet baselined the differences between our domain specific  bucketing implementation towards molecular data  against the generic micro-batching\cite{falcon2019pytorch_}
. Using Linear attention and bucketing allowed us to reduce the number of GPUs needed for quadratic attention and no bucketing from roughly 1000 to 16.

\subsection{Pre-training Scaleout}\label{app:scaleout}

\begin{figure}[ht!]
  \centering
  \includegraphics[width=.8\linewidth]{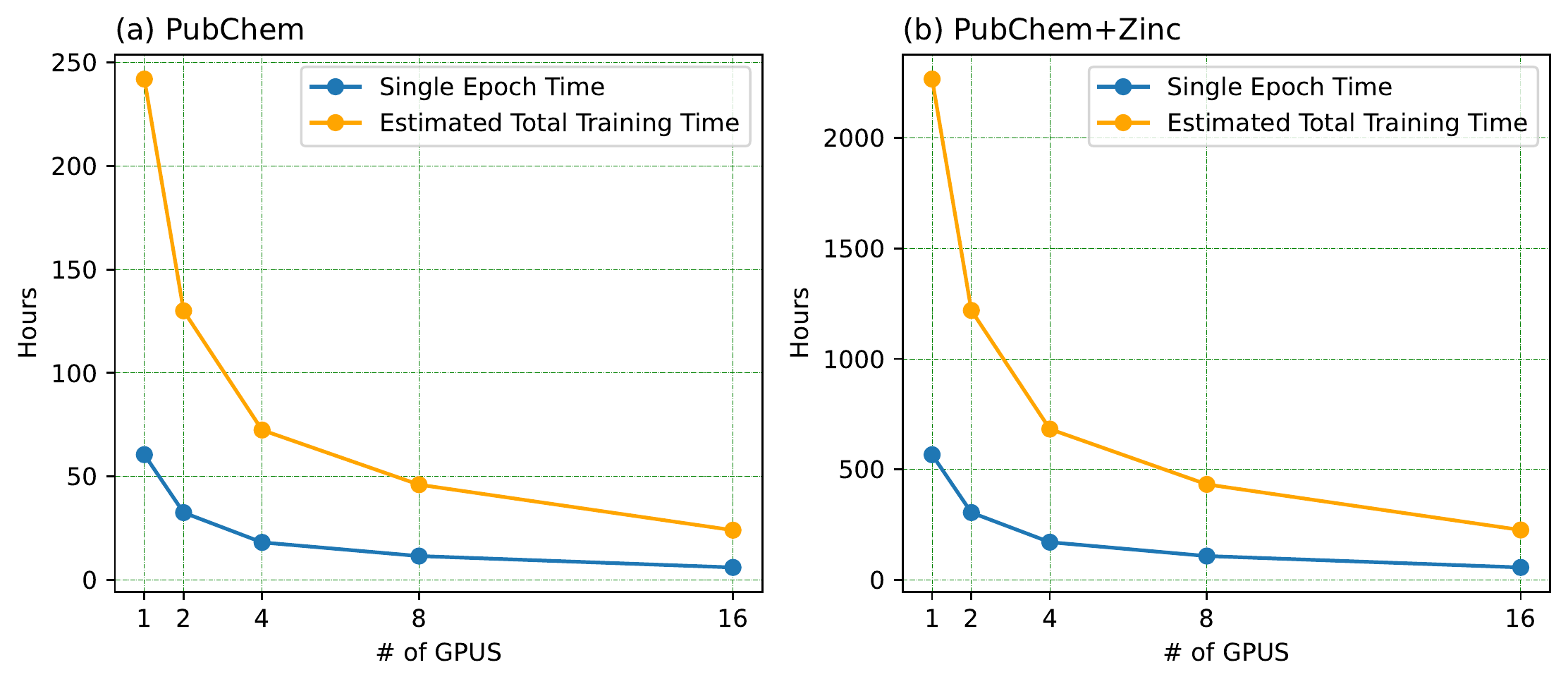}  
  \label{fig:pretrainingScaleout}
\caption{Estimated training times for our Linear attention \molformer with \emph{rotary} embeddings on a) PubChem and b) PubChem+ZINC datasets taken after 250 iterations. We see that training time decreases slightly sub-linearly as GPUs are added. Training time also scales approximately linearly as more data is added.}
\label{fig:pretrainingScaleout_fig}
\end{figure}

We give in Figure \ref{fig:pretrainingScaleout_fig} the estimated training times of \molformer as a function of the used GPUs in the parallelization.

\FloatBarrier

\section{Fine-tuning \molformer for Property Prediction }\label{app:finetunings params}

During the fine-tuning process, where the \molformer weights are not frozen, we experimented with different hyperparameters. In our experiments, we found that batch sizes of both 64 and 128 work best for the downstreams tasks. Also, among various learning rates, 3e-5 seems to be the best fit for all the measures on QM9 dataset. We found that the beta values used by the optimzier were import and were set to beta1 equaling $.9$ and beta2 equaling $.99$. We found the model to be very sensitive to the betas2 value during fine-tuning. The discriminator used was of a fixed size of $2$ layers with a hidden size of $768$ for all fine-tuning experiments.  

For the frozen strategy where the embeddings from the \molformer are fixed, we use a fully connected model to predict the properties. A hyperparameter sweep was performed for the frozen strategy using grid search and we randomly picked $25$ different variations for each task. The best model with the lowest validation loss was picked for further analysis. The different values of the frozen strategy hyperparameters are summarized in the table below.

\begin{table*}[ht!]
\setcounter{table}{0}
\caption{Different Values of the hyperparameters for the Frozen Models}

\centering
\begin{tabular}{l|l}
\toprule
Hyperparameter & Values\\
\midrule
Learning Rate & 0.001, 0.0001, 0.0005, 0.00005\\
Batch Size & 64, 128, 256\\
Hidden Dimension & 64, 128, 256, 512, 1024 \\
Number of Layers & 2, 3, 4\\
\bottomrule

\end{tabular}

\label{tab:hyperparams}
\end{table*}

\section{Dataset, Vocabulary, and Property Units}\label{app:dataset}
All our downstream evaluations are performed on tasks from the  MoleculeNet dataset~\cite{wu2018moleculenet}.  All the tasks mentioned in Table \ref{tab:esol_freesolv_result} use random splits as suggested in ~\cite{wu2018moleculenet}, while those in Table \ref{tab:molclr_comp} use scaffold splits as suggested in \cite{wang2021molclr}. A brief description of the downstream datasets are in Tables  ~\ref{tab:data_classification} and \ref{tab:data_regression}. We refer the reader to ~\cite{wu2018moleculenet} for more details on the specific tasks.

We have observed that various related work in the past have used different units for quantitative analysis on QM9 dataset without explicitly stating the units, making it difficult to compare the relative performance of different methods. We have listed the units of the measures used in this paper in Table \ref{tab:measure_unit}. 
We also give in Tables \ref{tab:data_stats} and \ref{tab:vocab_stats} the statistics of sequence length and vocabulary for the datasets considered in this work. While the vocabulary of each dataset varies the architecture of our models is identical for all experiments contained in this paper. The vocabulary for our models is defined by the union of the vocabularies of both PubChem and Zinc dataset. 

\begin{table*}[ht!]

\centering
\begin{tabular}{lcl}
\toprule
        & \# of compounds & Description                                                  \\
\midrule
BBBP    & 2039            & Blood brain barrier penetration dataset                      \\
Tox21   & 7831            & Toxicity measurements on 12 different targets                \\
Clintox & 1478            & Clinical trial toxicity of drugs                             \\
HIV     & 41127           & Ability of small molecules to inhibit HIV replication        \\
BACE    & 1513            & Binding results for a set of inhibitors for $\beta$ -- secretase 1 \\
SIDER   & 1427            & Drug side effect on different organ classes  \\         
\bottomrule
\end{tabular}
\caption{Description of classification datasets used for downstream evaluations}
\label{tab:data_classification}
\end{table*}

\begin{table*}[ht!]

\centering
\begin{tabular}{lcl}
\toprule
       & \# of compounds & Description                                                 \\
\midrule
QM9           & 133885 & 12 quantum mechanical calculations of small organic molecules with upto nine heavy atoms           \\
QM8           & 21786  & 12 excited state properties of small molecules  \\
ESOL          & 1128   & Water solubility dataset                                                                                                      \\
FreeSolv      & 642    & Hydration free energy of small molecules in water                                                                             \\
Lipophilicity & 4200   & Octanol/water distribution coefficient of molecules   \\
\bottomrule
\end{tabular}
\caption{Minimum, maximum and mean and standard deviation of sequence length for the datasets considered in this work. The vocabulary size after tokenization is also given in the last column. }
\label{tab:data_regression}
\end{table*}

\begin{table*}[ht!]
\centering
\begin{tabular}{lccccc}
Pre-trained Data & Min & Max & Mean & Std & Vocab Size\\
\midrule
QM9 & 1 & 22 & 14.76 & 2.02 & 30\\
Zinc &4 & 152 & 43.08 & 8.70 & 113 \\
10 PubChem + 10 Zinc & 2 & 2031 & 42.52 & 10.06 & 1044 \\
PubChem & 1 & 2211 & 43.24 & 22.21 & 2349\\
100 PubChem + 10 Zinc & 2 & 2211 & 42.86 & 16.93 & 2355 \\
PubChem + Zinc & 1 & 2211 & 44.76 & 14.55& 2362
\end{tabular}
\caption{Description of regression datasets used for downstream evaluations} 
\label{tab:data_stats}
\end{table*}

\begin{table*}[ht!]
\centering
\begin{tabular}{ll}
Pre-trained Data & Most Frequent Tokens\\
\midrule
QM9 & \texttt{C, 1, 0, 2, (} \\
PubChem &  \texttt{C, =, (, ), 0} \\
Zinc & \texttt{C, c, (, ), 1}\\
PubChem + Zinc & \texttt{C, c, (, ), =}\\
\bottomrule
\end{tabular}
\caption{Top five most frequent tokens for each data source.}

\label{tab:vocab_stats}
\end{table*}
\begin{table*}[ht!]
\centering
\begin{tabular}{lc}
Measure & Unit\\
\midrule
$\alpha$ & Bohr$^3$ \\
$C_v$ & cal/(mol*K)\\
G & Hartree\\
gap & Hartree\\
H & Hartree\\
$\epsilon_{homo}$ & Hartree \\
$\epsilon_{lumo}$  & Hartree\\
$\mu$ & Debye \\ 
$\langle R^2\rangle$ & Bohr$^2$\\
$U_0$ & Hartree \\
U & Hartree\\
ZPVE & Hartree\\
$*\_atom$ & eV \\ \bottomrule
\end{tabular}
\caption{Units of QM9 target measures}
\label{tab:measure_unit}
\end{table*}

\newpage
\section{Additional results on comparing \molformer-XL with geometry-aware GNNs on regression benchmarks}
In this section we show additional results in Table \ref{tab:esol_freesolv_result_3D} on comparison of \molformer with geometry-aware supervised (DimeNet) and self-supervised (GeomGCL, GEM) on the physical chemistry regression benchmarks from MoleculeNet.

\begin{table*}[ht]
\centering
\begin{tabular}{lcccc}
\toprule
\textcolor{black}{Task}  & \textcolor{black}{DimeNet~\cite{klicpera2020directional}} & \textcolor{black}{GeomGCL~\cite{liu2021pre}} & \textcolor{black}{GEM~\cite{fang2022geometry}}  & \textcolor{black}{\molformer-XL}  \\
\midrule
\textcolor{black}{ESOL (RMSE)} & \textcolor{black}{0.633}  & \textcolor{black}{0.575} & \textcolor{black}{0.798}&  \textcolor{black}{\textbf{0.2787}} \\
\textcolor{black}{FreeSolv (RMSE)} & \textcolor{black}{0.978} & \textcolor{black}{0.866} & \textcolor{black}{1.877} &
\textcolor{black}{\textbf{0.2308}} \\
\textcolor{black}{Lipophilicity (RMSE)} & \textcolor{black}{0.614} & \textcolor{black}{0.541} & \textcolor{black}{0.660}
  & \textcolor{black}{\textbf{0.5289}} \\
\bottomrule
\end{tabular}
\caption{\textcolor{black}{Performance of fine-tuned \molformer-XL and other supervised and self-supervised geometry-aware GNN baselines  on  ESOL, FreeSolv, and Lipophilicity regression benchmarks. Baseline performances are taken from references \protect{\cite{liu2021pre, fang2022geometry}}}. }
\label{tab:esol_freesolv_result_3D}
\end{table*}

\section{Additional Results and Ablations on QM9 Benchmark}
In this section we show additional results and  ablations on the pre-training and fine-tuning of \molformer on the QM9 benchmark. 
\paragraph{Fine-tuned \molformer-XL versus Baselines}
We further report \molformer-XL performance on all twelve property prediction  tasks individually within QM9, and compare that against several previously discussed baseline models as well as four additional baselines. The additional baselines included are as follows: (i) a more expressive GNN, specifically 123-GNN  \cite{morris2019weisfeiler}, (ii) two  neural nets that leverage 3D geometry -- a multitask neural net encoding the Coulomb Matrix (CM) \cite{PhysRevLett.108.058301} and  its GNN variant as in the deep tensor neural net (DTNN) \cite{schutt2017quantum}, and (iii) Chemberta \cite{chithrananda2020chemberta}.  
 which show that \molformer-XL achieves comparable or better performance to that of the majority of the  competitors. Specifically, \molformer-XL outperforms all baselines in term of average MAE and average standard MAE. ChemBERTa shows the highest average MAE among all. 
The more expressive 123-GNN performs better on most measures compared to \molformer-XL;  however, such powerful networks are known to be  difficult to scale  (see \cite{vignac2020building} for example).  As a comparison, the linear attention employed in  \molformer-XL ensures  a linear time complexity.


\begin{table*}[ht!]
\centering
\begin{tabular}{lrrrrrrrr}
\toprule
 & \multicolumn{3}{c|}{Graph-Based} & \multicolumn{3}{c|}{Geometry-Based} &  SMILES-Based\\
\midrule
Measure  & A-FP  & 123-gnn & \multicolumn{1}{l|}{GC}  & CM & DTNN & \multicolumn{1}{l|}{MPNN}  &  \molformer-XL & ChemBERTa \\
\midrule
$\alpha$           & 0.492  &  \rankone{0.27}  &  1.37  &   0.85    &  0.95  &  0.89 & \ranktwo{0.3327} & 0.8510 \\
$C_v$                 & 0.252  &  \rankone{0.0944}  &  0.65 & 0.39    &  0.27  &  0.42  &     \ranktwo{0.1447} & 0.4234 \\
G                  & 0.893   &  \rankone{0.0469}  &  3.41  & 2.27    &  2.43  &  2.02  &    \ranktwo{0.3362} & 4.1295 \\
gap                 & 0.00528   &  \ranktwo{0.0048}  &  0.01126  &  0.0086  &  0.0112  &  0.0066  &   \rankone{0.0038} & 0.0052\\
H                     & 0.893   &  \rankone{0.0419}  &  3.41  &     2.27    &  2.43  &  2.02  & \ranktwo{0.2522} & 4.0853 \\
$\epsilon_{homo}$    & 0.00358   &  \ranktwo{0.00337}  &  0.00716  &   0.00506 &  0.0038  &  0.00541  &   \rankone{0.0029} & 0.0044 \\
$\epsilon_{lumo}$     & 0.00415   &  \ranktwo{0.00351}  &  0.00921  & 0.00645 &   0.0051  &  0.00623  &    \rankone{0.0027} & 0.0041\\
$\mu$                 & 0.451  &  0.476  &  0.583  &    0.519   &  \rankone{0.244}  &  \ranktwo{0.358}  & 0.3616 & 0.4659 \\
$\langle R^2\rangle$   & 26.839  &  22.90  &  35.97   & 46.00     &  \rankone{17.00}  &  28.5  &     \ranktwo{17.0620} & 86.150 \\
$U_0$                 & 0.898   &  \rankone{0.0427}  &  3.41  & 2.27    &  2.43  &  2.05  &    \ranktwo{0.3211} & 3.9811\\
U                     & 0.893   &  \rankone{0.111}  &  3.41  &  2.27    &  2.43  &  2.00  &   \ranktwo{0.2522} & 4.3768 \\
ZPVE                 & 0.00207  &  \rankone{0.00019}  &  0.00299  &  0.00207 &  0.0017  &  0.00216  &     \ranktwo{0.0003} & 0.0023 \\
\midrule
Avg MAE 	& 2.6355 &	\ranktwo{1.9995} &	4.3536 & 4.7384 &	2.3504	& 3.1898 & \rankone{1.5894} & 8.7067 \\
Avg std MAE  &0.0854 & \ranktwo{0.0658} & 0.1683 & 0.1281 & 0.1008 & 0.1108 & \rankone{0.0567} & 0.1413 \\
\bottomrule
\end{tabular}
\caption{\molformer performance on QM9 test set. Our best \molformer variant is pre-trained on PubChem+ZINC dataset and fine-tuned for each measure. Baseline performance values are taken from \protect{\cite{wu2018moleculenet, xiong2019pushing, maron2020provably}.}   \rankone{Blue} and \ranktwo{Orange} indicates best and second-best performing model, respectively. 
\molformer trained with rotary embeddings on PubChem+ZINC achieves the best Avg MAE and Avg. std. MAE across all tasks. 
}
\label{tab:qm9_results_with_baseline}
\end{table*}

\begin{table*}[ht!]
\centering
\begin{tabular}{lccc}
\hline \textcolor{black}{QM9 Task} & \textcolor{black}{SchNet \cite{schnet}} & \textcolor{black}{DimeNet \cite{klicpera2020directional}} & \textcolor{black}{\molformer-XL} \\
\hline 
\textcolor{black}{$U_0$\_atom}  & \textcolor{black}{0.0140} & \textcolor{black}{\textbf{0.0080}}  & \textcolor{black}{0.0827}\\
\textcolor{black}{U\_atom}  & \textcolor{black}{0.0190} & \textcolor{black}{\textbf{0.0079}} & \textcolor{black}{0.0974}\\
\textcolor{black}{H\_atom}  & \textcolor{black}{0.0140} & \textcolor{black}{\textbf{0.0081}} & \textcolor{black}{0.0947}\\
\textcolor{black}{G\_atom}  & \textcolor{black}{0.0140} & \textcolor{black}{\textbf{0.0089}} & \textcolor{black}{0.0888}\\
\hline
\end{tabular}
\caption{\textcolor{black}{Comparison of \molformer-XL  with two exemplary 3D GNN models, SchNet and Dimenet, on QM9 atomization energy/enthalpy (in eV) regression benchmarks.  
}}
\label{tab:molformer_3dgnn}
\end{table*}

\begin{figure*}[ht!]
\centering
 n
  \includegraphics[scale=0.6]{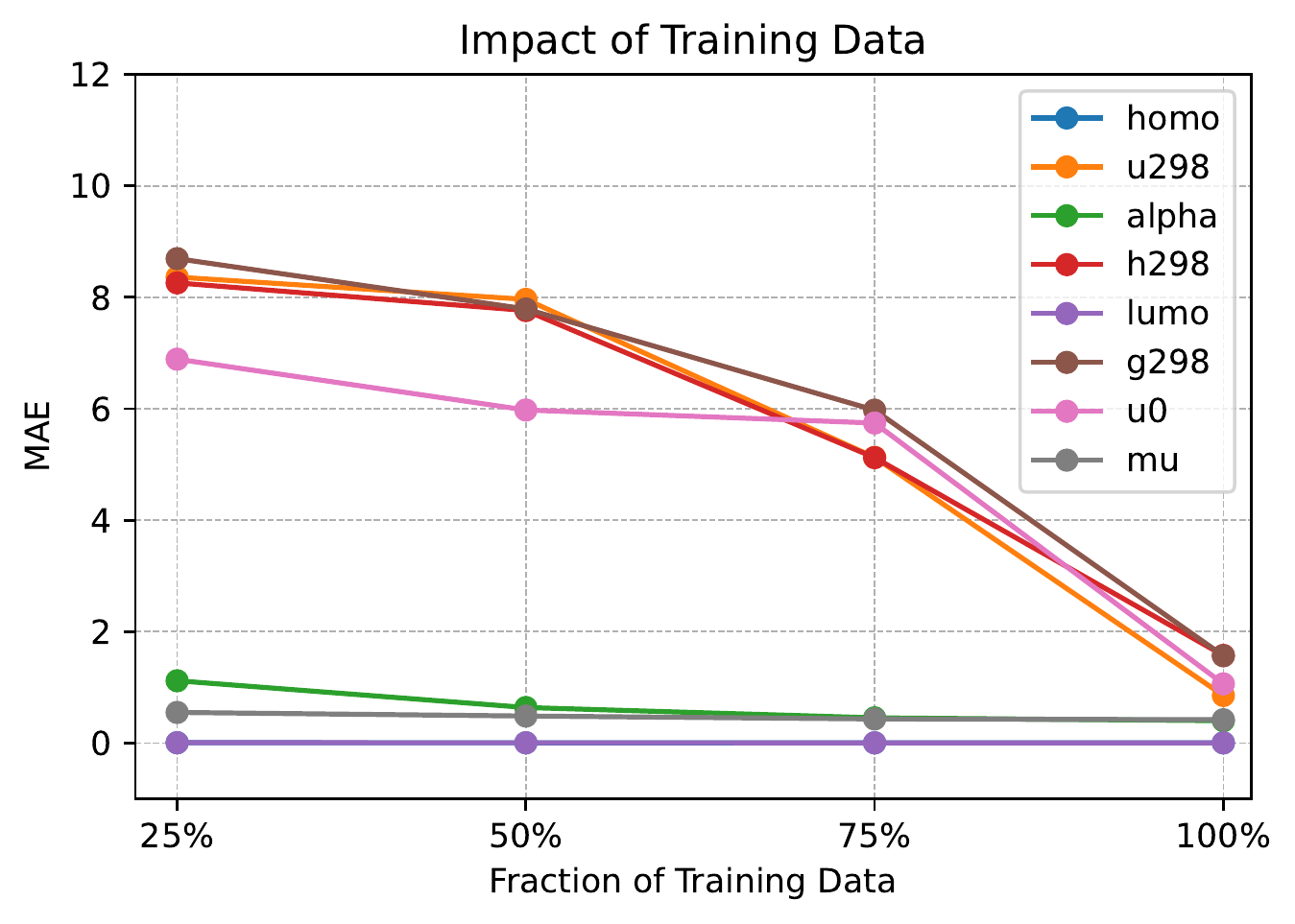}

\caption{Mean absolute errors  with  varying training set size. Fine-tuning of \molformer with rotary embeddings for prediction of various properties on QM9 Molecules for different training set sizes.}
\label{fig:training_fig_Qm9}
\end{figure*}

\paragraph{Impact of \molformer pre-training dataset on Downstream task} We present in Table \ref{tab:newTable_newResults_all}  and Fig. \ref{fig:training_fig_Qm9}  ablations on the impact of the pre-training dataset on the performance of fine-tuning \molformer in the downstream property prediction tasks on QM9. We see that as the pre-training dataset sizes becomes large, \molformer achieves better performance (i.e lower MAE). Note that \molformer-XL refers to \molformer pre-trained on PubChem+ ZINC.

\paragraph{Position embeddings} The positional embeddings ablation results are collected in Table \ref{tab:qm9_variants}.  Different pre-training datasets sizes are also investigated and are broken up into \molformer (1) pre-trained on only the QM9 training set (111k molecules) -- referred to as \molformer-qm9 (2) only PubChem (111M molecules) -- referred to as \molformer-PubChem   (3) PubChem+ZINC (1.1 Billion+ Molecules), \textit{i.e.} \molformer-XL. 
Results  presented in Table \ref{tab:qm9_variants} show that \molformer with Rotary embeddings and fine-tuning is behind the Absolute positional embedding model for the smaller datasets, but then wins as the dataset size passes 1 Billion molecules.
\textcolor{black}{ We note that two main differences  between our linear time relative rotary attention  formulation and the original one from \cite{su2021roformer} are: 1) our attention remains normalized, whereas the one proposed in\cite{su2021roformer} does not; 2) The  rotation in \cite{su2021roformer} is applied to  the transformed query and key in the random feature space, whereas we apply it at the key and query level. This results, in our case, to an approximation  of a kernel acting on the position-modulated key and queries that encode the relative position. On the other hand, the original formulation results in an approximation of a product kernel one on the space of positions and one on the space of keys and queries. We believe that both the normalization and having a kernel acting on a joint embedding of keys/queries and positions advantage our formulation on the original one from \cite{su2021roformer}. More rigorous investigation will be carried out in future work.       }

  With that said we can see that as the pre-training grows from PubChem only to the more extensive and diverse PubChem+Zinc corpus the representational power of the model increases, as showcased by the stronger performance on the QM9 benchmark. 
  
  \paragraph{Impact Fine-tuning versus Frozen Embedding/ Rotary versus Absolute on Downstream tasks } We show in Table
\ref{tab:std_results_pubchem_zinc} the impact of fine-tuning versus using frozen \molformer embeddings from pre-training phase on the performance on the QM9 benchmark in both rotary and absolute embeddings. We see that fine-tuning and rotary achieve the best performance.
 
 \paragraph{Robustness Across Data Folds} We  report performance comparison at the individual property prediction task within QM9   in SI Table \ref{tab:newTable_newResults_all}.
In order to ensure the robustness of these results across data splits we also provide,  the performance of \molformer-XL  on QM9 tasks using 5-cross validation folds  (SI Table \ref{tab:std_results_pubchem_zinc}).



\begin{table*}
\resizebox{\textwidth}{!}{\begin{tabular}{lrrr|rrr|rrr}
\toprule
Pre-training Data $\rightarrow$ &\multicolumn{3}{c|}{QM9 Only}& \multicolumn{3}{c|}{PubChem Only} & \multicolumn{3}{c}{PubChem+ZINC}\\
Measure $\downarrow$ & \begin{tabular}{@{}c@{}}Frozen \\ \redcross Rotary\end{tabular} & \begin{tabular}{@{}c@{}}Fine-tuned \\ \redcross Rotary\end{tabular} & \begin{tabular}{@{}c@{}}Fine-tuned \\ \greencheck Rotary\end{tabular} & \begin{tabular}{@{}c@{}}Frozen \\ \redcross Rotary\end{tabular} & \begin{tabular}{@{}c@{}}Fine-tuned \\ \redcross Rotary\end{tabular} & \begin{tabular}{@{}c@{}}Fine-tuned \\ \greencheck Rotary\end{tabular} & \begin{tabular}{@{}c@{}}Frozen \\ \redcross Rotary\end{tabular} & \begin{tabular}{@{}c@{}}Fine-tuned \\ \redcross Rotary\end{tabular} & \begin{tabular}{@{}c@{}}Fine-tuned \\ \greencheck Rotary\end{tabular}\\
\midrule
$\alpha$                & 1.6258 & \textbf{0.5078} & 0.6001        & 1.5470 & \textbf{0.5280} & 0.8452  &  0.5312 & 0.3713 & \textbf{0.3327} \\
$C_v$                   & 1.0176 & \textbf{0.1589} & 0.1906    & 0.9984 & \textbf{0.1506} & 0.2701 &  0.2303 & 0.1584 & \textbf{0.1447} \\
G                       & 3.2528 & 0.9985 & \textbf{0.7479}        & 2.0089 & \textbf{0.8626} & 1.5920  &  \textbf{0.3066} & 0.6861        & 0.3362 \\
gap                     & 0.0187 & \textbf{0.0057} & 0.0061    & 0.0182 & \textbf{0.0050} & 0.0109  &  \textbf{0.0036} & 0.0039 & 0.0038 \\
H                       & 1.9221 & 1.1579 & \textbf{1.0250}    & 2.3627 & 1.3342 & \textbf{0.7088}  & 0.3675 & .07369 & \textbf{0.2522} \\
$\epsilon_{homo}$       & 0.0115 & \textbf{0.0042} & 0.0046      & 0.0147 & \textbf{0.0038} & 0.0082 &  0.0062 & \textbf{0.0028}  & 0.0029 \\
$\epsilon_{lumo}$       & 0.0157 & \textbf{0.0041} & 0.0056        & 0.0148 & \textbf{0.0036} & 0.0080  &  0.0058 & \textbf{0.0025}        & 0.0027 \\
$\mu$                   & 0.8394 & \textbf{0.4380} & 0.4630    & 0.8509 & \textbf{0.4284} & 0.6166 &  0.6463 & 0.3921 & \textbf{0.3616} \\
$\langle R^2\rangle$    & 86.9461 & \textbf{24.0785} & 25.9482  & 87.2816 & \textbf{30.2904} & 34.0425        & 27.5962 & 18.8286 & \textbf{17.0620} \\
$U_0$                   & 3.0626 & \textbf{1.1462} & 1.3168    & 2.0613 & \textbf{0.8969} & 1.5503  &  0.4500 & 0.4244 & \textbf{0.3211} \\
U                       & 1.8555 & \textbf{1.0454} & 1.6158        & 1.9638 & \textbf{1.1122} & 1.1351   &  0.4480 &        0.7370 & \textbf{0.2522} \\
ZPVE                    & 0.0020 & \textbf{0.0011} & 0.0012    & 0.0020 & \textbf{0.0008} & 0.0012  &   0.0004 & \textbf{0.0002}& 0.0003 \\
\midrule
Avg MAE & 8.3808 & \textbf{2.4621} & 2.6604      & 8.260 & \textbf{2.968} & 3.3990 & 2.5497 & 1.8620 & \textbf{1.5894} \\
Avg std MAE & 0.2390 & 0.0843 & 0.0937 & 0.2447 & 0.0801 & 0.1355  & 0.0978 & 0.0611 & \textbf{0.0567} \\
\# Wins for fixed data & \multicolumn{1}{c}{0} & \multicolumn{1}{c}{\textbf{10}} & \multicolumn{1}{c|}{2} & \multicolumn{1}{c}{0} & \multicolumn{1}{c}{\textbf{11}} & \multicolumn{1}{c|}{1} & \multicolumn{1}{c}{2} & \multicolumn{1}{c}{3} & \multicolumn{1}{c}{\textbf{7}} \\
\bottomrule
\end{tabular}}
\caption{Comparison of different \molformer variants on the QM9  test set. Models on the left half of the table are pre-trained using QM9 only and the model in the middle is trained on PubChem only, whereas the models on right half are pre-trained on the PubChem+ZINC dataset. The variants with (\greencheck) and without (\redcross) rotary embeddings are compared. Our best candidate variant (for Table \ref{tab:qm9_results_with_baseline}) is picked based on the average MAE score.}
\label{tab:newTable_newResults_all}
\end{table*}

\begin{table*}[htb!]
\centering
 \begin{tabular}{lrrrr}
 \toprule
  & \multicolumn{3}{c}{PubChem+Zinc}\\
 \midrule
 Measure $\downarrow$ & \begin{tabular}{@{}c@{}}Frozen \\ \redcross Rotary\end{tabular} & \begin{tabular}{@{}c@{}}Frozen \\ \greencheck Rotary\end{tabular} & \begin{tabular}{@{}c@{}}Fine-tuned \\ \redcross Rotary\end{tabular} & \begin{tabular}{@{}c@{}}Fine-tuned \\ \greencheck Rotary\end{tabular} \\
 \midrule
 $\alpha$ &  0.6468 $\pm$ 0.0169  & 1.2956 $\pm$ 0.0244 & 0.4246  $\pm$ 0.0486 & \textbf{0.3455 $\pm$ 0.0066} \\
 $C_v$ &  0.2825 $\pm$ 0.0078 &  0.7959 $\pm$ 0.0141 &  0.2080 $\pm$ 0.0340 &  \textbf{0.1589 $\pm$ 0.0102} \\
 G &    1.2865 $\pm$ 0.1192 &  2.1535 $\pm$ 0.0503 &  0.7696 $\pm$ 0.0719 &  \textbf{0.3609 $\pm$ 0.0276} \\
 gap &  0.0089 $\pm$ 0.0000 &  0.0165 $\pm$ 0.0002 & \textbf{ 0.0038 $\pm$ 0.0001} &  0.0040 $\pm$ 0.0001 \\
 H &  1.0730 $\pm$ 0.3188 &  1.8490 $\pm$ 0.1469 &  0.8525 $\pm$ 0.0855 &  \textbf{0.3318 $\pm$ 0.0512} \\ 
 $\epsilon_{homo}$ &  0.0066 $\pm$ 0.0000 &  0.0103 $\pm$ 0.0001 &  0.0030 $\pm$ 0.0001 &  \textbf{0.0029 $\pm$ 0.0001} \\
 $\epsilon_{lumo}$ &  0.0063 $\pm$ 0.0000 &  0.0128 $\pm$ 0.0001 &  0.0039 $\pm$ 0.0002 &  \textbf{0.0029 $\pm$ 0.0001} \\
 $\mu$ &  0.6872 $\pm$ 0.0049 &  0.8089 $\pm$ 0.0205 &  0.3986 $\pm$ 0.0060 &  \textbf{0.3709 $\pm$ 0.0065} \\
 $\langle R^2\rangle$ &  29.3779 $\pm$ 0.2944 &  72.8752 $\pm$ 1.1180 &  19.9005 $\pm$ 0.3305 &  \textbf{17.8121 $\pm$ 0.8596} \\
 $U_0$ &  1.2877 $\pm$ 0.2373 &  2.1030 $\pm$ 0.0967 &  0.8492 $\pm$ 0.1102 &  \textbf{0.3795 $\pm$ 0.0820} \\
 U &  1.3238 $\pm$ 0.1954 &  1.8647 $\pm$ 0.0601 &  0.8631 $\pm$ 0.0999 &  \textbf{0.3677 $\pm$ 0.0382} \\
 ZPVE &  0.0005 $\pm$ 0.0000 &  0.0018 $\pm$ 0.0001 &  \textbf{0.0003 $\pm$ 0.0001} &  \textbf{0.0003 $\pm$ 0.0001} \\
 \bottomrule
 \end{tabular}
\caption{Mean and standard deviation of Mean Absolute Error (MAE) on the various QM9 tasks using a pre-trained model that is trained on both PubChem and Zinc}
\label{tab:std_results_pubchem_zinc}
\end{table*}

Specifically, we  report from Table \ref{tab:std_results_pubchem_zinc} the mean and  standard deviations of MAE for 5 different  folds of the data split into $80 \%$ training and $10 \%$ validation and $10\%$ test. Most of the related work does not perform cross validation and just report the results on a single split. We note that the standard deviations are quite low for most of the predictions and the mean errors are in line with the main paper for all folds  of all tasks which suggests the \molformer representations are robust.  

\newpage
\section{Insights into \molformer - tSNE Visualization}
\begin{figure}[ht!]
    \centering
    \includegraphics[width=1\textwidth]{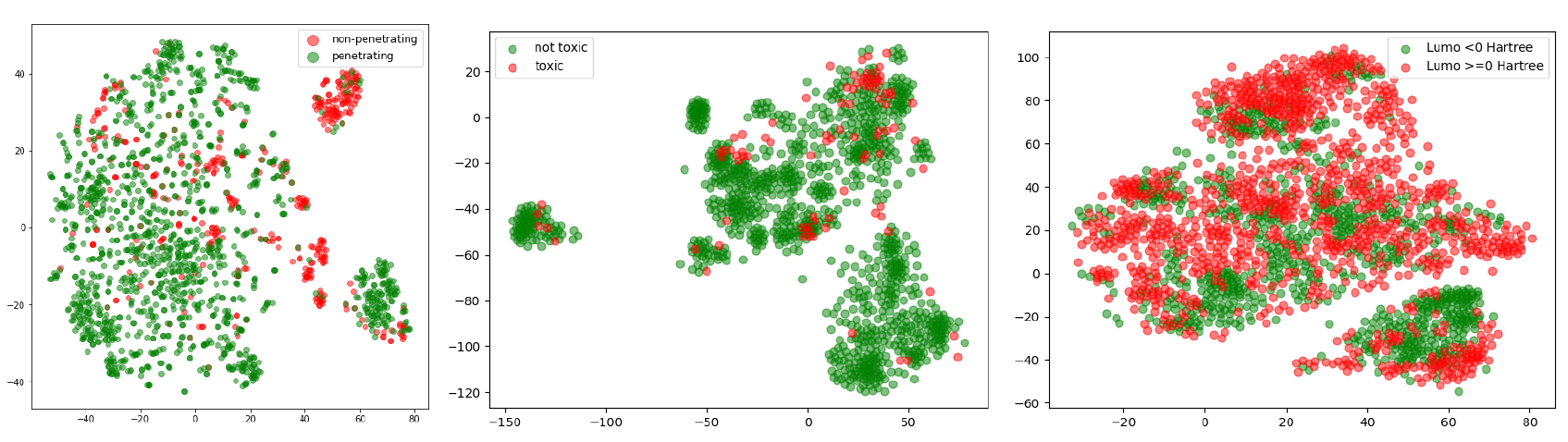}
    \caption{t-SNE projection  of frozen \molformer embeddings (no fine-tuning) of the BBBP (left) and ClinTox(middle) and discretized LUMO (right) datasets.} 
    \label{fig:tsne}
\end{figure}

We performed the following set of experiments in order to evaluate if the \molformer embeddings  capture molecular properties as well as structural aspects. First, we investigate a t-SNE \cite{vandermaaten08a} projection  of \molformer-XL embeddings sampled from two classes of the BBBP dataset \textcolor{black}{and two classes of the ClinTox dataset as shown in Figure \ref{fig:tsne}: left and middle. The two classes in BBBP are separated by molecules that penetrate (penetrating) the blood brain barrier and molecules that do not penetrate (non-penetrating) the blood brain barrier. Clintox is separated into a class of toxic molecules and  non-toxic molecules. Finally we discretize the QM9  dataset according to the LUMO energy, where one class consists  of molecules with LUMO energy value $<0$ Hartree and the other class is composed of molecules with  LUMO energy  $>=0$ Hartree (Figure \ref{fig:tsne}: right). }

\textcolor{black}{It can be seen in Figure \ref{fig:tsne}, starting with BBBP on the very left, that even without task-specific fine-tuning, \molformer-XL is able to discriminate between the two classes, suggesting that molecular property information has been captured in this universal representation.   For ClinTox, middle plot in Figure \ref{fig:tsne},  clustering of toxic and non-toxic is apparent but the clusters extend beyond the toxic and non-toxic classes. With Clintox one can see that as the center of the clusters becomes more concentrated, the  toxic and non-toxic molecules there is less overlap of classes in that area of the cluster. Also, generally, if a toxic molecule is located near a cluster of non-toxic molecules, it is on the border of the cluster. Figure \ref{fig:tsne}:right  shows the a separation tendency between the high/low LUMO classes but with an obvious amount of overlap in the non-finetuned embedding space.}

\section{\molformer Attention Visualization  and Structure Discovery}
\begin{figure}
  \begin{subfigure}{\linewidth}
  \includegraphics[width=.15\linewidth]{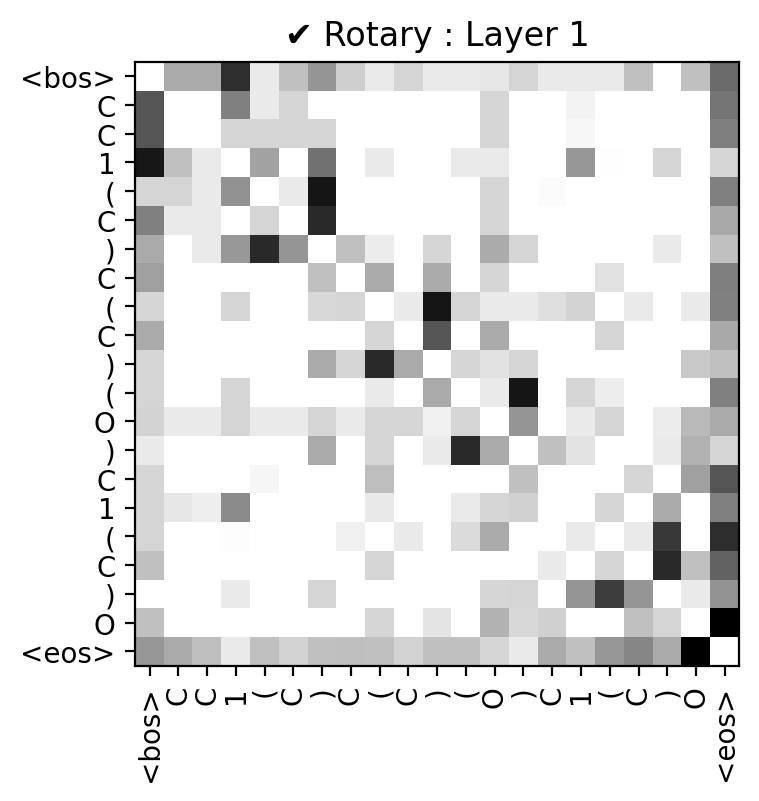}\hfill
  \includegraphics[width=.15\linewidth]{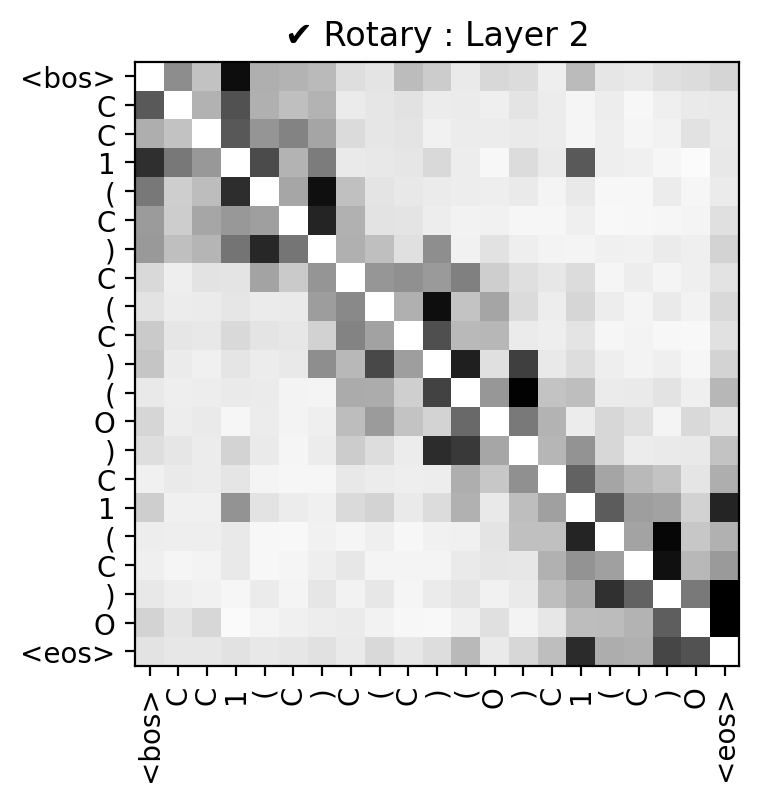}\hfill
  \includegraphics[width=.15\linewidth]{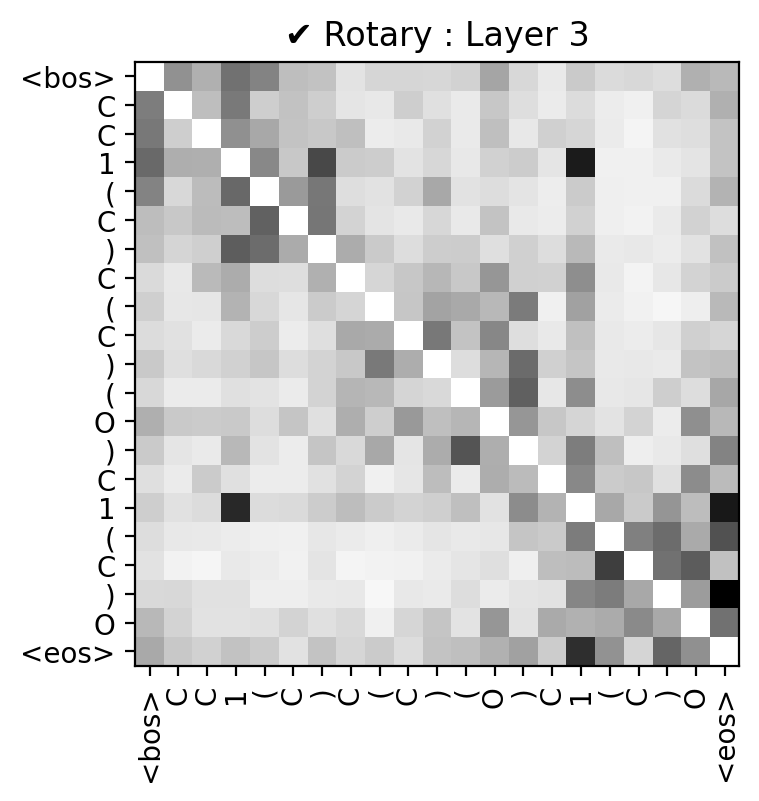}\hfill
  \includegraphics[width=.15\linewidth]{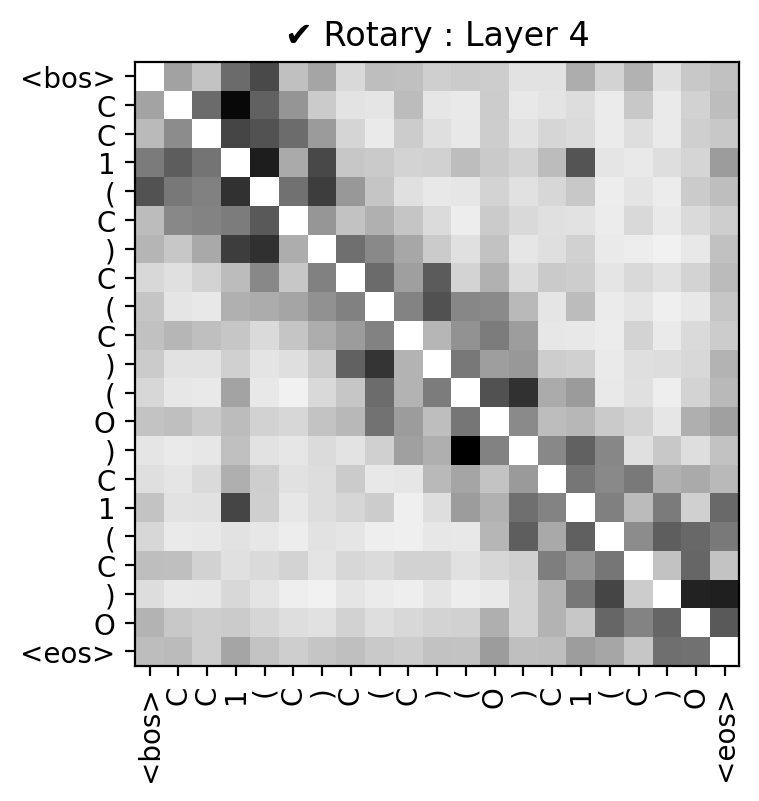}\hfill
  \includegraphics[width=.15\linewidth]{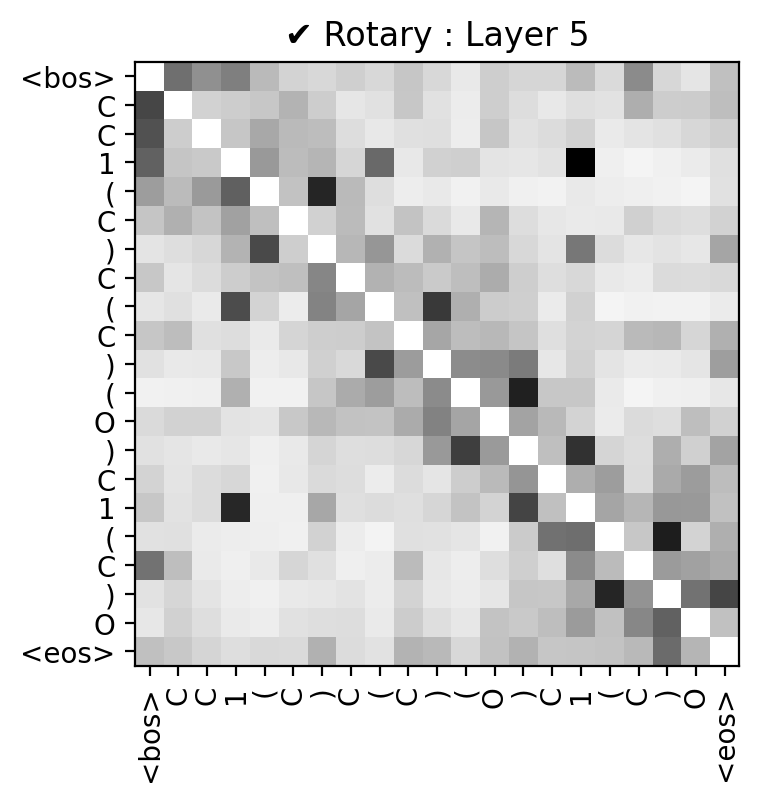}\hfill
  \includegraphics[width=.15\linewidth]{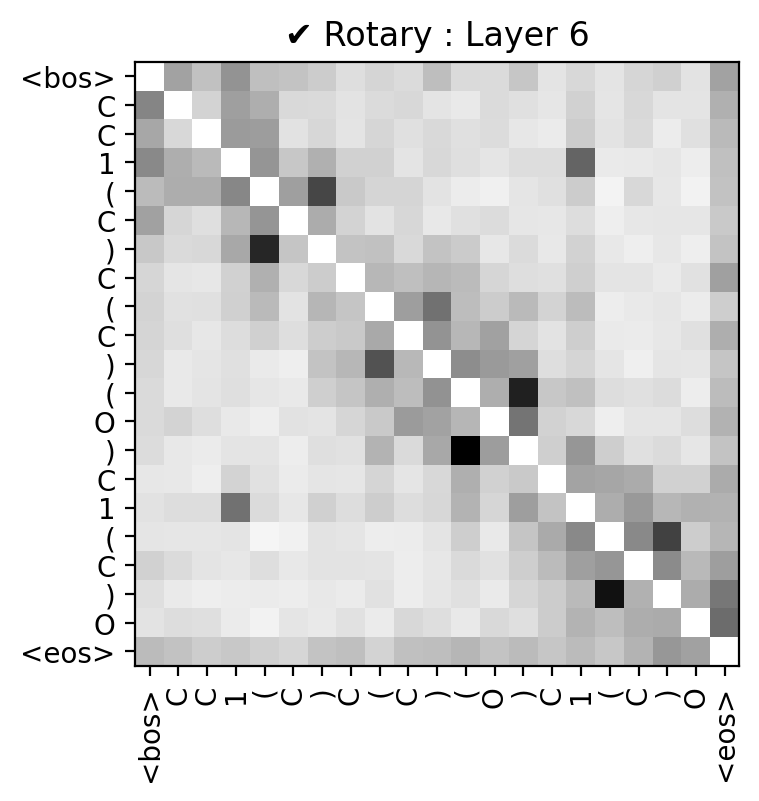}\hfill
  \includegraphics[width=.15\linewidth]{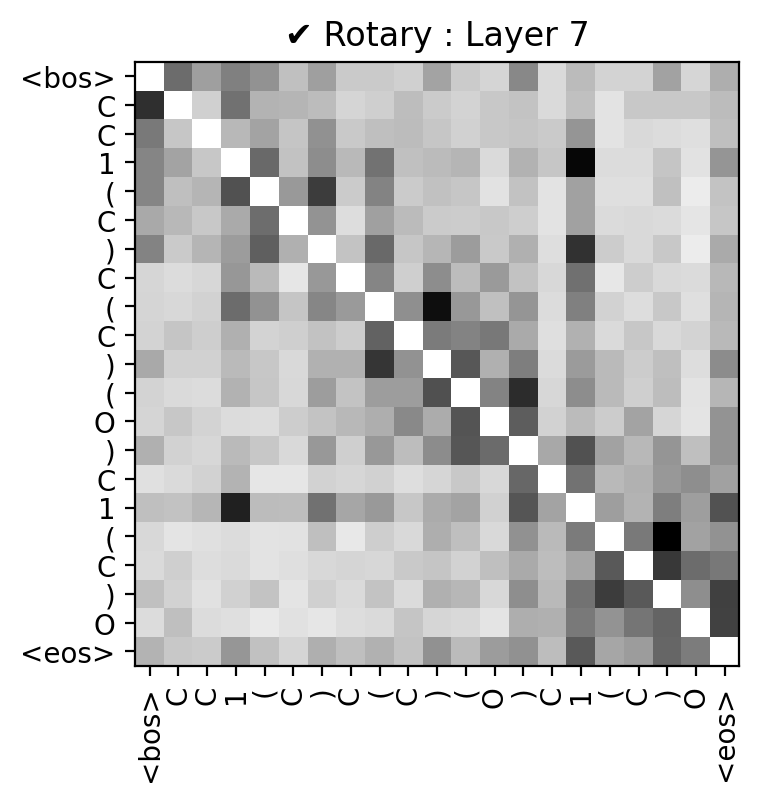}\hfill
  \includegraphics[width=.15\linewidth]{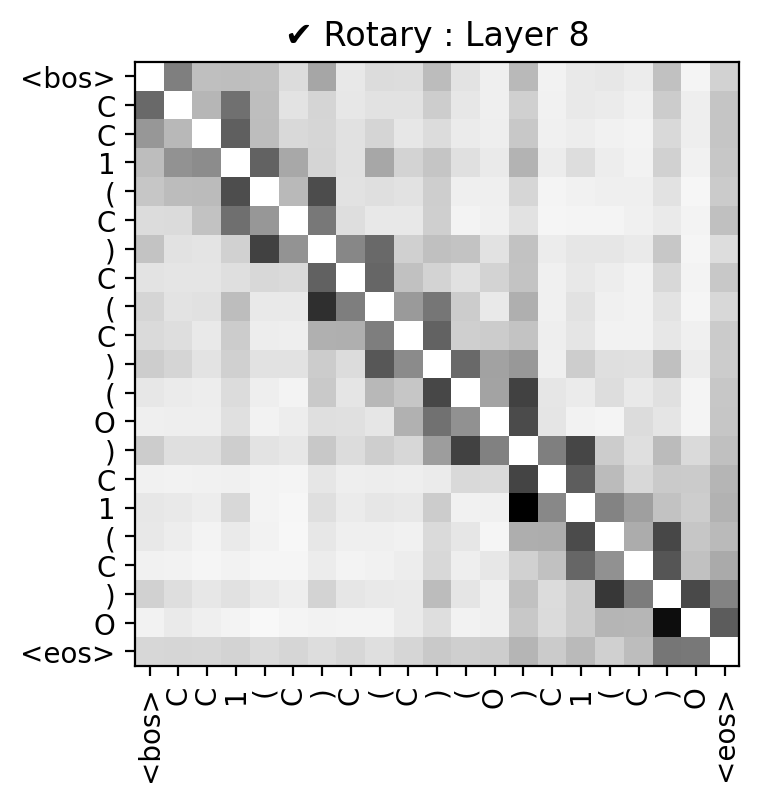}\hfill
  \includegraphics[width=.15\linewidth]{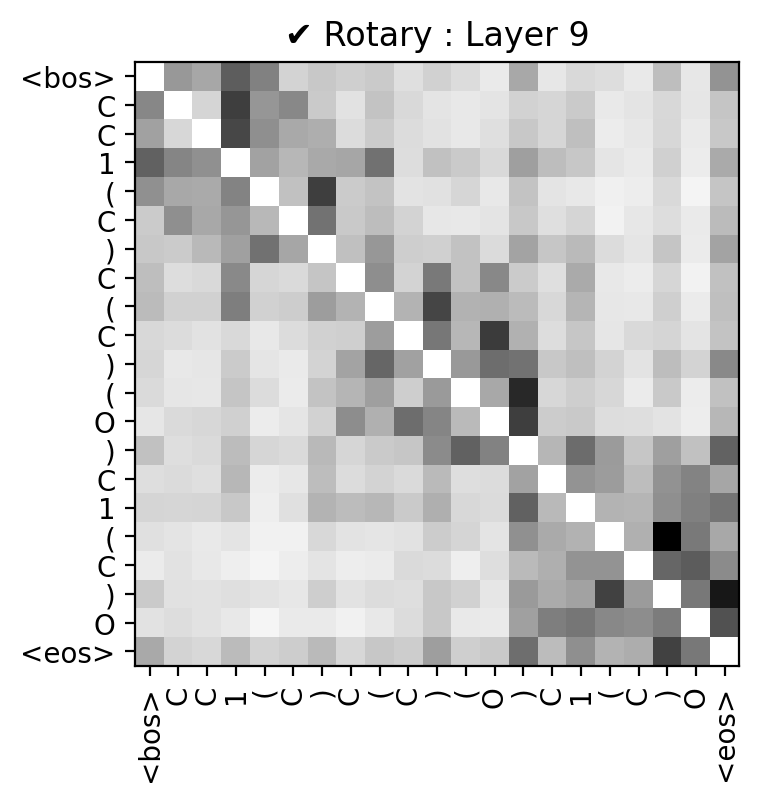}\hfill
  \includegraphics[width=.15\linewidth]{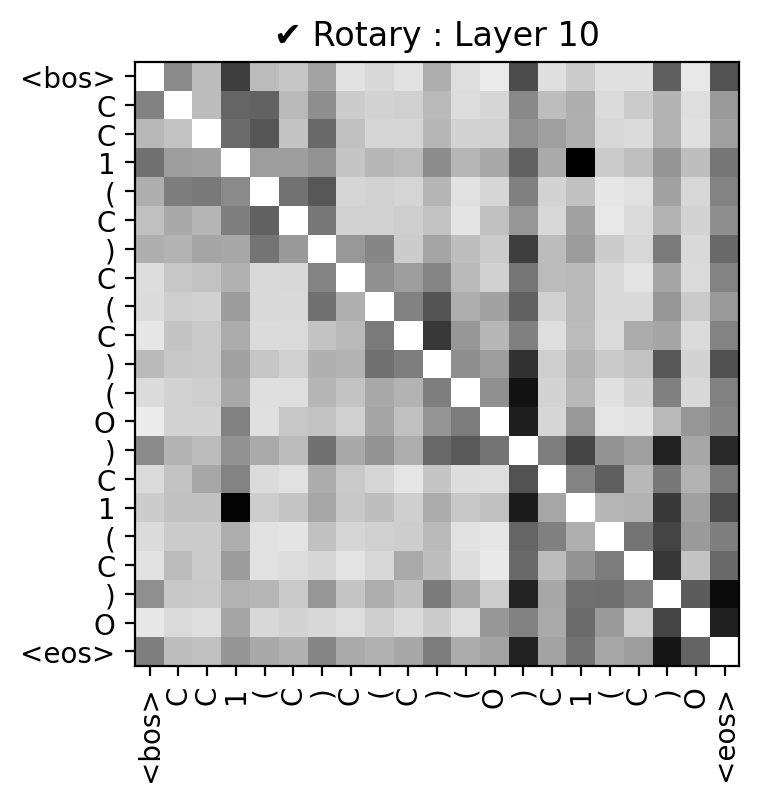}\hfill
  \includegraphics[width=.15\linewidth]{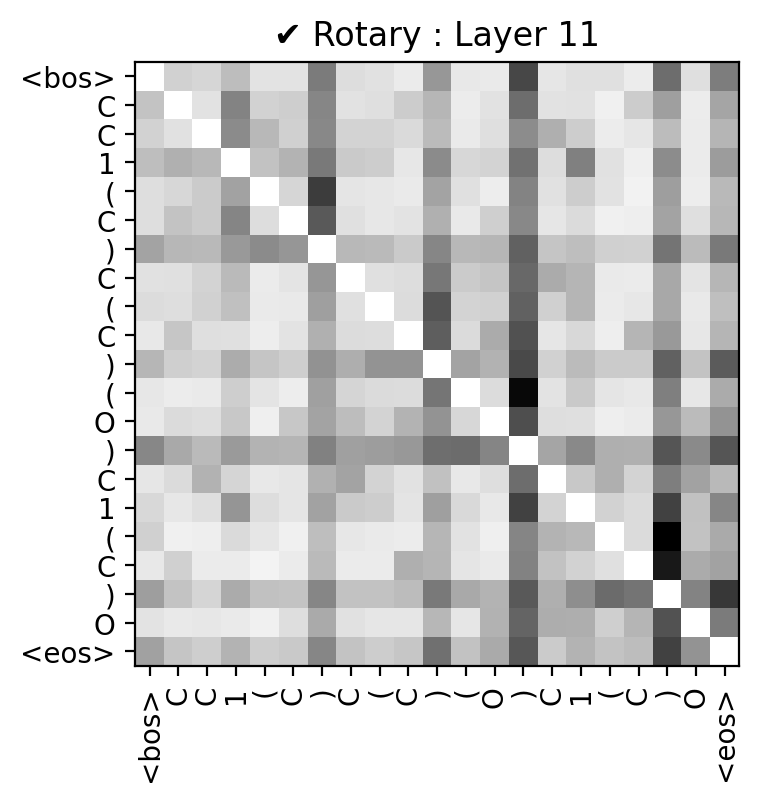}\hfill
  \includegraphics[width=.15\linewidth]{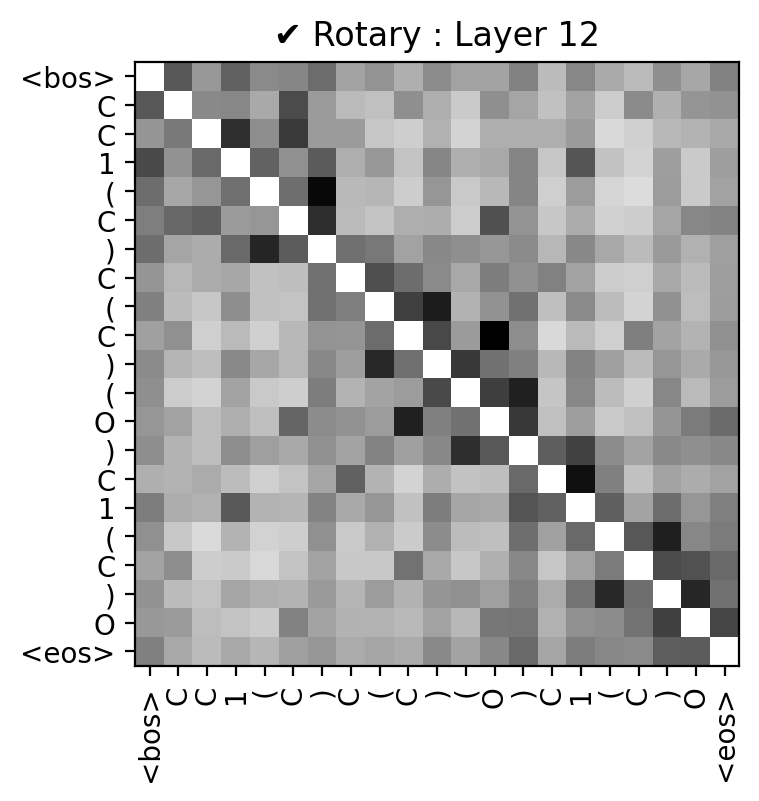}\hfill
  \caption{Attention Map for full-attention \& rotary/relative  embedding variant of \molformer}
  \end{subfigure}\par\medskip
  \begin{subfigure}{\linewidth}
  \includegraphics[width=.15\linewidth]{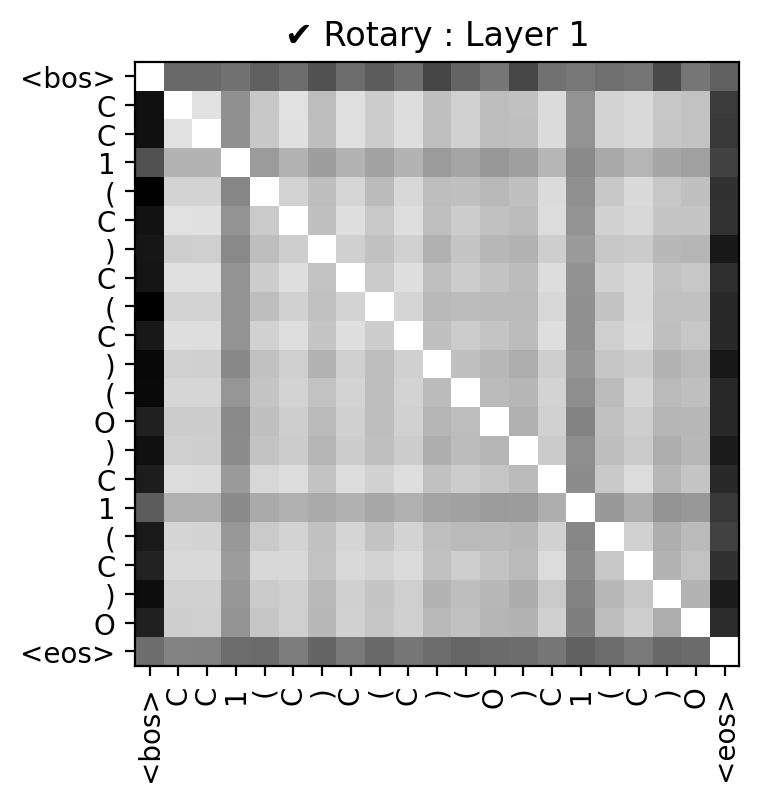}\hfill
  \includegraphics[width=.15\linewidth]{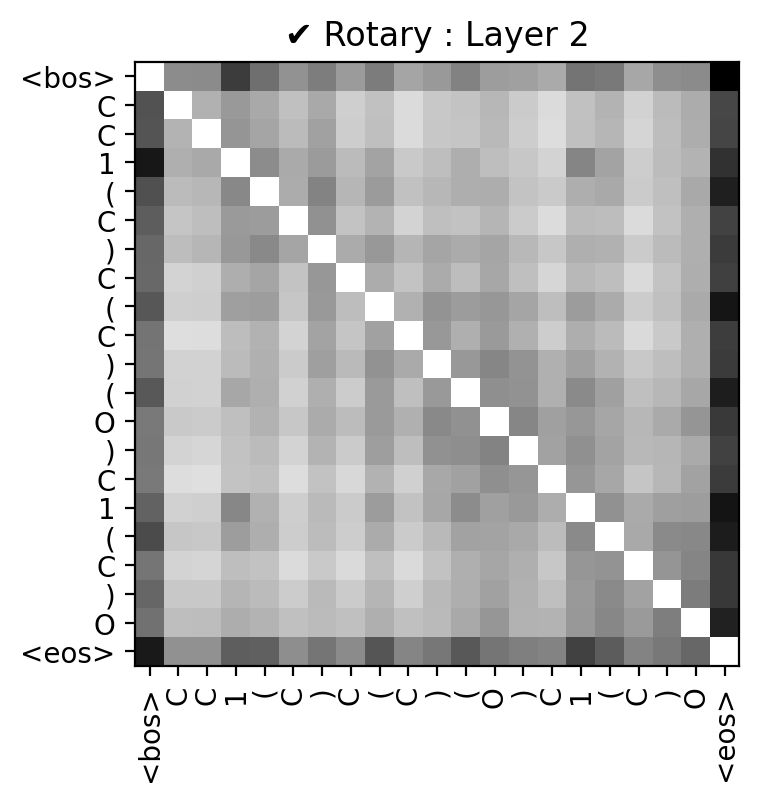}\hfill
  \includegraphics[width=.15\linewidth]{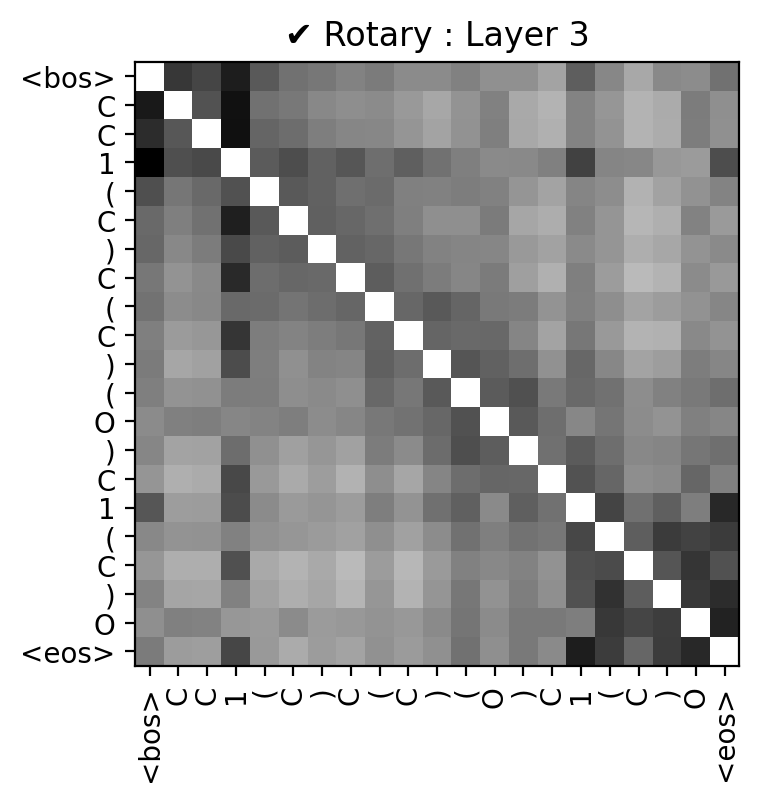}\hfill
  \includegraphics[width=.15\linewidth]{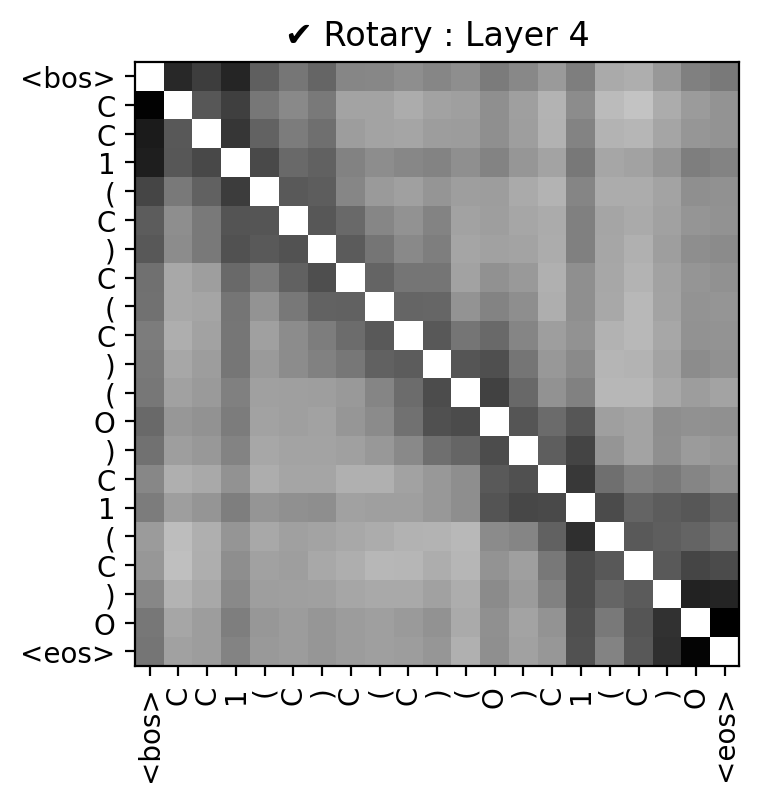}\hfill
  \includegraphics[width=.15\linewidth]{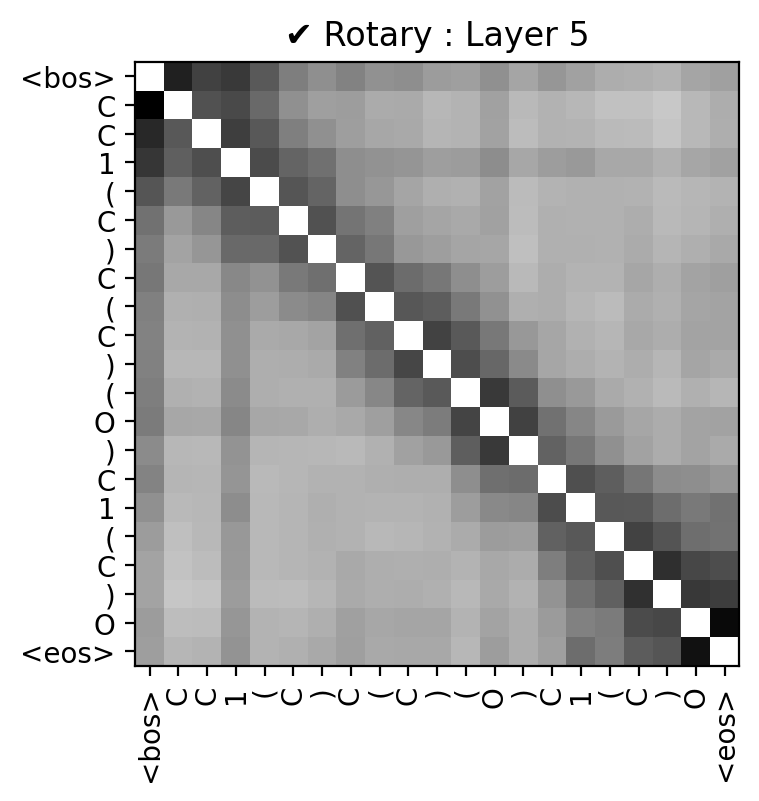}\hfill
  \includegraphics[width=.15\linewidth]{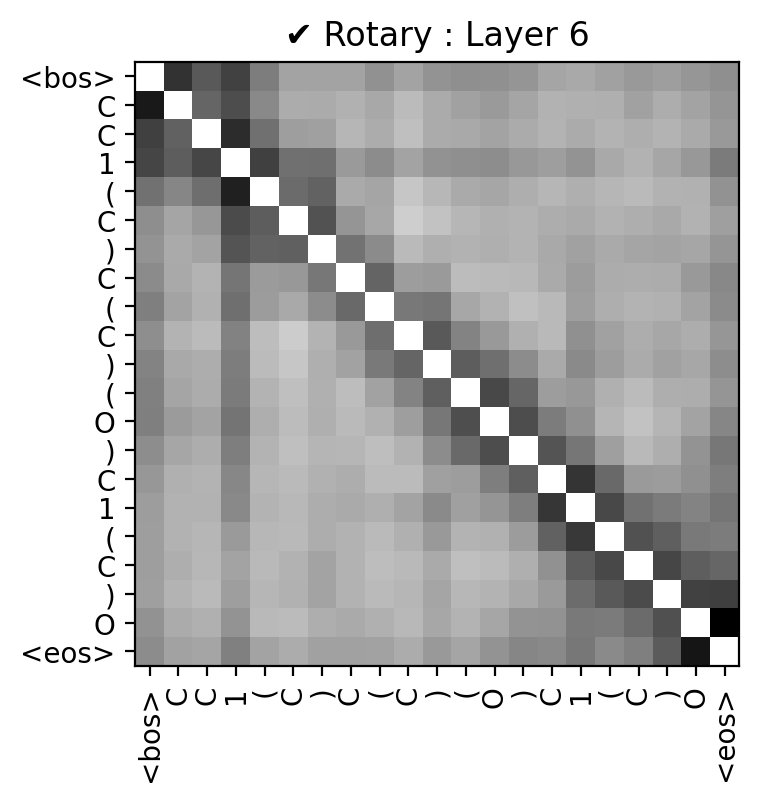}\hfill
  \includegraphics[width=.15\linewidth]{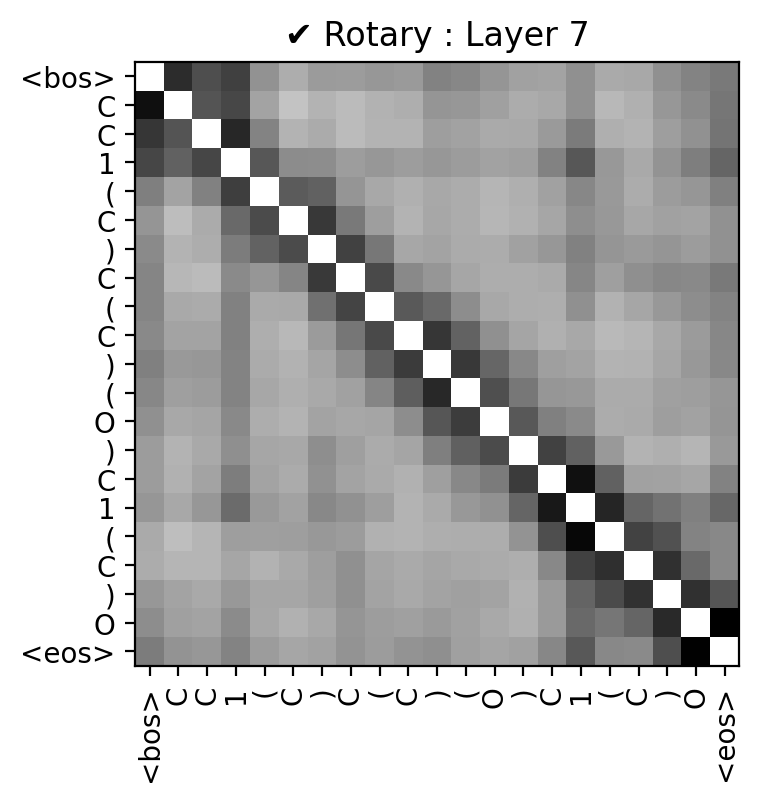}\hfill
  \includegraphics[width=.15\linewidth]{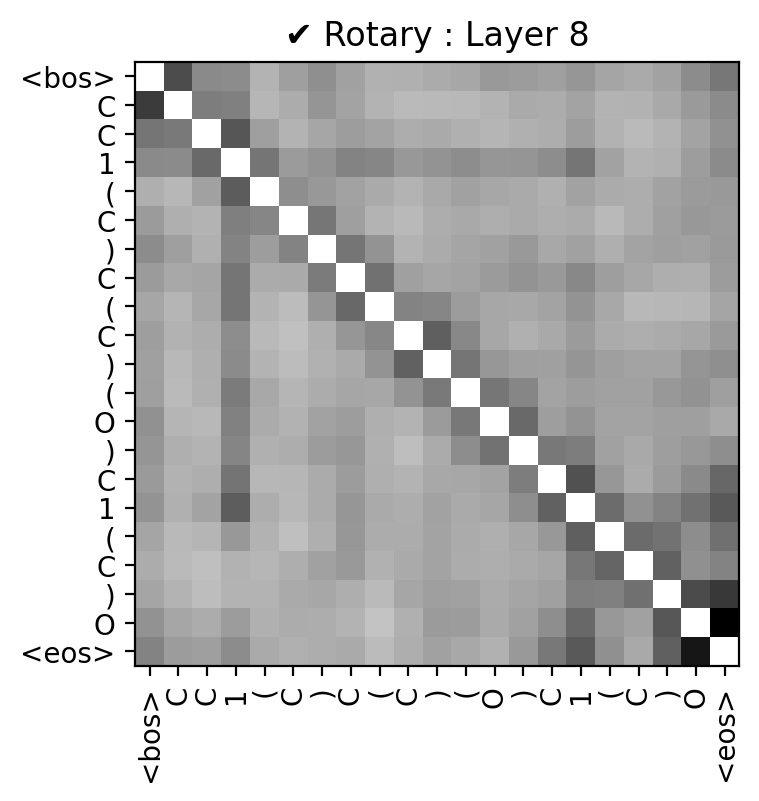}\hfill
  \includegraphics[width=.15\linewidth]{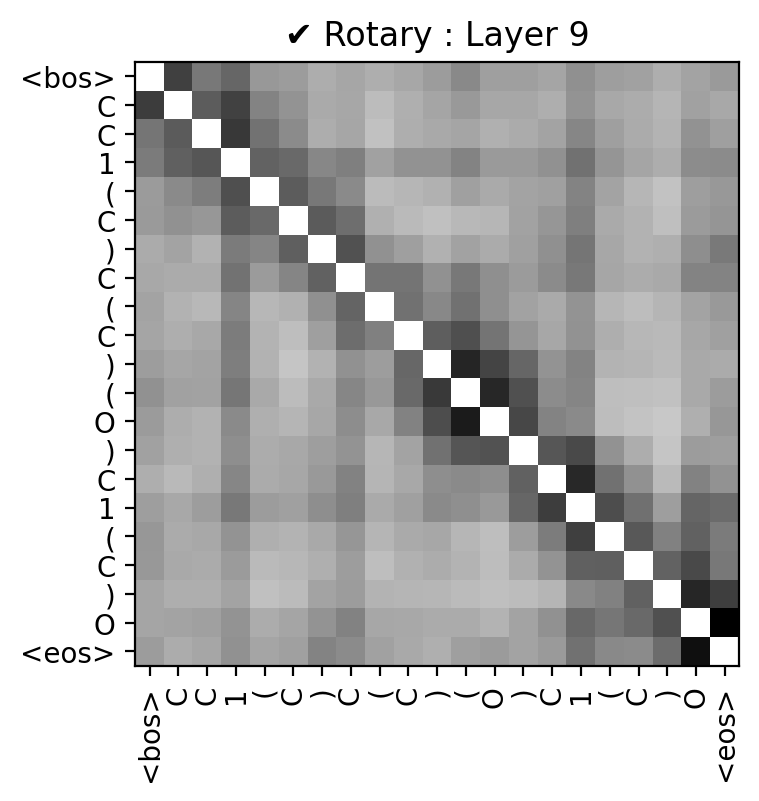}\hfill
  \includegraphics[width=.15\linewidth]{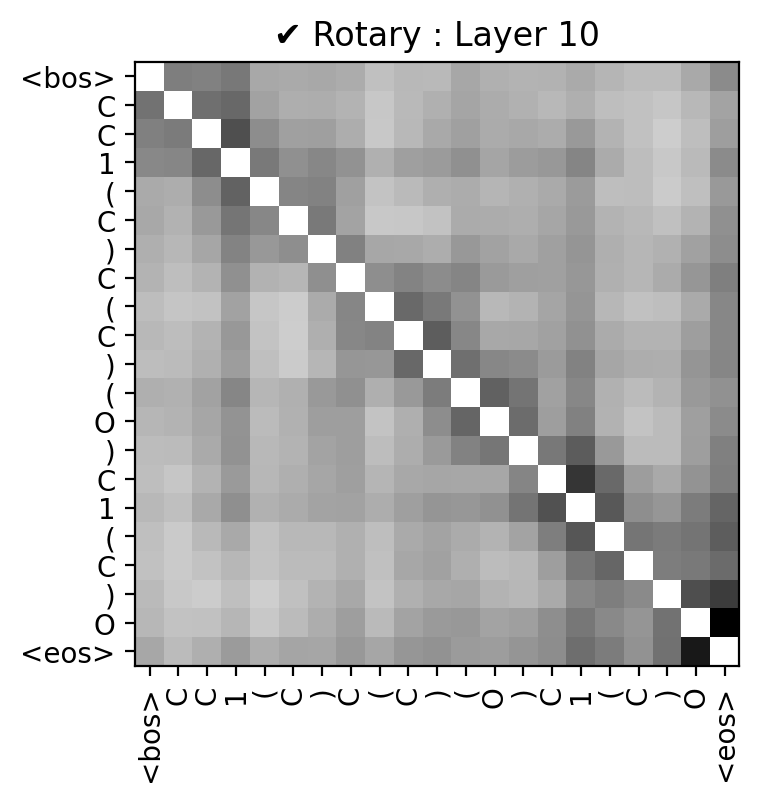}\hfill
  \includegraphics[width=.15\linewidth]{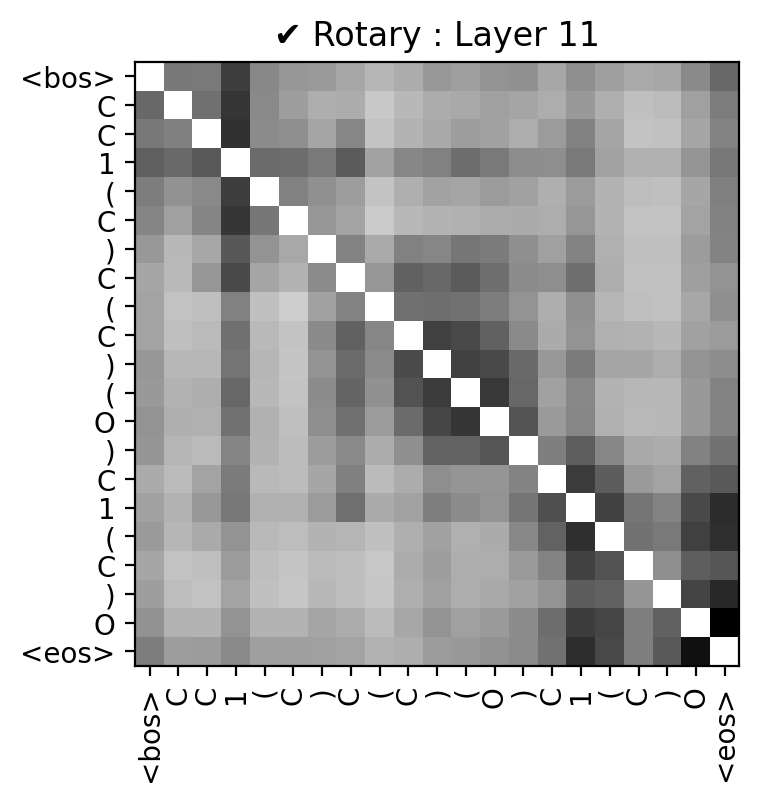}\hfill
  \includegraphics[width=.15\linewidth]{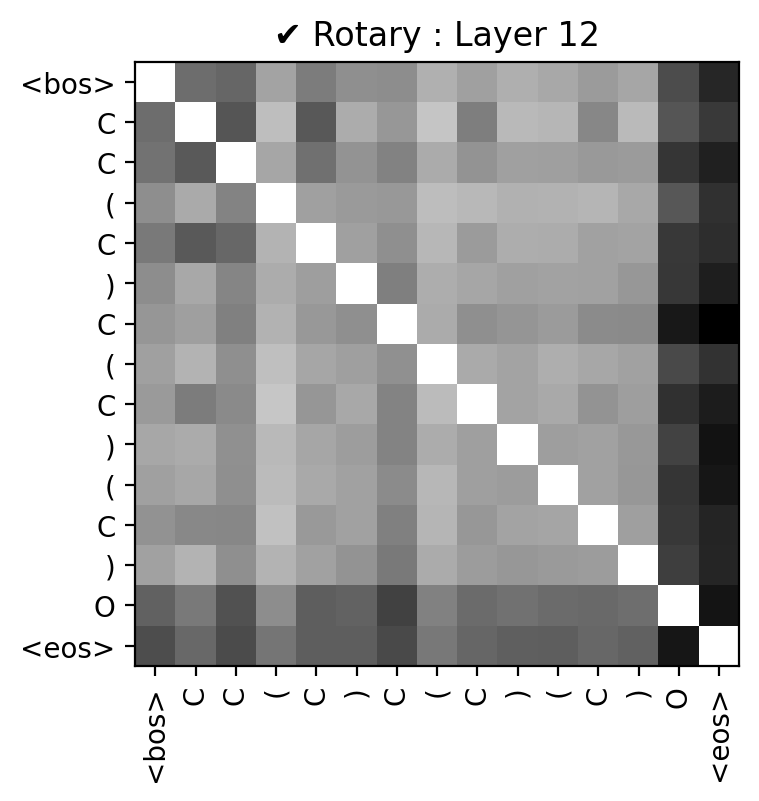}\hfill
  \caption{Attention Map for linear-attention \& rotary/relative  embedding variant of \molformer}
  \end{subfigure}\par\medskip
  \caption{Attention map shown for the Smiles sequence, \texttt{gdb\_62509}, \texttt{`CC1(C)C(C)(O)C1(C)O'} for all the layers of \molformer. For each layer, the attentions are average-pooled across all the heads.}
  \label{fig:attention_map_seq1}
\end{figure}

\begin{figure}
  \begin{subfigure}{\linewidth}
  \includegraphics[width=.15\linewidth]{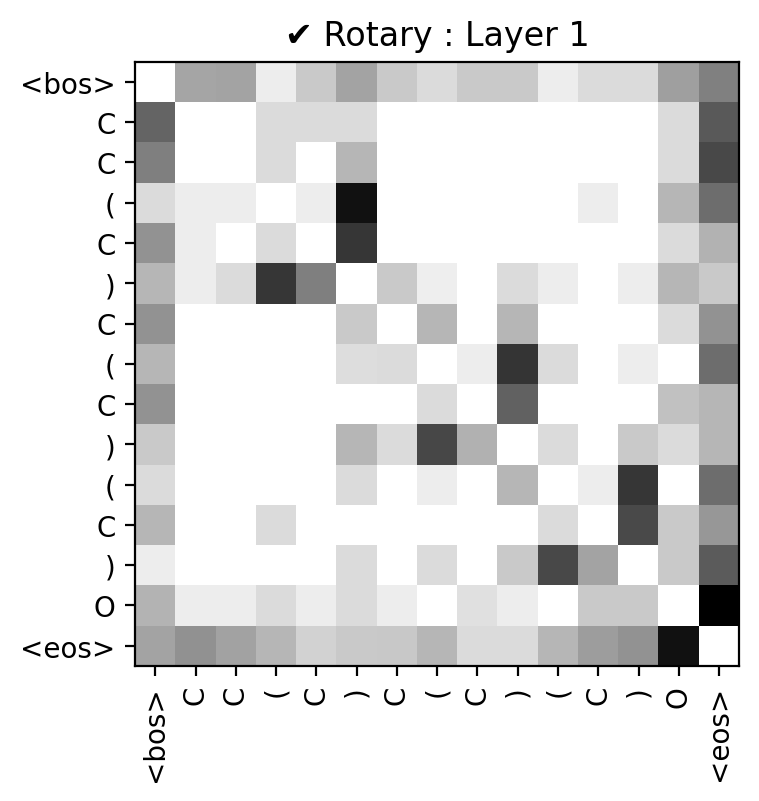}\hfill
  \includegraphics[width=.15\linewidth]{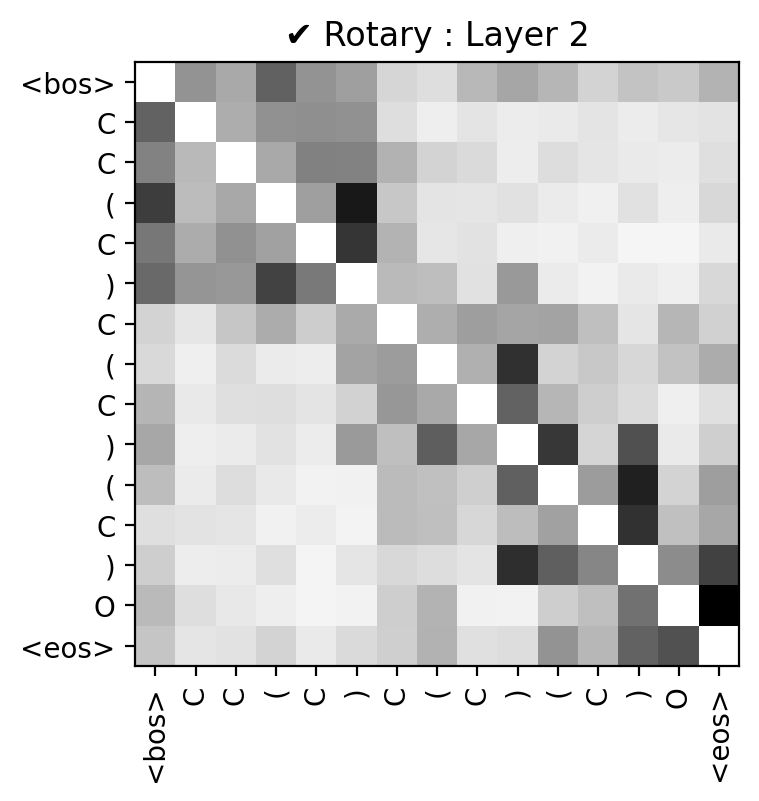}\hfill
  \includegraphics[width=.15\linewidth]{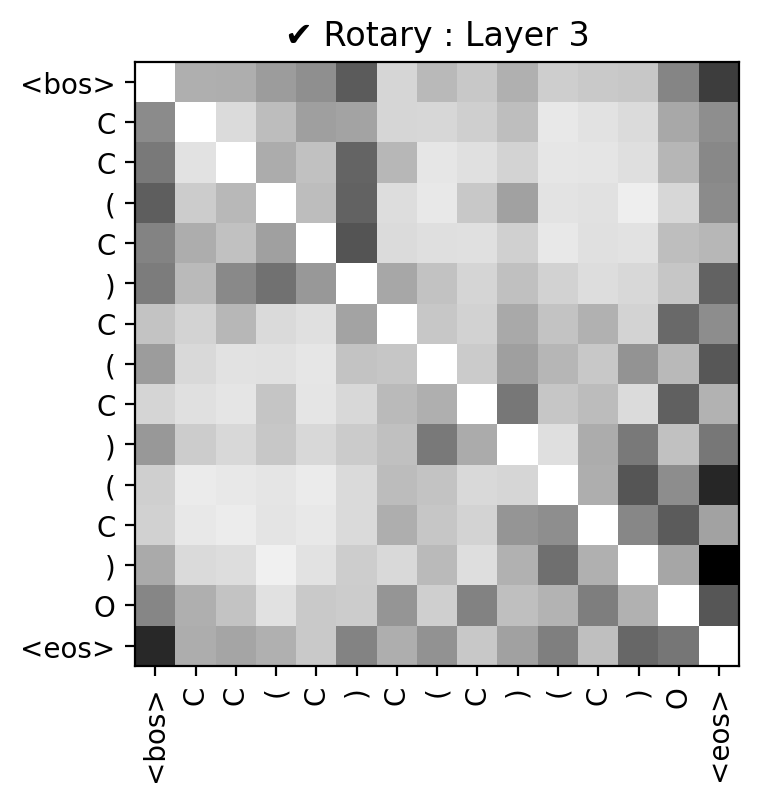}\hfill
  \includegraphics[width=.15\linewidth]{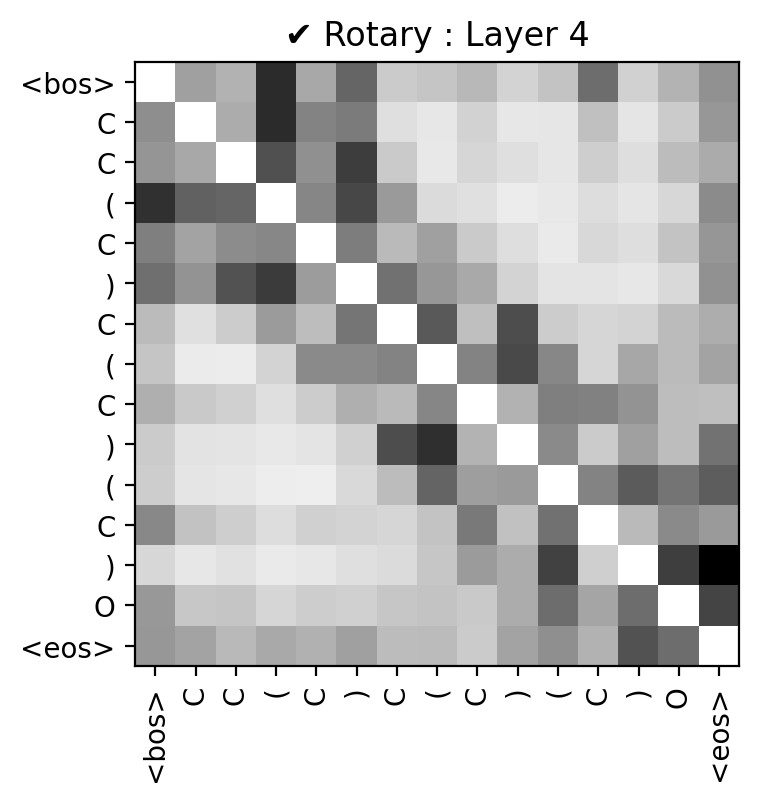}\hfill
  \includegraphics[width=.15\linewidth]{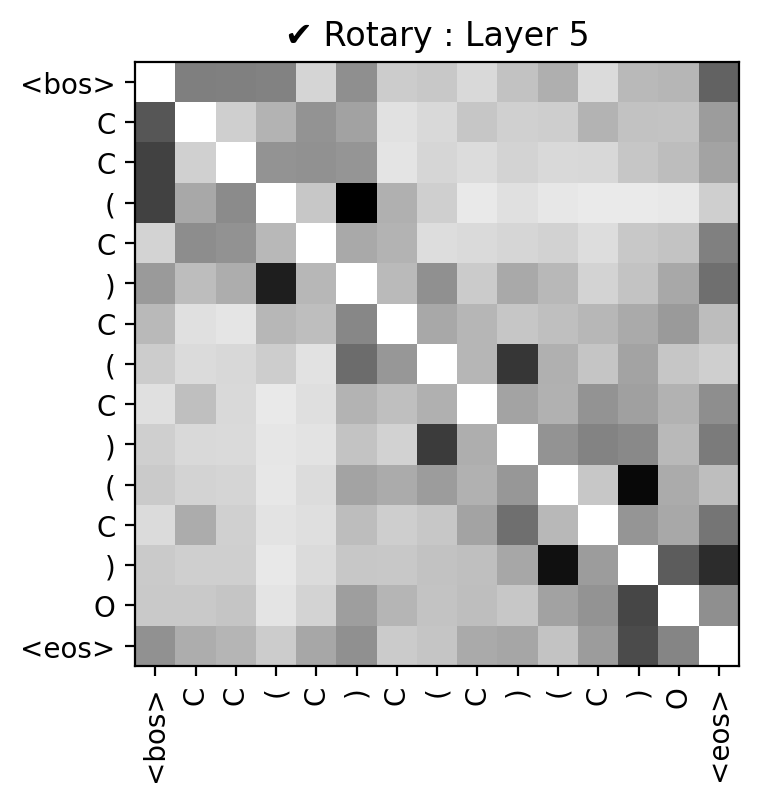}\hfill
  \includegraphics[width=.15\linewidth]{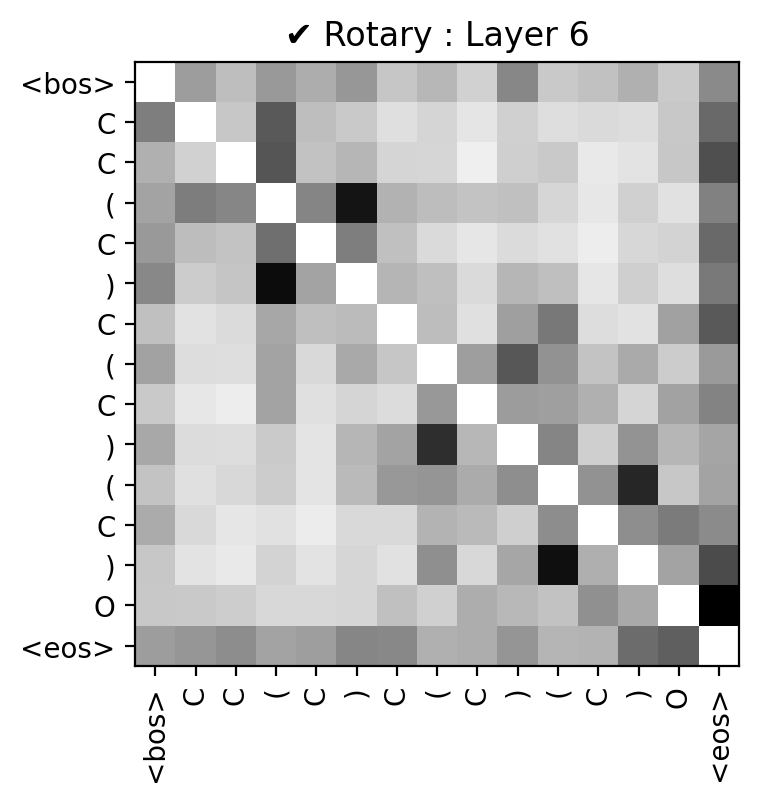}\hfill
  \includegraphics[width=.15\linewidth]{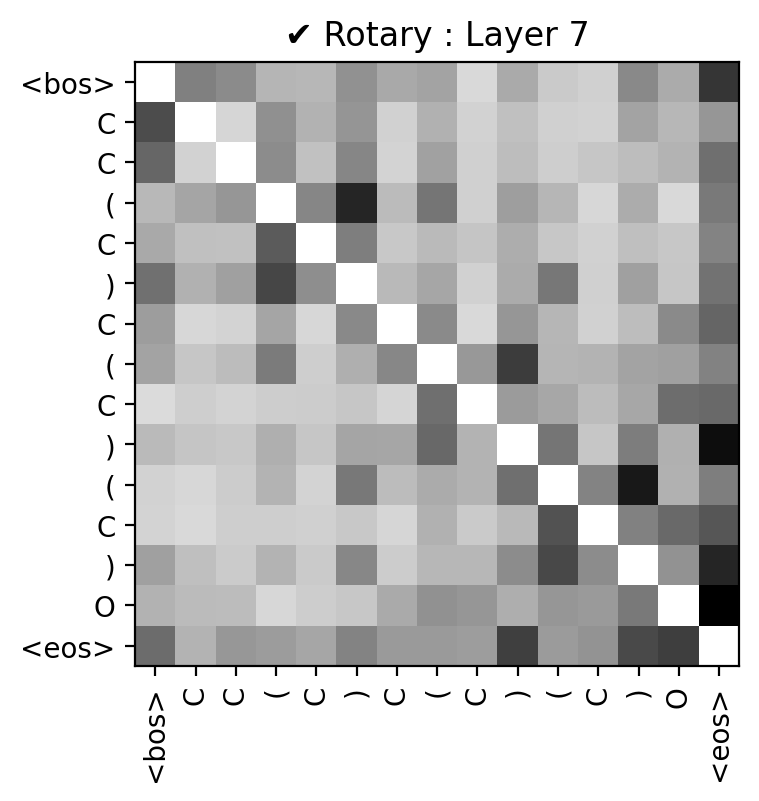}\hfill
  \includegraphics[width=.15\linewidth]{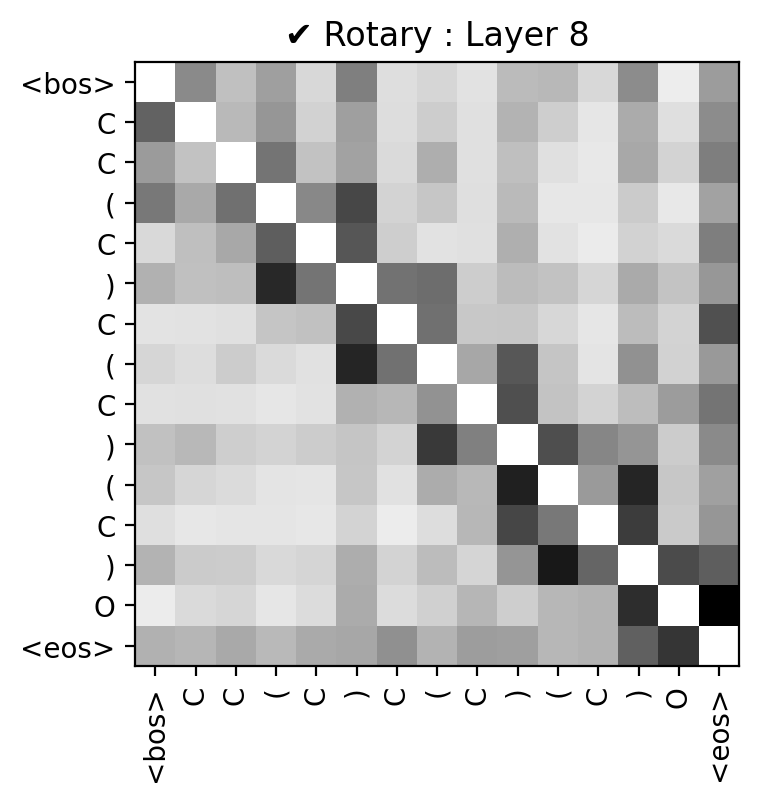}\hfill
  \includegraphics[width=.15\linewidth]{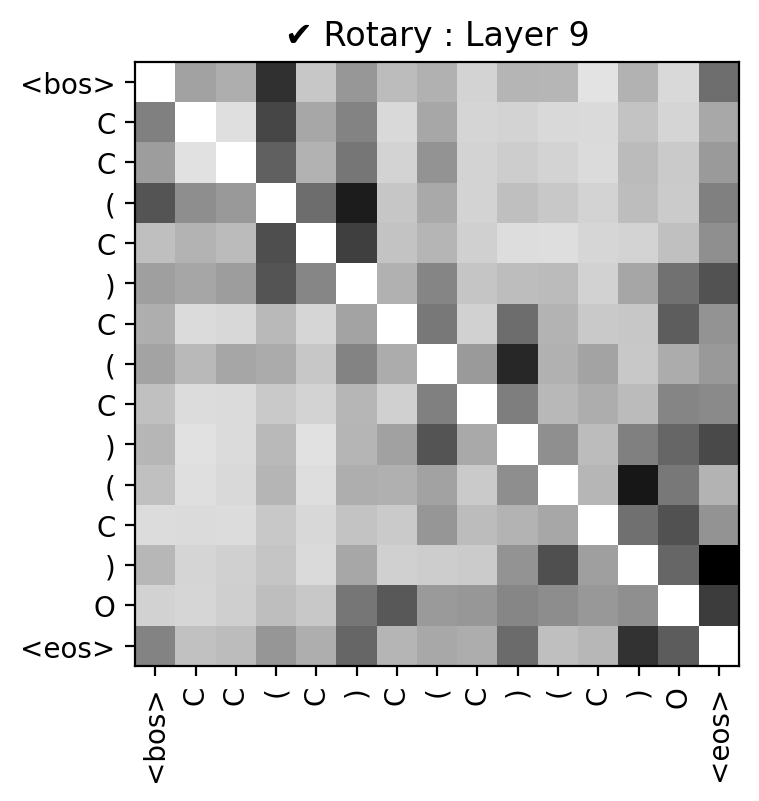}\hfill
  \includegraphics[width=.15\linewidth]{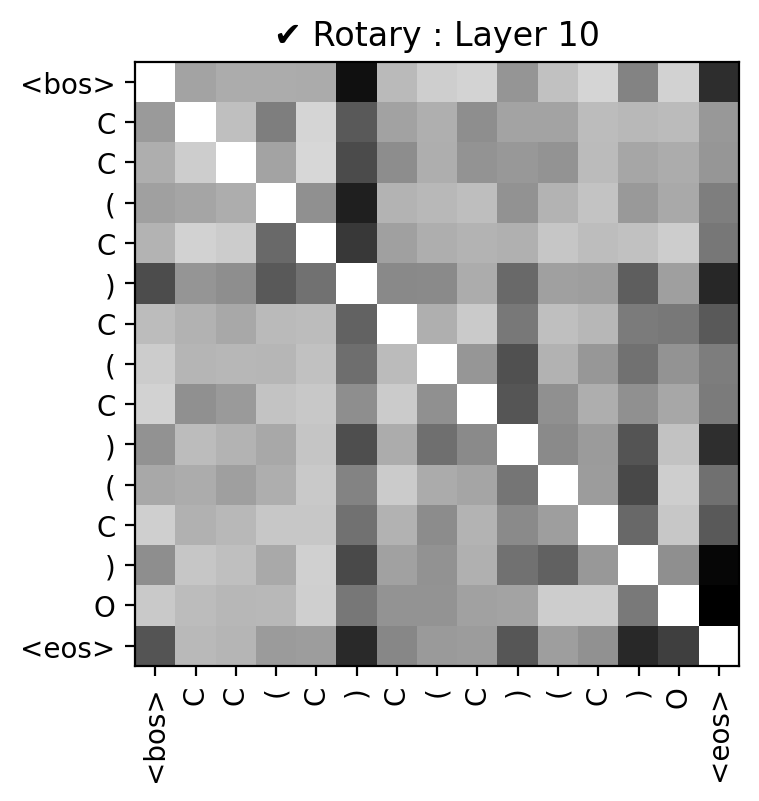}\hfill
  \includegraphics[width=.15\linewidth]{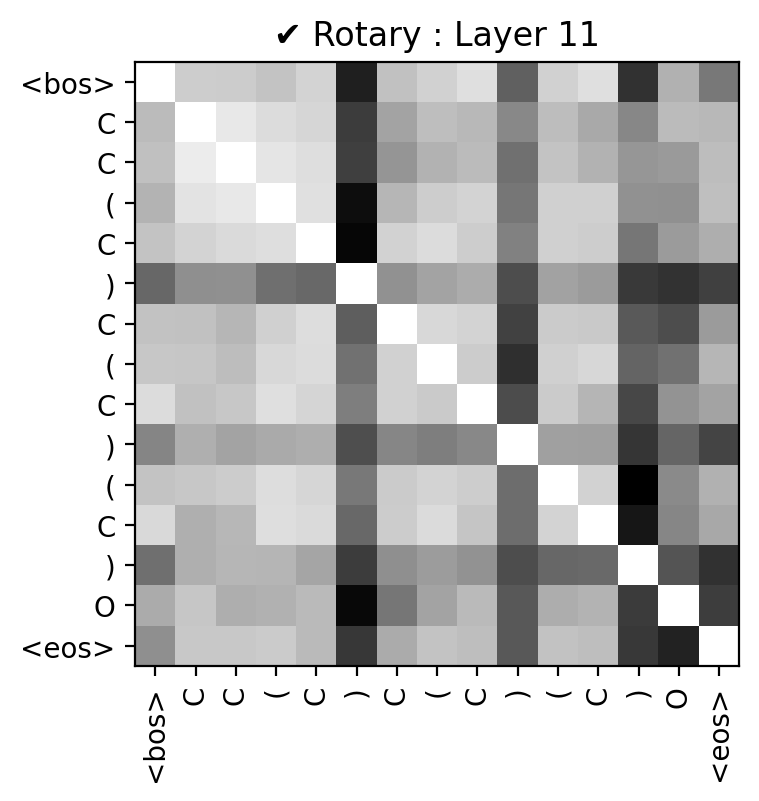}\hfill
  \includegraphics[width=.15\linewidth]{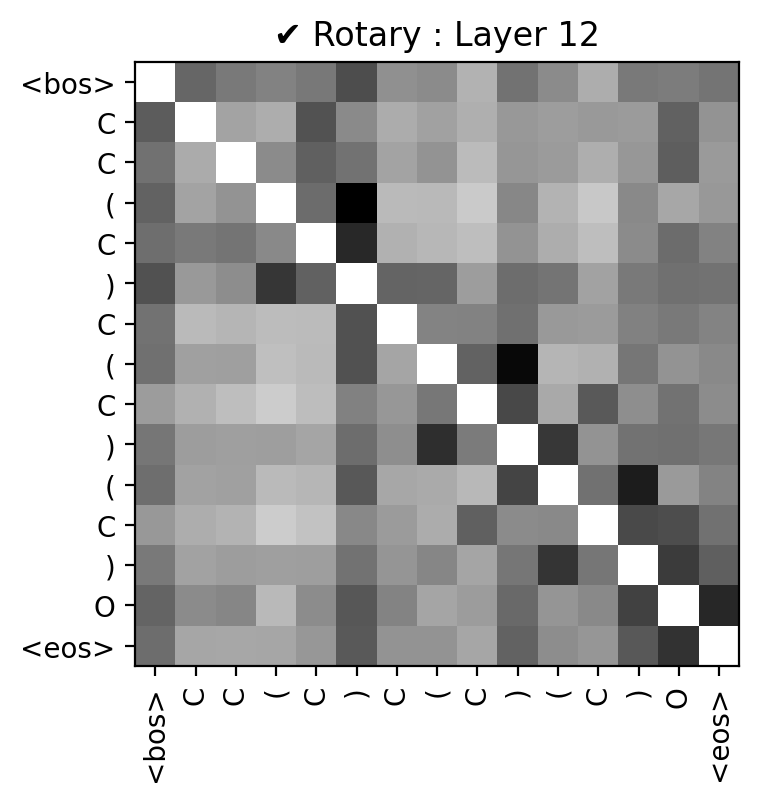}\hfill
  \caption{Attention map for full-attention \& rotary/relative embedding variant of \molformer}
  \end{subfigure}\par\medskip
  \begin{subfigure}{\linewidth}
  \includegraphics[width=.15\linewidth]{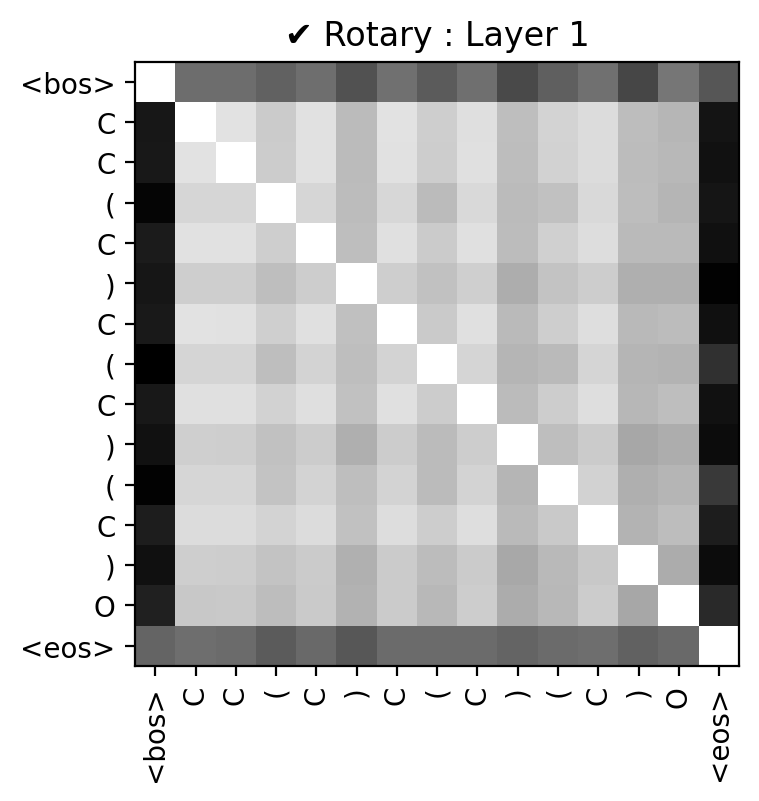}\hfill
  \includegraphics[width=.15\linewidth]{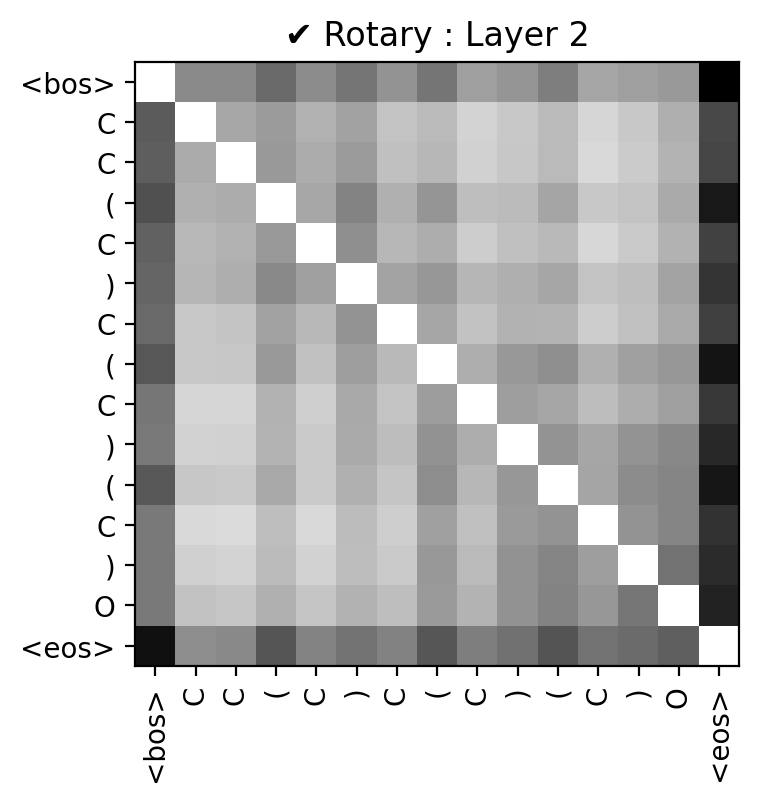}\hfill
  \includegraphics[width=.15\linewidth]{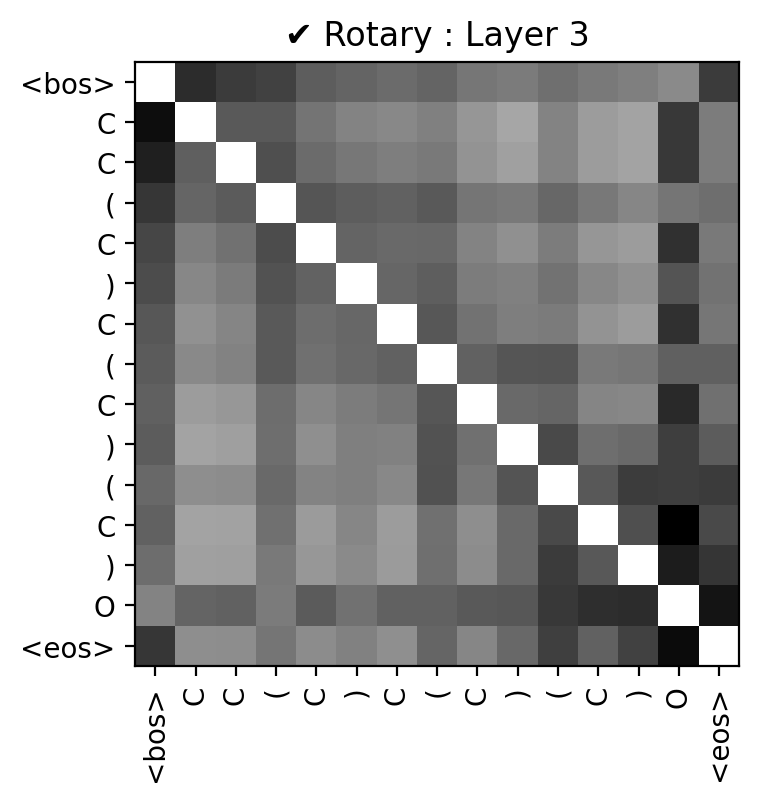}\hfill
  \includegraphics[width=.15\linewidth]{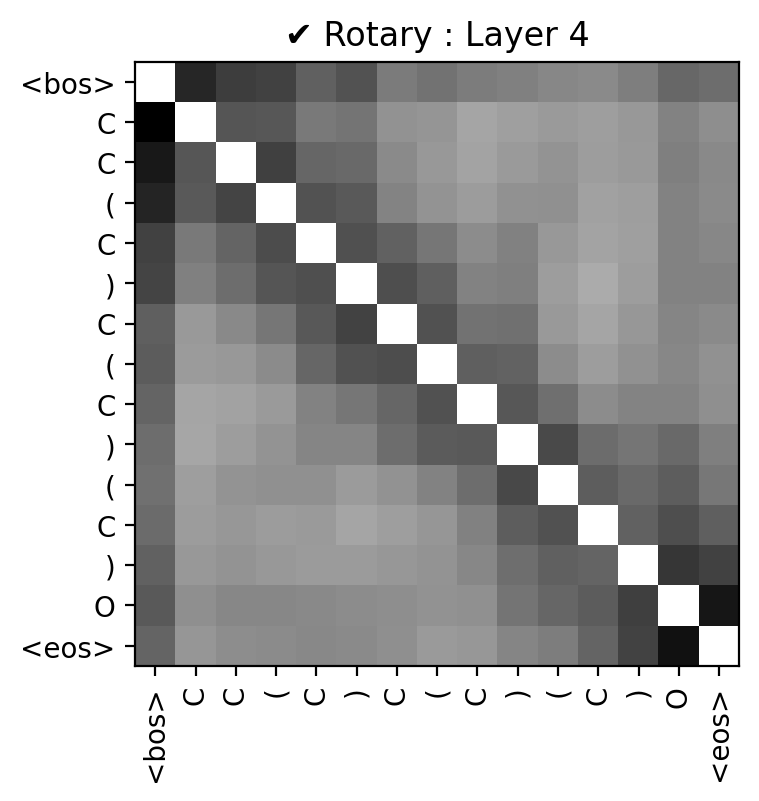}\hfill
  \includegraphics[width=.15\linewidth]{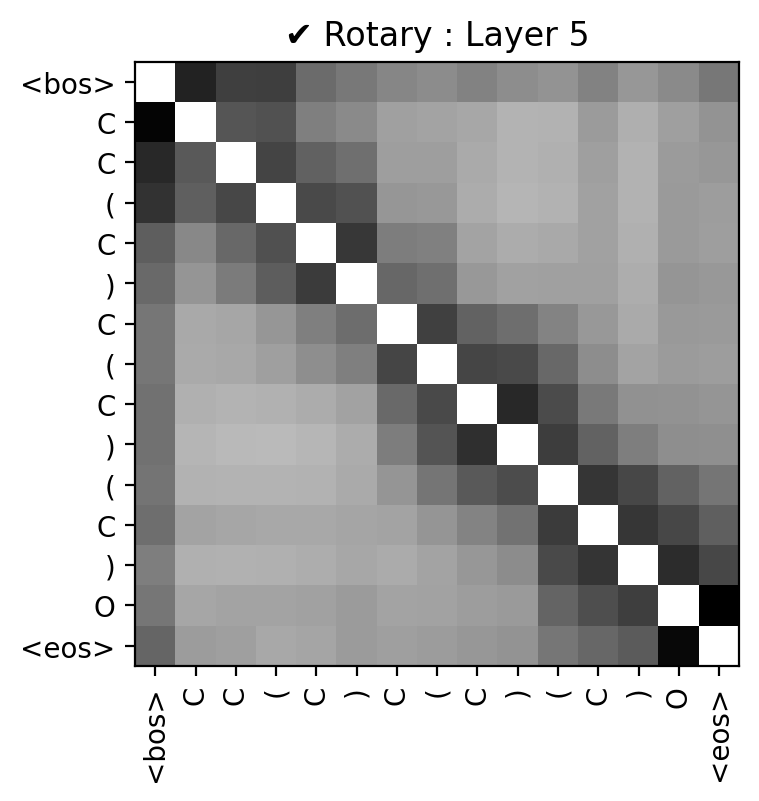}\hfill
  \includegraphics[width=.15\linewidth]{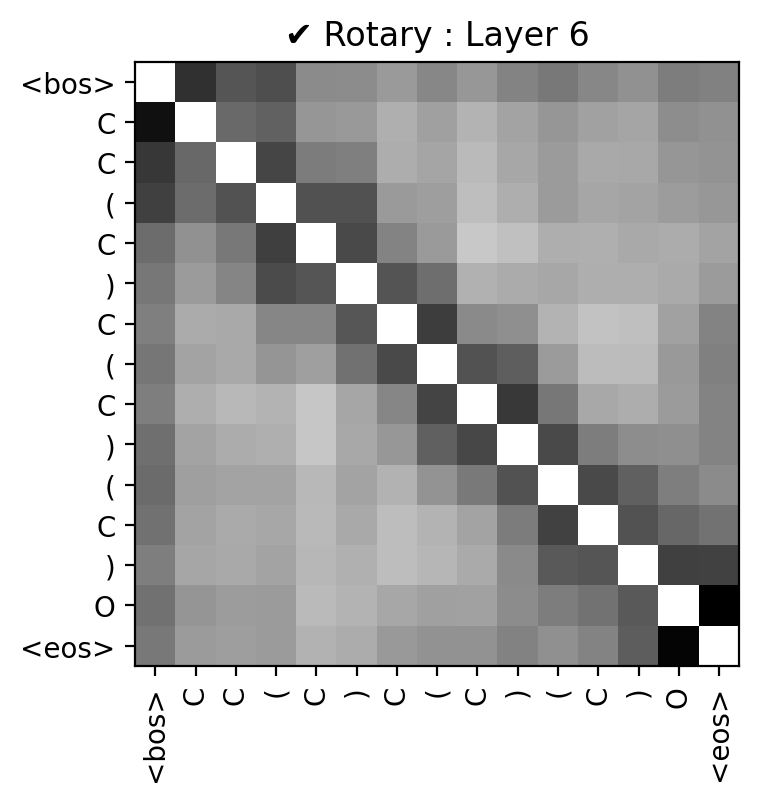}\hfill
  \includegraphics[width=.15\linewidth]{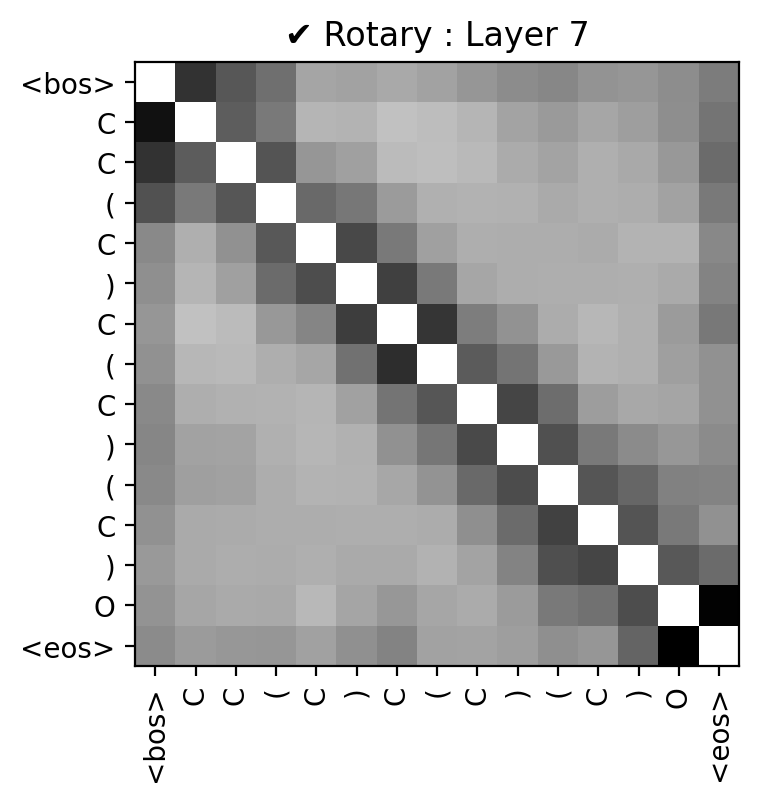}\hfill
  \includegraphics[width=.15\linewidth]{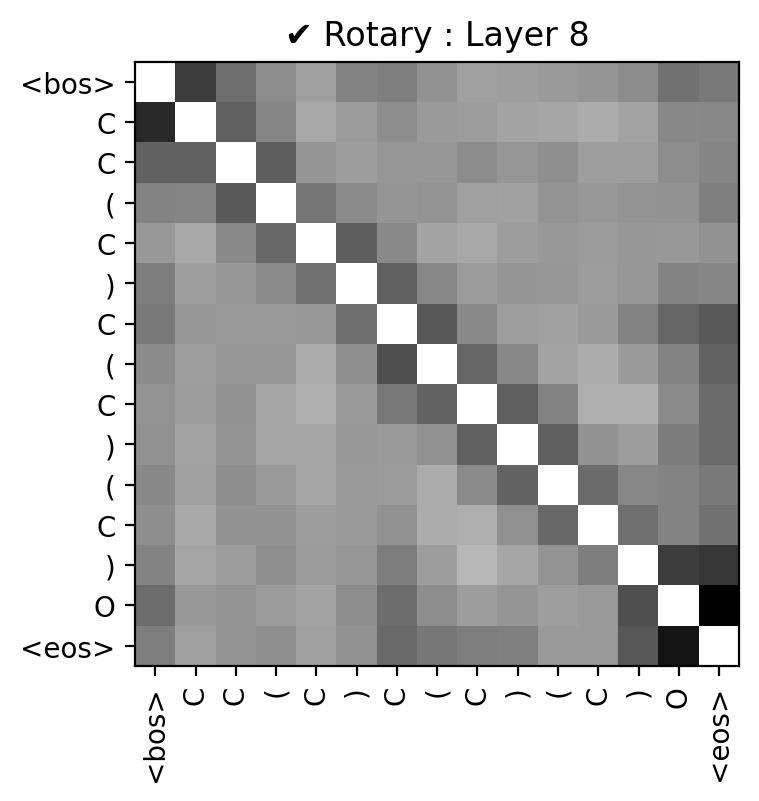}\hfill
  \includegraphics[width=.15\linewidth]{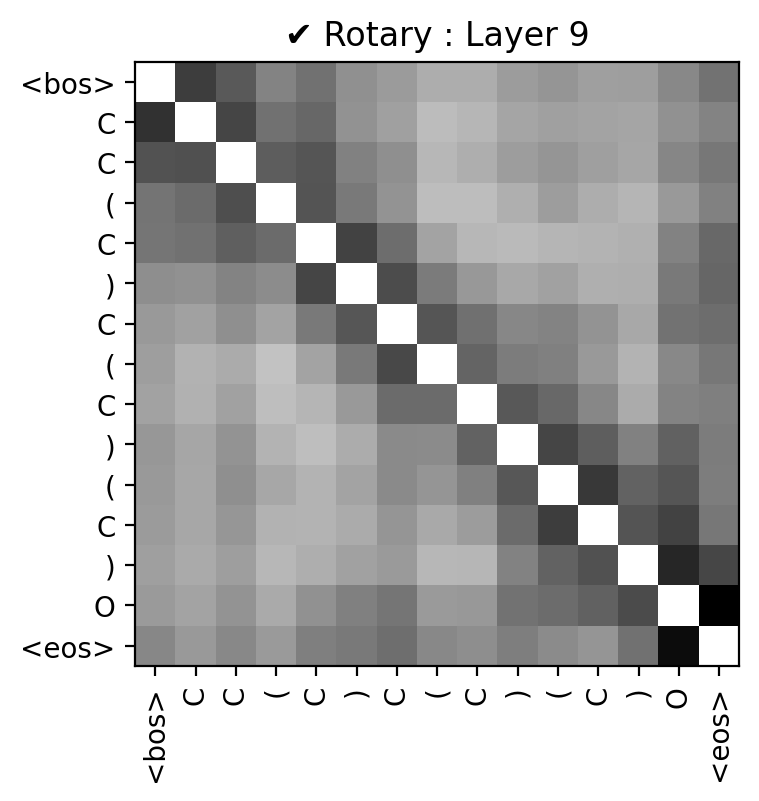}\hfill
  \includegraphics[width=.15\linewidth]{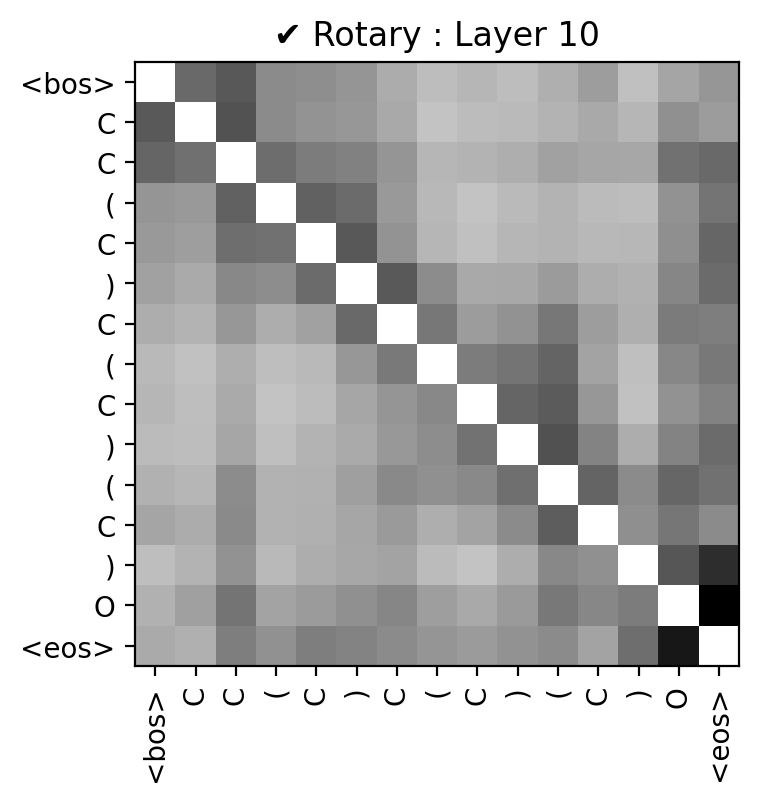}\hfill
  \includegraphics[width=.15\linewidth]{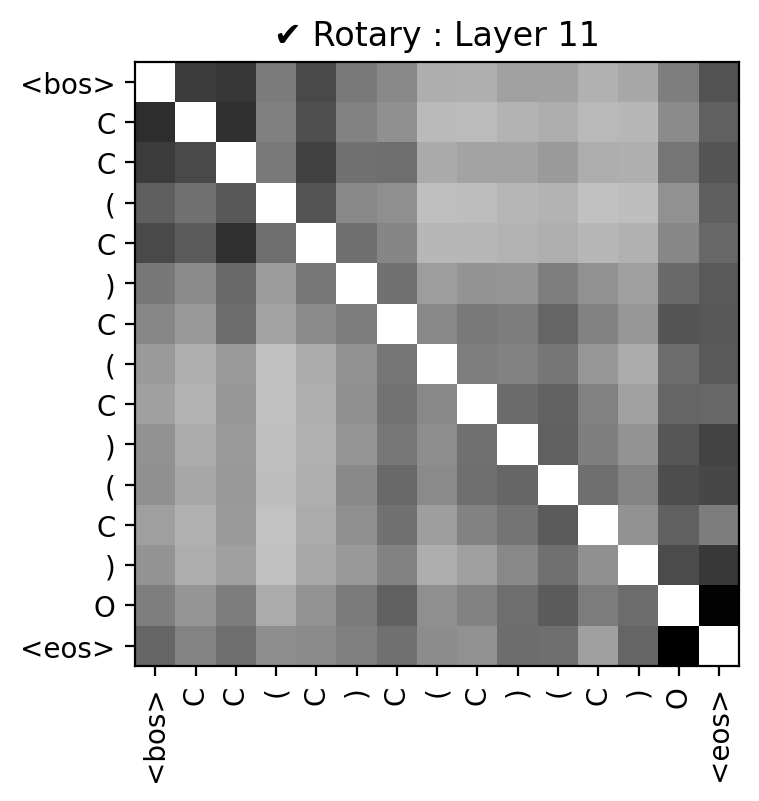}\hfill
  \includegraphics[width=.15\linewidth]{figures/figures_attention/linear_rotary/appendix_rot_linear_seq2_layer11.png}\hfill
  \caption{Attention Map for linear-attention \& rotary/relative embedding variant of \molformer}
  \end{subfigure}\par\medskip
  \caption{Attention map shown for the smile sequence, \texttt{gdb\_1105}, \texttt{`CC(C)C(C)(C)O'} for all the layers of \molformer. For each layer, the attentions are average-pooled across all the heads.}
\label{fig:attention_map_seq2}
\end{figure}

 In this section we present a visual comparison to show the attention representations of two molecules from the QM9 test dataset (\texttt{gdb\_62509} and \texttt{gdb\_1105}) generated by the linear and full attention model variants. Both these models of \molformer-XL have rotary positional embedding in place. We picked \texttt{gdb\_62509} and \texttt{gdb\_1105} based on the best cosine similarity from the medium bucket category, and are the same as the ones used in Table \ref{tab:quantitative_attention}. For full quantitative comparison on this metric, refer to Table \ref{tab:quantitative_attention}. The attention head weights of each layer are averaged and all layers are presented in figures \ref{fig:attention_map_seq1} and \ref{fig:attention_map_seq2}. Please note that the colorbar on the subplots are of different scale for different variants. Several interesting observations can be made from these representations. For both full-attention and its linear counterpart most of the attentions are directed toward the closing parentheses in higher layers (layer 9 and up). Therefore, it is reasonable to avoid other layers for identifying meaningful features with respect to structure. 
 We also notice that intermediate layers 8 and 9, from the linear-attention with rotary variant, capture the 3D structure of the molecules better. This also reinforces observation made in quantitative analysis reflected in Table \ref{tab:quantitative_attention}.   
 
 While the model variations have similar in downstream task performance in our evaluations the differences in their attention weights and in their ability of discovering structural information from SMILES representation are insightful  and intriguing. Specifically with regard to how the linear-attention embedding captures structural information of molecules. Also, it is worth observing that the \molformer-XL variants here are not fine-tuned for the QM9 dataset and yet the attentions manage to represent the structure.
 

\FloatBarrier
\section{Assets and License}\label{app:assets}

The following table summarizes the libraries utilized for our experiments and their accompanying license terms.

\begin{center}
\resizebox{\textwidth}{!}{\begin{tabular}{ l|l|l }
\toprule
 Asset & License & Link \\
\midrule
 Fast-Transformers & MIT License & https://github.com/idiap/fast-transformers/blob/master/LICENSE \\ 
 Pytorch & BSD Style License & https://github.com/pytorch/pytorch/blob/master/LICENSE \\  
 RDKit & BSD 3-Clause "New" or "Revised" License & https://github.com/rdkit/rdkit/blob/master/license.txt \\
 Pytorch Lightning & Apache 2.0 License & https://github.com/PyTorchLightning/pytorch-lightning/blob/master/LICENSE \\
 HuggingFace Datasets & Apache 2.0 License & https://github.com/huggingface/datasets/blob/master/LICENSE \\
 Nvidia APEX & BSD 3-Clause "New" or "Revised" License &  https://github.com/NVIDIA/apex/blob/master/LICENSE \\ 
\hline
\end{tabular}}
\end{center}

\end{document}